%% file: ic-si-2026-foiks2026-conditional_constraint_splittings.tex
\documentclass[a4paper,fleqn]{cas-sc}

\usepackage[numbers]{natbib}

\usepackage{graphicx}

\usepackage{subcaption}

\usepackage{url}
\usepackage{amsmath}
\usepackage{amssymb}
\usepackage{booktabs}
\usepackage{algorithm}
\usepackage[noend]{algpseudocode} 
\usepackage{aliascnt}
\usepackage{xspace}
\usepackage{cleveref}
\usepackage{xcolor}
\usepackage{esvect}
\usepackage{tikz}
\usetikzlibrary{calc, arrows.meta}
\usepackage[bibliography=common, appendix=inline]{apxproof}

\allowdisplaybreaks

\usepackage{csquotes}  
\usepackage[misc,geometry]{ifsym}

\input{macro_files_stripped}

\newdefinition{definition}{Definition}
\newdefinition{example}{Example}

\newtheorem{theorem}{Theorem}
\newtheoremrep{proposition}[theorem]{Proposition}
\newtheorem{lemma}[theorem]{Lemma}
\newtheorem{corollary}[theorem]{Corollary}

\begin{document}
	\let\WriteBookmarks\relax
	\def\floatpagepagefraction{1}
	\def\textpagefraction{.001}
	
	\shorttitle{Splitting Belief Bases Conditionally based on Constraints for c-Representations and their Solutions}    
	
	\shortauthors{L. Spiegel, C. Beierle, G. Kern-Isberner, A. Hahn}  
	
	\title [mode = title]{
		Splitting Belief Bases Conditionally based on Constraints for c-Representations and their Solutions
		}  
	
	
	
	%
	
	\author[1]{Lars-Phillip Spiegel}[orcid=0009-0001-1962-752X]
	\cormark[1]
	\ead{lars-phillip.spiegel@fernuni-hagen.de}
	\credit{Conceptualization, Writing -- original draft, Writing -- review and editing}
	
	\author[1]{Christoph Beierle}[orcid=0000-0002-0736-8516]
	\ead{christoph.beierle@fernuni-hagen.de}
	\credit{Conceptualization, Writing -- original draft, Writing -- review and editing, Funding acquisition}
	
	\author[2]{Gabriele Kern-Isberner}[orcid=0000-0001-8689-5391]
	\ead{gabriele.kern-isberner@cs.tu-dortmund.de} 
	\credit{Conceptualization, Writing -- original draft, Writing -- review and editing, Funding acquisition} 
	
	\author[2]{Alexander Hahn}[orcid=0009-0008-6114-2594]
	\ead{alexander.hahn@tu-dortmund.de}
	\credit{Conceptualization, Writing -- original draft, Writing -- review and editing}

	\affiliation[1]{%
		organization={FernUniversität in Hagen},
		addressline={Universitätsstraße 11}, 
		city={Hagen},
		postcode={58097}, 
		country={Germany}
	}
	
	\affiliation[2]{%
		organization={TU Dortmund University},
		addressline={August-Schmidt-Straße 1}, 
		city={Dortmund},
		postcode={44227},
		country={Germany}
	}
	
	\cortext[1]{Corresponding author}
	
	
	\begin{abstract}
		Splitting notions for conditional belief bases aim at localizing reasoning and computation tasks to subbases containing only the relevant parts of the belief base. Among the different reasoning mechanisms, c-representations, a special kind of Spohn's ranking models, have been shown to provide high-quality inferences. c-Representations can be conveniently specified via a constraint satisfaction problem. In this article, we introduce the notion of conditional constraint splittings, extending constraint splittings by also allowing the subbases to share conditionals. 
		Taking inspiration from conditional syntax splittings, we identify the subclasses of safe and \niceSafe\ conditional constraint splittings,
		allowing us to locally and independently solve the constraint system on the subbases.
		We also generalize 
		solution splittings for c-representations
		to \gcondCSolSplit s
		and 
		provide an in-depth analysis of their relationship to
		conditional constraint splittings.
		Additionally, 
		we elaborate on the relationship of these new splitting notions
		to
		safe and generalized safe conditional syntax splittings and also to
		safe covers,
		which have been proposed 
		for localized computation of c-representations.
	\end{abstract}
	
	\begin{highlights}
		\item We generalize constraint splittings to \emph{conditional} constraint splittings, allowing subbases to depend on shared conditionals.
		\item We identify the subclasses of \emph{safe} and \emph{\niceSafe} conditional constraint splittings, allowing us to solve the underlying constraint satisfaction problem locally on the subbases.
		\item We generalize c-solution splittings to \emph{safe conditional} c-solution splittings, allowing subbases to also share conditionals.
		\item We establish the precise relationships among safe conditional constraint splittings, \niceSafe\ conditional constraint splittings, and \gcondCSolSplit s.
		\item We prove that every \niceSafe\ conditional constraint splitting can be transformed into a safe cover and vice versa.
		\item We illustrate the relationship of our approaches to other splitting techniques by presenting a comprehensive hierarchy of splittings.
	\end{highlights}

	
	\maketitle

\input{textbody_stripped}
	
	\section*{Acknowledgments}
		This work was supported by the Deutsche Forschungsgemeinschaft (DFG, German Research Foundation), project number 512363537,
		grant BE 1700/12-1 awarded to Christoph Beierle, and 
		grant KE 1413/15-1 awarded to Gabriele Kern-Isberner.
		Lars-Phillip Spiegel was supported by grant BE 1700/12-1, and Alexander Hahn was supported by grant KE 1413/15-1.
%
%
%
%
%
%
%
%
%
%
%
%
	
	\printcredits
	
	\bibliographystyle{cas-model2-names}
	
	\newcommand{\verzeichnisBibtex}{\string~/BibTeXReferencesSVNlink} 
	
	
	
	\bibliography{%
		references
	}
	
\end{document}

%% file: macro_files_stripped.tex
\newcommand{\symbolForInfSkept}{\ensuremath{\mathit{sk}}}

\newcommand{\symbolInd}{\ensuremath{\mathit{\mathit{O}}}}

\newcommand{\kbableitModMinName}[3]{\ensuremath{\nmableit^{\!\!\!\!#2,#3}_{\!#1}}}

\newcommand{\indinf}[3]{\ensuremath{\kbableitModMinName{\R}{\symbolForInfSkept}{\symbolInd}}}

\newcommand{\BB}{belief base\xspace}   %
\newcommand{\BBs}{belief bases\xspace}   %
\newcommand{\NFallKonditionale}[1]{\ensuremath{\mathit{NFC}(#1)}}

\newcommand{\closureMatrixName}[1]{\ensuremath{\mathit{CM}}} %

\newcommand{\listL}[1]{\ensuremath{\lbrack\mathcal{L}_{\Sigma}\rbrack}} %
\newcommand{\listNFC}[1]{\ensuremath{\lbrack\NFallKonditionale{\Sigma}\rbrack}} %

\makeatletter
\newcommand{\preceqdotLOKAL}{\mathrel{\mathpalette\LOKALpr@ceqd@t\relax}}
\newcommand{\LOKALpr@ceqd@t}[2]{%
  \begingroup
  \sbox\z@{$#1\prec$}\sbox\tw@{$#1\preccurlyeq$}%
  \dimen@=0.5\dimexpr\ht\tw@-\ht\z@\relax
  {\preccurlyeq}%
  \mkern-4mu
  \raisebox{\dimen@}{$\m@th#1\cdot$}%
  \endgroup
}
\makeatother

\makeatletter
\newcommand{\preceqprecdotLOKAL}{\mathrel{\mathpalette\LOKALprpr@ceqd@t\relax}}
\newcommand{\LOKALprpr@ceqd@t}[2]{%
  \begingroup
  \sbox\z@{$#1\prec$}\sbox\tw@{$#1\preccurlyeq$}%
  \dimen@=0.5\dimexpr\ht\tw@-\ht\z@\relax
  {\preccurlyeq}%
  \mkern-9mu
  {\prec}%
  \mkern-4mu
  \raisebox{\dimen@}{$\m@th#1\cdot$}%
  \endgroup
}
\makeatother

\newcommand{\displayTransformationRule}[3]%
{#1 &  #2 & #3}

\newcommand{\displayTRrule}[3]%
{#1 &  #2 & #3}

\newcommand{\symbolSSVker}{\ensuremath{\textbf{R1}}\xspace}  %
\newcommand{\symbolSSFker}{\ensuremath{\textbf{R2}}\xspace}  %
\newcommand{\symbolELMker}{\ensuremath{\textbf{R3}}\xspace}  %
\newcommand{\symbolTRIVIAL}{\textrm{\textbf{R4}}\xspace}  %
\newcommand{\symbolSUBSETS}{\textrm{\textbf{R5}}\xspace}  %
\newcommand{\symbolCIRCLE}{\textrm{\textbf{R6}}\xspace}  %

\newcommand{\refTRIVIAL}{\ensuremath{\symbolTRIVIAL}\xspace}    %
\newcommand{\refSUBSETS}{\ensuremath{\symbolSUBSETS}\xspace}    %
\newcommand{\refCIRCLE}{\ensuremath{\symbolCIRCLE}\xspace}    %

\newcommand{\refSSVker}{\ensuremath{\symbolSSVker}\xspace}    %
\newcommand{\refSSFker}{\ensuremath{\symbolSSFker}\xspace}    %
\newcommand{\refELMker}{\ensuremath{\symbolELMker}\xspace}    %

\newcommand{\nameSSV}{\ensuremath{\textit{subset-V}}\xspace}  %
\newcommand{\nameSSF}{\ensuremath{\textit{subset-F}}\xspace}  %
\newcommand{\nameELM}{\ensuremath{\textit{element}}\xspace}  %
\newcommand{\nameTRIVIAL}{\ensuremath{\textit{trivial}}\xspace}  %
\newcommand{\nameSUBSETS}{\ensuremath{\textit{subsets}}\xspace}  %
\newcommand{\nameCIRCLE}{\ensuremath{\textit{circle}}\xspace}  %

\newcommand{\nsrPaarKER}[3]{\ensuremath{\langle #1,\; #2 \rangle_{#3}}}

{
     
    \begin{enumerate}
}
{
    \end{enumerate}
}

{

    \begin{enumerate}
}
{
    \end{enumerate}
}

\newcommand\satzCLab[4]%
{\ifthenelse{\equal{#1}{no}}{\ensuremath{\satzCL{\{#3\}}{\{#4\}}}}{\ensuremath{\hbox{\ensuremath{#1_{#2}\!\!:\!\,\satzCL{\{#3\}}{\{#4\}}}}}}}

\newcommand{\nfcab}[2][no]{%
    \IfEqCase{#2}{%
        {01}{\ensuremath{\satzCLab{#1}{#2}{3}{3,2}}}%
        {02}{\ensuremath{\satzCLab{#1}{#2}{2}{3,2}}}%
        {03}{\ensuremath{\satzCLab{#1}{#2}{3}{3,0}}}%
        {04}{\ensuremath{\satzCLab{#1}{#2}{0}{3,0}}}%
        {05}{\ensuremath{\satzCLab{#1}{#2}{2}{2,1}}}%
        {06}{\ensuremath{\satzCLab{#1}{#2}{2}{2,0}}}%
        {07}{\ensuremath{\satzCLab{#1}{#2}{0}{2,0}}}%
        {08}{\ensuremath{\satzCLab{#1}{#2}{3}{3,2,1}}}%
        {09}{\ensuremath{\satzCLab{#1}{#2}{2}{3,2,1}}}%
        {12}{\ensuremath{\cb{\satzCLab{#1}{#2}{3}{3,2,0}}}}%
        {13}{\ensuremath{\cb{\satzCLab{#1}{#2}{2}{3,2,0}}}}%
        {14}{\ensuremath{\cb{\satzCLab{#1}{#2}{0}{3,2,0}}}}%
        {18}{\ensuremath{\cb{\satzCLab{#1}{#2}{2}{2,1,0}}}}%
        {19}{\ensuremath{\cb{\satzCLab{#1}{#2}{0}{2,1,0}}}}%
        {10}{\ensuremath{\cb{\satzCLab{#1}{#2}{3,2}{3,2,1}}}}%
        {11}{\ensuremath{\cb{\satzCLab{#1}{#2}{2,1}{3,2,1}}}}%
        {15}{\ensuremath{\cb{\satzCLab{#1}{#2}{3,2}{3,2,0}}}}%
        {16}{\ensuremath{\cb{\satzCLab{#1}{#2}{3,0}{3,2,0}}}}%
        {17}{\ensuremath{\cb{\satzCLab{#1}{#2}{2,0}{3,2,0}}}}%
        {20}{\ensuremath{\satzCLab{#1}{#2}{2,1}{2,1,0}}}%
        {21}{\ensuremath{\satzCLab{#1}{#2}{2,0}{2,1,0}}}%
        {22}{\ensuremath{\satzCLab{#1}{#2}{3}{3,2,1,0}}}%
        {23}{\ensuremath{\satzCLab{#1}{#2}{2}{3,2,1,0}}}%
        {24}{\ensuremath{\satzCLab{#1}{#2}{0}{3,2,1,0}}}%
        {25}{\ensuremath{\satzCLab{#1}{#2}{3,2}{3,2,1,0}}}%
        {26}{\ensuremath{\satzCLab{#1}{#2}{3,0}{3,2,1,0}}}%
        {27}{\ensuremath{\satzCLab{#1}{#2}{2,1}{3,2,1,0}}}%
        {28}{\ensuremath{\satzCLab{#1}{#2}{2,0}{3,2,1,0}}}%
        {29}{\ensuremath{\satzCLab{#1}{#2}{3,2,1}{3,2,1,0}}}%
        {30}{\ensuremath{\satzCLab{#1}{#2}{3,2,0}{3,2,1,0}}}%
        {31}{\ensuremath{\satzCLab{#1}{#2}{2,1,0}{3,2,1,0}}}%
        {32}{\ensuremath{\satzCLab{#1}{#2}{3}{3,1}}}%
        {33}{\ensuremath{\satzCLab{#1}{#2}{1}{3,1}}}%
        {34}{\ensuremath{\satzCLab{#1}{#2}{1}{2,1}}}%
        {35}{\ensuremath{\satzCLab{#1}{#2}{1}{1,0}}}%
        {36}{\ensuremath{\satzCLab{#1}{#2}{0}{1,0}}}%
        {37}{\ensuremath{\satzCLab{#1}{#2}{1}{3,2,1}}}%
        {39}{\ensuremath{\cb{\satzCLab{#1}{#2}{3}{3,1,0}}}}%
        {40}{\ensuremath{\cb{\satzCLab{#1}{#2}{1}{3,1,0}}}}%
        {41}{\ensuremath{\cb{\satzCLab{#1}{#2}{0}{3,1,0}}}}%
        {45}{\ensuremath{\cb{\satzCLab{#1}{#2}{1}{2,1,0}}}}%
        {38}{\ensuremath{\cb{\satzCLab{#1}{#2}{3,1}{3,2,1}}}}%
        {42}{\ensuremath{\cb{\satzCLab{#1}{#2}{3,1}{3,1,0}}}}%
        {43}{\ensuremath{\cb{\satzCLab{#1}{#2}{3,0}{3,1,0}}}}%
        {44}{\ensuremath{\cb{\satzCLab{#1}{#2}{1,0}{3,1,0}}}}%
        {46}{\ensuremath{\satzCLab{#1}{#2}{1,0}{2,1,0}}}%
        {47}{\ensuremath{\satzCLab{#1}{#2}{1}{3,2,1,0}}}%
        {48}{\ensuremath{\satzCLab{#1}{#2}{3,1}{3,2,1,0}}}%
        {49}{\ensuremath{\satzCLab{#1}{#2}{1,0}{3,2,1,0}}}%
        {50}{\ensuremath{\satzCLab{#1}{#2}{3,1,0}{3,2,1,0}}}%
    }[\PackageError{nfcab}{Undefined option to nfcab: #1}{}]%
}%

\newcommand{\nfcabChar}[2][no]{%
    \IfEqCase{#2}{%
        {01}{\ensuremath{\satzCLab{#1}{#2}{ab}{ab,a\ol{b}}}}%
        {02}{\ensuremath{\satzCLab{#1}{#2}{a\ol{b}}{ab,a\ol{b}}}}%
        {03}{\ensuremath{\satzCLab{#1}{#2}{ab}{ab,\ol{a}\ol{b}}}}%
        {04}{\ensuremath{\satzCLab{#1}{#2}{\ol{a}\ol{b}}{ab,\ol{a}\ol{b}}}}%
        {05}{\ensuremath{\satzCLab{#1}{#2}{a\ol{b}}{a\ol{b},\ol{a}b}}}%
        {06}{\ensuremath{\satzCLab{#1}{#2}{a\ol{b}}{a\ol{b},\ol{a}\ol{b}}}}%
        {07}{\ensuremath{\satzCLab{#1}{#2}{\ol{a}\ol{b}}{a\ol{b},\ol{a}\ol{b}}}}%
        {08}{\ensuremath{\satzCLab{#1}{#2}{ab}{ab,a\ol{b},\ol{a}b}}}%
        {09}{\ensuremath{\satzCLab{#1}{#2}{a\ol{b}}{ab,a\ol{b},\ol{a}b}}}%
        {12}{\ensuremath{\satzCLab{#1}{#2}{ab}{ab,a\ol{b},\ol{a}\ol{b}}}}%
        {13}{\ensuremath{\satzCLab{#1}{#2}{a\ol{b}}{ab,a\ol{b},\ol{a}\ol{b}}}}%
        {14}{\ensuremath{\satzCLab{#1}{#2}{\ol{a}\ol{b}}{ab,a\ol{b},\ol{a}\ol{b}}}}%
        {18}{\ensuremath{\satzCLab{#1}{#2}{a\ol{b}}{a\ol{b},\ol{a}b,\ol{a}\ol{b}}}}%
        {19}{\ensuremath{\satzCLab{#1}{#2}{\ol{a}\ol{b}}{a\ol{b},\ol{a}b,\ol{a}\ol{b}}}}%
        {10}{\ensuremath{\satzCLab{#1}{#2}{ab,a\ol{b}}{ab,a\ol{b},\ol{a}b}}}%
        {11}{\ensuremath{\satzCLab{#1}{#2}{a\ol{b},\ol{a}b}{ab,a\ol{b},\ol{a}b}}}%
        {15}{\ensuremath{\satzCLab{#1}{#2}{ab,a\ol{b}}{ab,a\ol{b},\ol{a}\ol{b}}}}%
        {16}{\ensuremath{\satzCLab{#1}{#2}{ab,\ol{a}\ol{b}}{ab,a\ol{b},\ol{a}\ol{b}}}}%
        {17}{\ensuremath{\satzCLab{#1}{#2}{a\ol{b},\ol{a}\ol{b}}{ab,a\ol{b},\ol{a}\ol{b}}}}%
        {20}{\ensuremath{\satzCLab{#1}{#2}{a\ol{b},\ol{a}b}{a\ol{b},\ol{a}b,\ol{a}\ol{b}}}}%
        {21}{\ensuremath{\satzCLab{#1}{#2}{a\ol{b},\ol{a}\ol{b}}{a\ol{b},\ol{a}b,\ol{a}\ol{b}}}}%
        {22}{\ensuremath{\satzCLab{#1}{#2}{ab}{ab,a\ol{b},\ol{a}b,\ol{a}\ol{b}}}}%
        {23}{\ensuremath{\satzCLab{#1}{#2}{a\ol{b}}{ab,a\ol{b},\ol{a}b,\ol{a}\ol{b}}}}%
        {24}{\ensuremath{\satzCLab{#1}{#2}{\ol{a}\ol{b}}{ab,a\ol{b},\ol{a}b,\ol{a}\ol{b}}}}%
        {25}{\ensuremath{\satzCLab{#1}{#2}{ab,a\ol{b}}{ab,a\ol{b},\ol{a}b,\ol{a}\ol{b}}}}%
        {26}{\ensuremath{\satzCLab{#1}{#2}{ab,\ol{a}\ol{b}}{ab,a\ol{b},\ol{a}b,\ol{a}\ol{b}}}}%
        {27}{\ensuremath{\satzCLab{#1}{#2}{a\ol{b},\ol{a}b}{ab,a\ol{b},\ol{a}b,\ol{a}\ol{b}}}}%
        {28}{\ensuremath{\satzCLab{#1}{#2}{a\ol{b},\ol{a}\ol{b}}{ab,a\ol{b},\ol{a}b,\ol{a}\ol{b}}}}%
        {29}{\ensuremath{\satzCLab{#1}{#2}{ab,a\ol{b},\ol{a}b}{ab,a\ol{b},\ol{a}b,\ol{a}\ol{b}}}}%
        {30}{\ensuremath{\satzCLab{#1}{#2}{ab,a\ol{b},\ol{a}\ol{b}}{ab,a\ol{b},\ol{a}b,\ol{a}\ol{b}}}}%
        {31}{\ensuremath{\satzCLab{#1}{#2}{a\ol{b},\ol{a}b,\ol{a}\ol{b}}{ab,a\ol{b},\ol{a}b,\ol{a}\ol{b}}}}%
        {32}{\ensuremath{\satzCLab{#1}{#2}{ab}{ab,\ol{a}b}}}%
        {33}{\ensuremath{\satzCLab{#1}{#2}{\ol{a}b}{ab,\ol{a}b}}}%
        {34}{\ensuremath{\satzCLab{#1}{#2}{\ol{a}b}{a\ol{b},\ol{a}b}}}%
        {35}{\ensuremath{\satzCLab{#1}{#2}{\ol{a}b}{\ol{a}b,\ol{a}\ol{b}}}}%
        {36}{\ensuremath{\satzCLab{#1}{#2}{\ol{a}\ol{b}}{\ol{a}b,\ol{a}\ol{b}}}}%
        {37}{\ensuremath{\satzCLab{#1}{#2}{\ol{a}b}{ab,a\ol{b},\ol{a}b}}}%
        {39}{\ensuremath{\satzCLab{#1}{#2}{ab}{ab,\ol{a}b,\ol{a}\ol{b}}}}%
        {40}{\ensuremath{\satzCLab{#1}{#2}{\ol{a}b}{ab,\ol{a}b,\ol{a}\ol{b}}}}%
        {41}{\ensuremath{\satzCLab{#1}{#2}{\ol{a}\ol{b}}{ab,\ol{a}b,\ol{a}\ol{b}}}}%
        {45}{\ensuremath{\satzCLab{#1}{#2}{\ol{a}b}{a\ol{b},\ol{a}b,\ol{a}\ol{b}}}}%
        {38}{\ensuremath{\satzCLab{#1}{#2}{ab,\ol{a}b}{ab,a\ol{b},\ol{a}b}}}%
        {42}{\ensuremath{\satzCLab{#1}{#2}{ab,\ol{a}b}{ab,\ol{a}b,\ol{a}\ol{b}}}}%
        {43}{\ensuremath{\satzCLab{#1}{#2}{ab,\ol{a}\ol{b}}{ab,\ol{a}b,\ol{a}\ol{b}}}}%
        {44}{\ensuremath{\satzCLab{#1}{#2}{\ol{a}b,\ol{a}\ol{b}}{ab,\ol{a}b,\ol{a}\ol{b}}}}%
        {46}{\ensuremath{\satzCLab{#1}{#2}{\ol{a}b,\ol{a}\ol{b}}{a\ol{b},\ol{a}b,\ol{a}\ol{b}}}}%
        {47}{\ensuremath{\satzCLab{#1}{#2}{\ol{a}b}{ab,a\ol{b},\ol{a}b,\ol{a}\ol{b}}}}%
        {48}{\ensuremath{\satzCLab{#1}{#2}{ab,\ol{a}b}{ab,a\ol{b},\ol{a}b,\ol{a}\ol{b}}}}%
        {49}{\ensuremath{\satzCLab{#1}{#2}{\ol{a}b,\ol{a}\ol{b}}{ab,a\ol{b},\ol{a}b,\ol{a}\ol{b}}}}%
        {50}{\ensuremath{\satzCLab{#1}{#2}{ab,\ol{a}b,\ol{a}\ol{b}}{ab,a\ol{b},\ol{a}b,\ol{a}\ol{b}}}}%
    }[\PackageError{nfcab}{Undefined option to nfcab: #1}{}]%
}%

\newcommand{\nfcabNeu}[2][no]{%
    \IfEqCase{#2}{%
        {01}{\ensuremath{\satzCLab{#1}{#2}{3}{3,2}}}%
        {02}{\ensuremath{\satzCLab{#1}{#2}{3}{3,1}}}%
        {03}{\ensuremath{\satzCLab{#1}{#2}{2}{3,2}}}%
        {04}{\ensuremath{\satzCLab{#1}{#2}{1}{3,1}}}%
        {05}{\ensuremath{\satzCLab{#1}{#2}{3}{3,0}}}%
        {06}{\ensuremath{\satzCLab{#1}{#2}{0}{3,0}}}%
        {07}{\ensuremath{\satzCLab{#1}{#2}{2}{2,1}}}%
        {08}{\ensuremath{\satzCLab{#1}{#2}{1}{2,1}}}%
        {09}{\ensuremath{\satzCLab{#1}{#2}{2}{2,0}}}%
        {10}{\ensuremath{\satzCLab{#1}{#2}{1}{1,0}}}%
        {11}{\ensuremath{\satzCLab{#1}{#2}{0}{2,0}}}%
        {12}{\ensuremath{\satzCLab{#1}{#2}{0}{1,0}}}%
        {13}{\ensuremath{\satzCLab{#1}{#2}{3}{3,2,1}}}%
        {14}{\ensuremath{\satzCLab{#1}{#2}{2}{3,2,1}}}%
        {15}{\ensuremath{\satzCLab{#1}{#2}{1}{3,2,1}}}%
        {16}{\ensuremath{\satzCLab{#1}{#2}{3,2}{3,2,1}}}%
        {17}{\ensuremath{\satzCLab{#1}{#2}{3,1}{3,2,1}}}%
        {18}{\ensuremath{\satzCLab{#1}{#2}{2,1}{3,2,1}}}%
        {19}{\ensuremath{\satzCLab{#1}{#2}{3}{3,2,0}}}%
        {20}{\ensuremath{\satzCLab{#1}{#2}{3}{3,1,0}}}%
        {21}{\ensuremath{\satzCLab{#1}{#2}{2}{3,2,0}}}%
        {22}{\ensuremath{\satzCLab{#1}{#2}{1}{3,1,0}}}%
        {23}{\ensuremath{\satzCLab{#1}{#2}{0}{3,2,0}}}%
        {24}{\ensuremath{\satzCLab{#1}{#2}{0}{3,1,0}}}%
        {25}{\ensuremath{\satzCLab{#1}{#2}{3,2}{3,2,0}}}%
        {26}{\ensuremath{\satzCLab{#1}{#2}{3,1}{3,1,0}}}%
        {27}{\ensuremath{\satzCLab{#1}{#2}{3,0}{3,2,0}}}%
        {28}{\ensuremath{\satzCLab{#1}{#2}{3,0}{3,1,0}}}%
        {29}{\ensuremath{\satzCLab{#1}{#2}{2,0}{3,2,0}}}%
        {30}{\ensuremath{\satzCLab{#1}{#2}{1,0}{3,1,0}}}%
        {31}{\ensuremath{\satzCLab{#1}{#2}{2}{2,1,0}}}%
        {32}{\ensuremath{\satzCLab{#1}{#2}{1}{2,1,0}}}%
        {33}{\ensuremath{\satzCLab{#1}{#2}{0}{2,1,0}}}%
        {34}{\ensuremath{\satzCLab{#1}{#2}{2,1}{2,1,0}}}%
        {35}{\ensuremath{\satzCLab{#1}{#2}{2,0}{2,1,0}}}%
        {36}{\ensuremath{\satzCLab{#1}{#2}{1,0}{2,1,0}}}%
        {37}{\ensuremath{\satzCLab{#1}{#2}{3}{3,2,1,0}}}%
        {38}{\ensuremath{\satzCLab{#1}{#2}{2}{3,2,1,0}}}%
        {39}{\ensuremath{\satzCLab{#1}{#2}{1}{3,2,1,0}}}%
        {40}{\ensuremath{\satzCLab{#1}{#2}{0}{3,2,1,0}}}%
        {41}{\ensuremath{\satzCLab{#1}{#2}{3,2}{3,2,1,0}}}%
        {42}{\ensuremath{\satzCLab{#1}{#2}{3,1}{3,2,1,0}}}%
        {43}{\ensuremath{\satzCLab{#1}{#2}{3,0}{3,2,1,0}}}%
        {44}{\ensuremath{\satzCLab{#1}{#2}{2,1}{3,2,1,0}}}%
        {45}{\ensuremath{\satzCLab{#1}{#2}{2,0}{3,2,1,0}}}%
        {46}{\ensuremath{\satzCLab{#1}{#2}{1,0}{3,2,1,0}}}%
        {47}{\ensuremath{\satzCLab{#1}{#2}{3,2,1}{3,2,1,0}}}%
        {48}{\ensuremath{\satzCLab{#1}{#2}{3,2,0}{3,2,1,0}}}%
        {49}{\ensuremath{\satzCLab{#1}{#2}{3,1,0}{3,2,1,0}}}%
        {50}{\ensuremath{\satzCLab{#1}{#2}{2,1,0}{3,2,1,0}}}%
    }[\PackageError{nfcab}{Undefined option to nfcab: #1}{}]%
}%

\newcommand{\nfcabCharNeu}[2][no]{%
    \IfEqCase{#2}{%
        {01}{\ensuremath{\satzCLab{#1}{#2}{ab}{ab,a\ol{b}}}}%
        {02}{\ensuremath{\satzCLab{#1}{#2}{ab}{ab,\ol{a}b}}}%
        {03}{\ensuremath{\satzCLab{#1}{#2}{a\ol{b}}{ab,a\ol{b}}}}%
        {04}{\ensuremath{\satzCLab{#1}{#2}{\ol{a}b}{ab,\ol{a}b}}}%
        {05}{\ensuremath{\satzCLab{#1}{#2}{ab}{ab,\ol{a}\ol{b}}}}%
        {06}{\ensuremath{\satzCLab{#1}{#2}{\ol{a}\ol{b}}{ab,\ol{a}\ol{b}}}}%
        {07}{\ensuremath{\satzCLab{#1}{#2}{a\ol{b}}{a\ol{b},\ol{a}b}}}%
        {08}{\ensuremath{\satzCLab{#1}{#2}{\ol{a}b}{a\ol{b},\ol{a}b}}}%
        {09}{\ensuremath{\satzCLab{#1}{#2}{a\ol{b}}{a\ol{b},\ol{a}\ol{b}}}}%
        {10}{\ensuremath{\satzCLab{#1}{#2}{\ol{a}b}{\ol{a}b,\ol{a}\ol{b}}}}%
        {11}{\ensuremath{\satzCLab{#1}{#2}{\ol{a}\ol{b}}{a\ol{b},\ol{a}\ol{b}}}}%
        {12}{\ensuremath{\satzCLab{#1}{#2}{\ol{a}\ol{b}}{\ol{a}b,\ol{a}\ol{b}}}}%
        {13}{\ensuremath{\satzCLab{#1}{#2}{ab}{ab,a\ol{b},\ol{a}b}}}%
        {14}{\ensuremath{\satzCLab{#1}{#2}{a\ol{b}}{ab,a\ol{b},\ol{a}b}}}%
        {15}{\ensuremath{\satzCLab{#1}{#2}{\ol{a}b}{ab,a\ol{b},\ol{a}b}}}%
        {16}{\ensuremath{\satzCLab{#1}{#2}{ab,a\ol{b}}{ab,a\ol{b},\ol{a}b}}}%
        {17}{\ensuremath{\satzCLab{#1}{#2}{ab,\ol{a}b}{ab,a\ol{b},\ol{a}b}}}%
        {18}{\ensuremath{\satzCLab{#1}{#2}{a\ol{b},\ol{a}b}{ab,a\ol{b},\ol{a}b}}}%
        {19}{\ensuremath{\satzCLab{#1}{#2}{ab}{ab,a\ol{b},\ol{a}\ol{b}}}}%
        {20}{\ensuremath{\satzCLab{#1}{#2}{ab}{ab,\ol{a}b,\ol{a}\ol{b}}}}%
        {21}{\ensuremath{\satzCLab{#1}{#2}{a\ol{b}}{ab,a\ol{b},\ol{a}\ol{b}}}}%
        {22}{\ensuremath{\satzCLab{#1}{#2}{\ol{a}b}{ab,\ol{a}b,\ol{a}\ol{b}}}}%
        {23}{\ensuremath{\satzCLab{#1}{#2}{\ol{a}\ol{b}}{ab,a\ol{b},\ol{a}\ol{b}}}}%
        {24}{\ensuremath{\satzCLab{#1}{#2}{\ol{a}\ol{b}}{ab,\ol{a}b,\ol{a}\ol{b}}}}%
        {25}{\ensuremath{\satzCLab{#1}{#2}{ab,a\ol{b}}{ab,a\ol{b},\ol{a}\ol{b}}}}%
        {26}{\ensuremath{\satzCLab{#1}{#2}{ab,\ol{a}b}{ab,\ol{a}b,\ol{a}\ol{b}}}}%
        {27}{\ensuremath{\satzCLab{#1}{#2}{ab,\ol{a}\ol{b}}{ab,a\ol{b},\ol{a}\ol{b}}}}%
        {28}{\ensuremath{\satzCLab{#1}{#2}{ab,\ol{a}\ol{b}}{ab,\ol{a}b,\ol{a}\ol{b}}}}%
        {29}{\ensuremath{\satzCLab{#1}{#2}{a\ol{b},\ol{a}\ol{b}}{ab,a\ol{b},\ol{a}\ol{b}}}}%
        {30}{\ensuremath{\satzCLab{#1}{#2}{\ol{a}b,\ol{a}\ol{b}}{ab,\ol{a}b,\ol{a}\ol{b}}}}%
        {31}{\ensuremath{\satzCLab{#1}{#2}{a\ol{b}}{a\ol{b},\ol{a}b,\ol{a}\ol{b}}}}%
        {32}{\ensuremath{\satzCLab{#1}{#2}{\ol{a}b}{a\ol{b},\ol{a}b,\ol{a}\ol{b}}}}%
        {33}{\ensuremath{\satzCLab{#1}{#2}{\ol{a}\ol{b}}{a\ol{b},\ol{a}b,\ol{a}\ol{b}}}}%
        {34}{\ensuremath{\satzCLab{#1}{#2}{a\ol{b},\ol{a}b}{a\ol{b},\ol{a}b,\ol{a}\ol{b}}}}%
        {35}{\ensuremath{\satzCLab{#1}{#2}{a\ol{b},\ol{a}\ol{b}}{a\ol{b},\ol{a}b,\ol{a}\ol{b}}}}%
        {36}{\ensuremath{\satzCLab{#1}{#2}{\ol{a}b,\ol{a}\ol{b}}{a\ol{b},\ol{a}b,\ol{a}\ol{b}}}}%
        {37}{\ensuremath{\satzCLab{#1}{#2}{ab}{ab,a\ol{b},\ol{a}b,\ol{a}\ol{b}}}}%
        {38}{\ensuremath{\satzCLab{#1}{#2}{a\ol{b}}{ab,a\ol{b},\ol{a}b,\ol{a}\ol{b}}}}%
        {39}{\ensuremath{\satzCLab{#1}{#2}{\ol{a}b}{ab,a\ol{b},\ol{a}b,\ol{a}\ol{b}}}}%
        {40}{\ensuremath{\satzCLab{#1}{#2}{\ol{a}\ol{b}}{ab,a\ol{b},\ol{a}b,\ol{a}\ol{b}}}}%
        {41}{\ensuremath{\satzCLab{#1}{#2}{ab,a\ol{b}}{ab,a\ol{b},\ol{a}b,\ol{a}\ol{b}}}}%
        {42}{\ensuremath{\satzCLab{#1}{#2}{ab,\ol{a}b}{ab,a\ol{b},\ol{a}b,\ol{a}\ol{b}}}}%
        {43}{\ensuremath{\satzCLab{#1}{#2}{ab,\ol{a}\ol{b}}{ab,a\ol{b},\ol{a}b,\ol{a}\ol{b}}}}%
        {44}{\ensuremath{\satzCLab{#1}{#2}{a\ol{b},\ol{a}b}{ab,a\ol{b},\ol{a}b,\ol{a}\ol{b}}}}%
        {45}{\ensuremath{\satzCLab{#1}{#2}{a\ol{b},\ol{a}\ol{b}}{ab,a\ol{b},\ol{a}b,\ol{a}\ol{b}}}}%
        {46}{\ensuremath{\satzCLab{#1}{#2}{\ol{a}b,\ol{a}\ol{b}}{ab,a\ol{b},\ol{a}b,\ol{a}\ol{b}}}}%
        {47}{\ensuremath{\satzCLab{#1}{#2}{ab,a\ol{b},\ol{a}b}{ab,a\ol{b},\ol{a}b,\ol{a}\ol{b}}}}%
        {48}{\ensuremath{\satzCLab{#1}{#2}{ab,a\ol{b},\ol{a}\ol{b}}{ab,a\ol{b},\ol{a}b,\ol{a}\ol{b}}}}%
        {49}{\ensuremath{\satzCLab{#1}{#2}{ab,\ol{a}b,\ol{a}\ol{b}}{ab,a\ol{b},\ol{a}b,\ol{a}\ol{b}}}}%
        {50}{\ensuremath{\satzCLab{#1}{#2}{a\ol{b},\ol{a}b,\ol{a}\ol{b}}{ab,a\ol{b},\ol{a}b,\ol{a}\ol{b}}}}%
    }[\PackageError{nfcab}{Undefined option to nfcab: #1}{}]%
}%

\newcommand{\paarKlNr}[2]{\ensuremath{#1.#2}}

\newcommand{\nfcabNeuKlNr}[2][no]{%
  \IfEqCase{#2}{%
        {01}{\ensuremath{\satzCLab{#1}{\paarKlNr{01}{1}}{3}{3,2}}}%
        {02}{\ensuremath{\satzCLab{#1}{\paarKlNr{01}{2}}{3}{3,1}}}%
        {03}{\ensuremath{\satzCLab{#1}{\paarKlNr{02}{1}}{2}{3,2}}}%
        {04}{\ensuremath{\satzCLab{#1}{\paarKlNr{02}{2}}{1}{3,1}}}%
        {05}{\ensuremath{\satzCLab{#1}{\paarKlNr{03}{1}}{3}{3,0}}}%
        {06}{\ensuremath{\satzCLab{#1}{\paarKlNr{04}{1}}{0}{3,0}}}%
        {07}{\ensuremath{\satzCLab{#1}{\paarKlNr{05}{1}}{2}{2,1}}}%
        {08}{\ensuremath{\satzCLab{#1}{\paarKlNr{05}{2}}{1}{2,1}}}%
        {09}{\ensuremath{\satzCLab{#1}{\paarKlNr{06}{1}}{2}{2,0}}}%
        {10}{\ensuremath{\satzCLab{#1}{\paarKlNr{06}{2}}{1}{1,0}}}%
        {11}{\ensuremath{\satzCLab{#1}{\paarKlNr{07}{1}}{0}{2,0}}}%
        {12}{\ensuremath{\satzCLab{#1}{\paarKlNr{07}{2}}{0}{1,0}}}%
        {13}{\ensuremath{\satzCLab{#1}{\paarKlNr{08}{1}}{3}{3,2,1}}}%
        {14}{\ensuremath{\satzCLab{#1}{\paarKlNr{09}{1}}{2}{3,2,1}}}%
        {15}{\ensuremath{\satzCLab{#1}{\paarKlNr{09}{2}}{1}{3,2,1}}}%
        {16}{\ensuremath{\satzCLab{#1}{\paarKlNr{10}{1}}{3,2}{3,2,1}}}%
        {17}{\ensuremath{\satzCLab{#1}{\paarKlNr{10}{2}}{3,1}{3,2,1}}}%
        {18}{\ensuremath{\satzCLab{#1}{\paarKlNr{11}{1}}{2,1}{3,2,1}}}%
        {19}{\ensuremath{\satzCLab{#1}{\paarKlNr{12}{1}}{3}{3,2,0}}}%
        {20}{\ensuremath{\satzCLab{#1}{\paarKlNr{12}{2}}{3}{3,1,0}}}%
        {21}{\ensuremath{\satzCLab{#1}{\paarKlNr{13}{1}}{2}{3,2,0}}}%
        {22}{\ensuremath{\satzCLab{#1}{\paarKlNr{13}{1}}{1}{3,1,0}}}%
        {23}{\ensuremath{\satzCLab{#1}{\paarKlNr{14}{1}}{0}{3,2,0}}}%
        {24}{\ensuremath{\satzCLab{#1}{\paarKlNr{14}{1}}{0}{3,1,0}}}%
        {25}{\ensuremath{\satzCLab{#1}{\paarKlNr{15}{1}}{3,2}{3,2,0}}}%
        {26}{\ensuremath{\satzCLab{#1}{\paarKlNr{15}{2}}{3,1}{3,1,0}}}%
        {27}{\ensuremath{\satzCLab{#1}{\paarKlNr{16}{1}}{3,0}{3,2,0}}}%
        {28}{\ensuremath{\satzCLab{#1}{\paarKlNr{16}{2}}{3,0}{3,1,0}}}%
        {29}{\ensuremath{\satzCLab{#1}{\paarKlNr{17}{1}}{2,0}{3,2,0}}}%
        {30}{\ensuremath{\satzCLab{#1}{\paarKlNr{17}{2}}{1,0}{3,1,0}}}%
        {31}{\ensuremath{\satzCLab{#1}{\paarKlNr{18}{1}}{2}{2,1,0}}}%
        {32}{\ensuremath{\satzCLab{#1}{\paarKlNr{18}{2}}{1}{2,1,0}}}%
        {33}{\ensuremath{\satzCLab{#1}{\paarKlNr{19}{1}}{0}{2,1,0}}}%
        {34}{\ensuremath{\satzCLab{#1}{\paarKlNr{20}{1}}{2,1}{2,1,0}}}%
        {35}{\ensuremath{\satzCLab{#1}{\paarKlNr{21}{1}}{2,0}{2,1,0}}}%
        {36}{\ensuremath{\satzCLab{#1}{\paarKlNr{21}{2}}{1,0}{2,1,0}}}%
        {37}{\ensuremath{\satzCLab{#1}{\paarKlNr{22}{1}}{3}{3,2,1,0}}}%
        {38}{\ensuremath{\satzCLab{#1}{\paarKlNr{23}{1}}{2}{3,2,1,0}}}%
        {39}{\ensuremath{\satzCLab{#1}{\paarKlNr{23}{2}}{1}{3,2,1,0}}}%
        {40}{\ensuremath{\satzCLab{#1}{\paarKlNr{24}{1}}{0}{3,2,1,0}}}%
        {41}{\ensuremath{\satzCLab{#1}{\paarKlNr{25}{1}}{3,2}{3,2,1,0}}}%
        {42}{\ensuremath{\satzCLab{#1}{\paarKlNr{25}{2}}{3,1}{3,2,1,0}}}%
        {43}{\ensuremath{\satzCLab{#1}{\paarKlNr{26}{1}}{3,0}{3,2,1,0}}}%
        {44}{\ensuremath{\satzCLab{#1}{\paarKlNr{27}{1}}{2,1}{3,2,1,0}}}%
        {45}{\ensuremath{\satzCLab{#1}{\paarKlNr{28}{1}}{2,0}{3,2,1,0}}}%
        {46}{\ensuremath{\satzCLab{#1}{\paarKlNr{28}{2}}{1,0}{3,2,1,0}}}%
        {47}{\ensuremath{\satzCLab{#1}{\paarKlNr{29}{1}}{3,2,1}{3,2,1,0}}}%
        {48}{\ensuremath{\satzCLab{#1}{\paarKlNr{30}{1}}{3,2,0}{3,2,1,0}}}%
        {49}{\ensuremath{\satzCLab{#1}{\paarKlNr{30}{2}}{3,1,0}{3,2,1,0}}}%
        {50}{\ensuremath{\satzCLab{#1}{\paarKlNr{31}{1}}{2,1,0}{3,2,1,0}}}%
    }[\PackageError{nfcab}{Undefined option to nfcab: #1}{}]%
}%

\newcommand{\id}{\ensuremath{\mathrm{id}}}

\newcommand{\beweisDirektDetailliert}[1]{}   %

\newcommand{\kuerzerFuerLICS}[1]{}           %

\newlength{\absetzenkleinerlaenge}
\newcount\stunde\newcount\minute
\stunde=\time 
\divide\stunde by 60 \minute=\time
\newcommand{\uhrzeit}{\makebox{\number\stunde\,:\,\ifnum\minute<10 %
0\fi\number\minute}}
\newcommand{\dotcup}{\mathbin{\dot\cup}}

\newcommand{\thistheoremname}{}
\newtheorem{generischespostulat}{\thistheoremname}   %

\newcommand{\solutionsRof}[1]{\ensuremath{\solutionsOf{\cspRof{#1}}}}

\newcommand{\hookdownarrow}{\mathrel{\rotatebox[origin=c]{-90}{$\hookrightarrow$}}}

\newcommand{\CondSynSplitGen}{\ensuremath{\CondSynSplitMacro{\Delta}{\Delta^1}{\Delta^2}{\Sigma_1}{\Sigma_2}{\Sigma_3}}}

\newcommand{\safeCondSynSplitGen}{\ensuremath{\safeCondSynSplitMacro{\Delta}{\Delta^1}{\Delta^2}{\Sigma_1}{\Sigma_2}{\Sigma_3}}}

\newcommand{\ol}[1]{\overline{#1}}

\renewcommand{\leq}{\leqslant}

\newcommand{\cspMIof}[2]{\ensuremath{\mathit{CR}^{#2}(#1)}} %
\newcommand{\solutionsR}{\ensuremath{\solutionsOfhelp{\cspR}}}
\newcommand{\solutionsOf}[1]{\ensuremath{\mathit{Sol}(#1)}}

\newcommand{\OCFsolutionsOf}[1]{\ensuremath{\mathit{Sol_{OCF}}(#1)}}

\newcommand{\OCFsolutionsRnMI}[1]{\ensuremath{\OCFsolutionsOf{\cspMIof{\R_n}{n-1}}}}

\newcommand{\induzierteOCF}[1]{\ensuremath{\kappa_{\!#1}}} %
\newcommand{\syntheticKB}[1]{\texttt{kb\_synth<\(n\)>\_c<\(2n\!\!-\!\!1\)>.pl}}

\newcommand{\cL}{\ensuremath{\mathcal L}}
\newcommand{\R}{\ensuremath{\mathcal R}}

\newcommand{\condAB}{(B | A)}

\newcommand{\notA}{\overline{A}}
\newcommand{\AnotB}{A \overline{B}}

\newcommand{\condL}{\mbox{$(\cL \mid \cL)$}}

\newcommand{\naturals}{\mathbb{N}}

\newcommand{\Mod}{\mbox{\it Mod}\,}

\newcommand{\kappaiminus}[1]{\eta_{#1}}

\newcommand{\kappaiminusVektor}{\ensuremath{\vv{\eta}}}

\makeatletter
\ifx\centernot\undefined
\newcommand*{\centernot}{%
	\mathpalette\@centernot
}
\fi
\def\@centernot#1#2{%
	\mathrel{%
		\rlap{%
			\settowidth\dimen@{$\m@th#1{#2}$}%
			\kern.5\dimen@
			\settowidth\dimen@{$\m@th#1=$}%
			\kern-.5\dimen@
			$\m@th#1\not$%
		}%
		{#2}%
	}%
}
\makeatother
\DeclareRobustCommand\nmableitSymb{\mathrel|\mkern-.5mu\joinrel\sim} %
\newcommand{\nmableit}{\ensuremath{\mbox{$\,\nmableitSymb\,$}}} %

\newcommand{\beweisendezeichen}%
{\penalty50\hspace*{0pt plus 1fil}\parfillskip=0pt\mbox{$\Box$}}

\newcommand{\fussnoteOhneMarkierung}[1]%
{%
\footnote{#1}%
\addtocounter{footnote}{-1}%
}

\newcommand{\satzCL}[2]{\ensuremath{(#1|#2)}}

\newlength{\abstand}
\newcommand{\signa}{\Sigma}
\newcommand{\subsigna}{\Theta}
\newcommand{\subsignabar}{\ol{\subsigna}}

\newcommand{\omegasub}{\omega^{\subsigna}}
\newcommand{\omegasubbar}{\omega^{\subsignabar}}

\newcommand{\omegasubj}[1]{\omega^{#1}}

\renewcommand{\ast}{*}

\newcommand{\marg}[2]{{#1|}_{#2}}
\newcommand{\margpw}[2]{{#1}^{#2}}

\makeatletter
\def\moverlay{\mathpalette\mov@rlay}
\def\mov@rlay#1#2{\leavevmode\vtop{%
		\baselineskip\z@skip \lineskiplimit-\maxdimen
		\ialign{\hfil$\m@th#1##$\hfil\cr#2\crcr}}}
\newcommand{\charfusion}[3][\mathord]{%
	#1{\ifx#1\mathop\vphantom{#2}\fi
		\mathpalette\mov@rlay{#2\cr#3}%
	}%
	\ifx#1\mathop\expandafter\displaylimits\fi}
\makeatother

\newcommand{\perpendicular}%
{\begin{sideways}$\models$\end{sideways}}

\newlength{\spalteAbst}
\newlength{\spalteAbstGr}
\newlength{\spalteAbstGGr}
\newlength{\zeileAbst}
\newcommand{\ALvariableE}{\ensuremath{a}}
\newcommand{\ALvariableZ}{\ensuremath{b}}
\newcommand{\ALvariableD}{\ensuremath{c}}
\newcommand{\ALvariableV}{\ensuremath{d}}
\newcommand{\ALvariableF}{\ensuremath{e}}

\newlength{\AbstandZwischenLiteralen}
\newcommand{\setzeAbstandZwischenLiteralen}{\hspace{\AbstandZwischenLiteralen}}

\newcommand{\omegaFuenf}[1]{%
    \IfEqCase{#1}{%
        {01}{\ensuremath{\omegaAafuenf{\ALvariableE}{\ALvariableZ}{\ALvariableD}{\ALvariableV}{\ALvariableF}}}%
        {02}{\ensuremath{\omegaAbfuenf{\ALvariableE}{\ALvariableZ}{\ALvariableD}{\ALvariableV}{\ALvariableF}}}%
        {03}{\ensuremath{\omegaBafuenf{\ALvariableE}{\ALvariableZ}{\ALvariableD}{\ALvariableV}{\ALvariableF}}}%
        {04}{\ensuremath{\omegaBbfuenf{\ALvariableE}{\ALvariableZ}{\ALvariableD}{\ALvariableV}{\ALvariableF}}}%
        {05}{\ensuremath{\omegaCafuenf{\ALvariableE}{\ALvariableZ}{\ALvariableD}{\ALvariableV}{\ALvariableF}}}%
        {06}{\ensuremath{\omegaCbfuenf{\ALvariableE}{\ALvariableZ}{\ALvariableD}{\ALvariableV}{\ALvariableF}}}%
        {07}{\ensuremath{\omegaDafuenf{\ALvariableE}{\ALvariableZ}{\ALvariableD}{\ALvariableV}{\ALvariableF}}}%
        {08}{\ensuremath{\omegaDbfuenf{\ALvariableE}{\ALvariableZ}{\ALvariableD}{\ALvariableV}{\ALvariableF}}}%
        {09}{\ensuremath{\omegaEafuenf{\ALvariableE}{\ALvariableZ}{\ALvariableD}{\ALvariableV}{\ALvariableF}}}%
        {10}{\ensuremath{\omegaEbfuenf{\ALvariableE}{\ALvariableZ}{\ALvariableD}{\ALvariableV}{\ALvariableF}}}%
        {11}{\ensuremath{\omegaFafuenf{\ALvariableE}{\ALvariableZ}{\ALvariableD}{\ALvariableV}{\ALvariableF}}}%
        {12}{\ensuremath{\omegaFbfuenf{\ALvariableE}{\ALvariableZ}{\ALvariableD}{\ALvariableV}{\ALvariableF}}}%
        {13}{\ensuremath{\omegaGafuenf{\ALvariableE}{\ALvariableZ}{\ALvariableD}{\ALvariableV}{\ALvariableF}}}%
        {14}{\ensuremath{\omegaGbfuenf{\ALvariableE}{\ALvariableZ}{\ALvariableD}{\ALvariableV}{\ALvariableF}}}%
        {15}{\ensuremath{\omegaHafuenf{\ALvariableE}{\ALvariableZ}{\ALvariableD}{\ALvariableV}{\ALvariableF}}}%
        {16}{\ensuremath{\omegaHbfuenf{\ALvariableE}{\ALvariableZ}{\ALvariableD}{\ALvariableV}{\ALvariableF}}}%
        {17}{\ensuremath{\omegaIafuenf{\ALvariableE}{\ALvariableZ}{\ALvariableD}{\ALvariableV}{\ALvariableF}}}%
        {18}{\ensuremath{\omegaIbfuenf{\ALvariableE}{\ALvariableZ}{\ALvariableD}{\ALvariableV}{\ALvariableF}}}%
        {19}{\ensuremath{\omegaJafuenf{\ALvariableE}{\ALvariableZ}{\ALvariableD}{\ALvariableV}{\ALvariableF}}}%
        {20}{\ensuremath{\omegaJbfuenf{\ALvariableE}{\ALvariableZ}{\ALvariableD}{\ALvariableV}{\ALvariableF}}}%
        {21}{\ensuremath{\omegaKafuenf{\ALvariableE}{\ALvariableZ}{\ALvariableD}{\ALvariableV}{\ALvariableF}}}%
        {22}{\ensuremath{\omegaKbfuenf{\ALvariableE}{\ALvariableZ}{\ALvariableD}{\ALvariableV}{\ALvariableF}}}%
        {23}{\ensuremath{\omegaLafuenf{\ALvariableE}{\ALvariableZ}{\ALvariableD}{\ALvariableV}{\ALvariableF}}}%
        {24}{\ensuremath{\omegaLbfuenf{\ALvariableE}{\ALvariableZ}{\ALvariableD}{\ALvariableV}{\ALvariableF}}}%
        {25}{\ensuremath{\omegaMafuenf{\ALvariableE}{\ALvariableZ}{\ALvariableD}{\ALvariableV}{\ALvariableF}}}%
        {26}{\ensuremath{\omegaMbfuenf{\ALvariableE}{\ALvariableZ}{\ALvariableD}{\ALvariableV}{\ALvariableF}}}%
        {27}{\ensuremath{\omegaNafuenf{\ALvariableE}{\ALvariableZ}{\ALvariableD}{\ALvariableV}{\ALvariableF}}}%
        {28}{\ensuremath{\omegaNbfuenf{\ALvariableE}{\ALvariableZ}{\ALvariableD}{\ALvariableV}{\ALvariableF}}}%
        {29}{\ensuremath{\omegaOafuenf{\ALvariableE}{\ALvariableZ}{\ALvariableD}{\ALvariableV}{\ALvariableF}}}%
        {30}{\ensuremath{\omegaObfuenf{\ALvariableE}{\ALvariableZ}{\ALvariableD}{\ALvariableV}{\ALvariableF}}}%
        {31}{\ensuremath{\omegaPafuenf{\ALvariableE}{\ALvariableZ}{\ALvariableD}{\ALvariableV}{\ALvariableF}}}%
        {32}{\ensuremath{\omegaPbfuenf{\ALvariableE}{\ALvariableZ}{\ALvariableD}{\ALvariableV}{\ALvariableF}}}%
    }[\PackageError{omegaFuenf}{Undefined option to omegaFuenf: #1}{}]%
}

\newcommand{\druckeWeltFuenf}[5]{\ensuremath{\mathit{#1\setzeAbstandZwischenLiteralen#2\setzeAbstandZwischenLiteralen#3\setzeAbstandZwischenLiteralen#4\setzeAbstandZwischenLiteralen#5}}} %

\newcommand{\omegaAafuenf}[5]{\ensuremath{\druckeWeltFuenf{#1}{#2}{#3}{#4}{#5}}}
\newcommand{\omegaAbfuenf}[5]{\ensuremath{\druckeWeltFuenf{#1}{#2}{#3}{#4}{\ol{#5}}}}
\newcommand{\omegaBafuenf}[5]{\ensuremath{\druckeWeltFuenf{#1}{#2}{#3}{\ol{#4}}{#5}}}
\newcommand{\omegaBbfuenf}[5]{\ensuremath{\druckeWeltFuenf{#1}{#2}{#3}{\ol{#4}}{\ol{#5}}}}
\newcommand{\omegaCafuenf}[5]{\ensuremath{\druckeWeltFuenf{#1}{#2}{\ol{#3}}{#4}{#5}}}
\newcommand{\omegaCbfuenf}[5]{\ensuremath{\druckeWeltFuenf{#1}{#2}{\ol{#3}}{#4}{\ol{#5}}}}
\newcommand{\omegaDafuenf}[5]{\ensuremath{\druckeWeltFuenf{#1}{#2}{\ol{#3}}{\ol{#4}}{#5}}}
\newcommand{\omegaDbfuenf}[5]{\ensuremath{\druckeWeltFuenf{#1}{#2}{\ol{#3}}{\ol{#4}}{\ol{#5}}}}

\newcommand{\omegaEafuenf}[5]{\ensuremath{\druckeWeltFuenf{#1}{\ol{#2}}{#3}{#4}{#5}}}
\newcommand{\omegaEbfuenf}[5]{\ensuremath{\druckeWeltFuenf{#1}{\ol{#2}}{#3}{#4}{\ol{#5}}}}
\newcommand{\omegaFafuenf}[5]{\ensuremath{\druckeWeltFuenf{#1}{\ol{#2}}{#3}{\ol{#4}}{#5}}}
\newcommand{\omegaFbfuenf}[5]{\ensuremath{\druckeWeltFuenf{#1}{\ol{#2}}{#3}{\ol{#4}}{\ol{#5}}}}
\newcommand{\omegaGafuenf}[5]{\ensuremath{\druckeWeltFuenf{#1}{\ol{#2}}{\ol{#3}}{#4}{#5}}}
\newcommand{\omegaGbfuenf}[5]{\ensuremath{\druckeWeltFuenf{#1}{\ol{#2}}{\ol{#3}}{#4}{\ol{#5}}}}
\newcommand{\omegaHafuenf}[5]{\ensuremath{\druckeWeltFuenf{#1}{\ol{#2}}{\ol{#3}}{\ol{#4}}{#5}}}
\newcommand{\omegaHbfuenf}[5]{\ensuremath{\druckeWeltFuenf{#1}{\ol{#2}}{\ol{#3}}{\ol{#4}}{\ol{#5}}}}

\newcommand{\omegaIafuenf}[5]{\ensuremath{\druckeWeltFuenf{\ol{#1}}{#2}{#3}{#4}{#5}}}
\newcommand{\omegaIbfuenf}[5]{\ensuremath{\druckeWeltFuenf{\ol{#1}}{#2}{#3}{#4}{\ol{#5}}}}
\newcommand{\omegaJafuenf}[5]{\ensuremath{\druckeWeltFuenf{\ol{#1}}{#2}{#3}{\ol{#4}}{#5}}}
\newcommand{\omegaJbfuenf}[5]{\ensuremath{\druckeWeltFuenf{\ol{#1}}{#2}{#3}{\ol{#4}}{\ol{#5}}}}
\newcommand{\omegaKafuenf}[5]{\ensuremath{\druckeWeltFuenf{\ol{#1}}{#2}{\ol{#3}}{#4}{#5}}}
\newcommand{\omegaKbfuenf}[5]{\ensuremath{\druckeWeltFuenf{\ol{#1}}{#2}{\ol{#3}}{#4}{\ol{#5}}}}
\newcommand{\omegaLafuenf}[5]{\ensuremath{\druckeWeltFuenf{\ol{#1}}{#2}{\ol{#3}}{\ol{#4}}{#5}}}
\newcommand{\omegaLbfuenf}[5]{\ensuremath{\druckeWeltFuenf{\ol{#1}}{#2}{\ol{#3}}{\ol{#4}}{\ol{#5}}}}

\newcommand{\omegaMafuenf}[5]{\ensuremath{\druckeWeltFuenf{\ol{#1}}{\ol{#2}}{#3}{#4}{#5}}}
\newcommand{\omegaMbfuenf}[5]{\ensuremath{\druckeWeltFuenf{\ol{#1}}{\ol{#2}}{#3}{#4}{\ol{#5}}}}
\newcommand{\omegaNafuenf}[5]{\ensuremath{\druckeWeltFuenf{\ol{#1}}{\ol{#2}}{#3}{\ol{#4}}{#5}}}
\newcommand{\omegaNbfuenf}[5]{\ensuremath{\druckeWeltFuenf{\ol{#1}}{\ol{#2}}{#3}{\ol{#4}}{\ol{#5}}}}
\newcommand{\omegaOafuenf}[5]{\ensuremath{\druckeWeltFuenf{\ol{#1}}{\ol{#2}}{\ol{#3}}{#4}{#5}}}
\newcommand{\omegaObfuenf}[5]{\ensuremath{\druckeWeltFuenf{\ol{#1}}{\ol{#2}}{\ol{#3}}{#4}{\ol{#5}}}}
\newcommand{\omegaPafuenf}[5]{\ensuremath{\druckeWeltFuenf{\ol{#1}}{\ol{#2}}{\ol{#3}}{\ol{#4}}{#5}}}
\newcommand{\omegaPbfuenf}[5]{\ensuremath{\druckeWeltFuenf{\ol{#1}}{\ol{#2}}{\ol{#3}}{\ol{#4}}{\ol{#5}}}}

\newcommand{\nullZero}{0}

\newcommand{\DeltaIWoJ}[2]{\Delta_{#1\setminus#2}}
\newcommand{\DeltaOneWoThree}{\DeltaIWoJ{1}{3}}
\newcommand{\DeltaTwoWoThree}{\DeltaIWoJ{2}{3}}
\newcommand{\DeltaIWoThree}{\DeltaIWoJ{i}{3}}
\newcommand{\DeltaJWoThree}{\DeltaIWoJ{i'}{3}}

\newcommand{\DeltaVar}[1]{\Delta_{#1}}
\newcommand{\DeltaI}{\DeltaVar{i}}
\newcommand{\DeltaJ}{\DeltaVar{i'}}
\newcommand{\DeltaOne}{\DeltaVar{1}}
\newcommand{\DeltaTwo}{\DeltaVar{2}}
\newcommand{\DeltaThree}{\DeltaVar{3}}
\newcommand{\DeltaZero}{\Delta_{\nullZero}}

\newcommand{\omegacomp}[2]{\omega^{#1}\omega^{#2}}
\newcommand{\omegacompThree}[3]{\omega^{#1}\omega^{#2}\omega^{#3}}

\newcommand{\omegaIThree}{\omegacomp{i}{3}}

\newcommand{\omegaIThreeJ}{\omegacompThree{i}{3}{i'}}

\newcommand{\omegaJ}{\omega^{i'}}

\newcommand{\genSafeCondSynSplitMacro}[6]{\ensuremath{#1 = #2 \bigcup^{\sf gs}_{#4,#5} #3 \mid #6}}
\newcommand{\genSafeCondSynSplitGen}{\genSafeCondSynSplitMacro{\Delta}{\DeltaOne}{\DeltaTwo}{\Sigma_1}{\Sigma_2}{\Sigma_3}}

\newcommand{\genSafe}{generalized safe}
\newcommand{\genSafety}{generalized safety}

\newcommand{\nsrV}{\ensuremath{\mathcal{V}}}
\newcommand{\nsrF}{\ensuremath{\mathcal{F}}}
\newcommand{\leereMenge}{\ensuremath{\varnothing}}

\renewcommand{\R}{\Delta}

\renewcommand{\marg}{|}

\renewcommand{\CondSynSplitGen}{\ensuremath{\Delta=\Delta_1\bigcup_{\Sigma_1,\Sigma_2} \Delta_2\mid \Sigma_3}}

\renewcommand{\safeCondSynSplitGen}{\ensuremath{\Delta=\Delta_1\bigcup_{\Sigma_1,\Sigma_2}^{\sf s} \Delta_2\mid \Sigma_3}}

\renewcommand{\solutionsOf}[1]{\mathit{Sol}(#1)}
\renewcommand{\solutionsRof}[1]{\solutionsOf{\CSP(#1)}}
\renewcommand{\solutionsR}{\solutionsRof{\Delta}}

\newcommand{\DeltaBird}{\Delta^{b}}

\newcommand{\wrt}{with respect to}

\newcommand{\CSP}{\ensuremath{\mathit{CSP}}\xspace}
\newcommand{\CSPs}{\ensuremath{\mathit{CSPs}}\xspace}

\renewcommand{\CSP}{\ensuremath{\mathit{C\!S\!P}}\xspace}
\renewcommand{\CSPs}{\ensuremath{\mathit{C\!S\!P\!s}}\xspace}

\newcommand{\spcr}{\psi}
\newcommand{\scpr}{\psi}

\newcommand{\natSpcr}{\Theta}

\newcommand{\directlyDependsSingle}{\leadsto^{dp}}

\newcommand{\directlyDependsSingleOfWrtSpcr}[1]{\leadsto^{dp}_{#1,\scpr}}

\newcommand{\directlyDepends}{\leadsto^{\ast}}

\newcommand{\directlyDependsOfWrt}[2]{\leadsto^{\ast}_{#1,#2}}
\newcommand{\directlyDependsOfWrtSpcr}[1]{\leadsto^{\ast}_{#1,\scpr}}

\newcommand{\unabhSymbol}{\ \raisebox{0.05em}{\rotatebox[origin=c]{90}{$\models$}}\ }

\renewcommand{\unabhSymbol}{\perp\!\!\!\perp}

\newcommand{\unabhSymbolOfWrt}[2]{\unabhSymbol_{#1}^{#2}}
\newcommand{\unabhSymbolOfWrtSpcr}[1]{\unabhSymbol_{#1}^{\spcr}}

\newcommand{\nutzeTextstyle}{}

\newcommand{\safeMid}{\mid^{\sf s}}

\newcommand{\constraintCupWrt}[1]{\mathchoice{{\textstyle \,\bigcup\limits^{\sf cs}_{#1}\,}}{\bigcup^{\sf cs}_{#1}}{}{}}

\newcommand{\notunabhSymbolOfWrt}[2]{\not\unabhSymbol_{#1}^{#2}}

\newcommand{\constraintCup}{\constraintCupWrt{\spcr}}
\newcommand{\constraintCupNatSpcr}{\constraintCupWrt{\natSpcr}}

\newcommand{\constraintSplitMacro}[3]{#1=#2\constraintCup\dots\constraintCup #3}
\newcommand{\constraintSplitGen}{\constraintSplitMacro{\Delta}{\Delta_1}{\Delta_m}}

\newcommand{\condConstraintSplitMacro}[4]{#1=#2\constraintCup\dots\constraintCup #3\mid #4}
\newcommand{\condConstraintSplitGen}{\condConstraintSplitMacro{\Delta}{\Delta_1}{\Delta_m}{\DeltaZero}}

\newcommand{\condConstraintSplitMacroTwoNatSpcr}[4]{#1=#2\constraintCupNatSpcr #3\mid #4}

\newcommand{\safeCondConstraintSplitMacro}[4]{#1=#2\constraintCup\dots\constraintCup #3\safeMid #4}
\newcommand{\safeCondConstraintSplitGen}{\safeCondConstraintSplitMacro{\Delta}{\Delta_1}{\Delta_m}{\DeltaZero}}

\newcommand{\OmegaOf}[1]{\Omega(#1)}
\newcommand{\calD}{\mathcal{D}}
\newcommand{\ver}{\ensuremath{\mathit{ver}}}
\newcommand{\fal}{\ensuremath{\mathit{fal}}}
\renewcommand{\fal}{\ensuremath{\mathit{f\!al}}}

\newcommand{\VSet}{\ensuremath{\mathcal{V}}}
\newcommand{\FSet}{\ensuremath{\mathcal{F}}}

\newcommand{\DeltaC}{\Delta^{c}}
\newcommand{\DeltaO}{\Delta^{a}}

\newcommand{\constraintOfWrtI}[3]{C_{#1}^{#2}(\delta_{#3})}
\newcommand{\constraintOfSpcrI}[2]{C_{#1}^{\spcr}(\delta_{#2})}
\newcommand{\constraintOfNatspcrI}[2]{C_{#1}^{\natSpcr}(\delta_{#2})}
\newcommand{\constraintOfNatspcrDI}[2]{C_{#1}^{\natSpcr}({#2})}

\newcommand*{\scopeOf}[3]{\mathit{scope}_{#1}^{#2}(#3)}

\newcommand{\solutionsRofNatSpcr}[1]{\solutionsRof{#1}}

\newcommand{\Aufteilung}{splt.}
\newcommand{\gcondCSolSplit}{safe conditional c-solution splitting}
\newcommand{\safecondCSolSplitAbr}{safe cond. c-solution \Aufteilung}

\newcommand{\cSolSplit}{c-solution splitting}
\newcommand{\cSolSplitAbr}{c-solution \Aufteilung}

\newcommand{\conditionalIndepOfInWrtGiven}[5]{#1\unabhSymbolOfWrt{#3}{#4} #2\mid #5}
\newcommand{\conditionalIndepGen}{\conditionalIndepOfInWrtGiven{\Delta_i}{\Delta_j}{\Delta}{\spcr}{\DeltaZero}}

\newcommand{\spcrSubset}{\theta}

\newcommand{\safeCondConstraintSplitTwoMacroWrt}[5]{{#1}={#2}\constraintCupWrt{#5}{#3}\safeMid{#4}}

\newcommand{\crFunc}{\hookdownarrow}
\newcommand{\crFuncWrt}[1]{\crFunc_{\!#1}}
\newcommand{\crFuncOfWrt}[2]{{#1\!\!\crFuncWrt{#2}}}

\newcommand{\crFuncGen}{\crFuncWrt{\Delta_1}}

\newcommand{\crFuncGenOf}[1]{\crFuncOfWrt{#1}{\Delta_1}}

\newcommand{\niceSafe}{fully localized}
\newcommand{\niceSafeAbbr}{fl}
\newcommand{\niceSafeMid}{\mid^{\sf \niceSafeAbbr}}

\newcommand{\niceSafeSplittAbr}{\niceSafeAbbr. cond. constraint splt.}

\newcommand{\niceSafeCondConstraintSplitMacroWrt}[5]{{#1}={#2}\constraintCupWrt{#5}\dots\constraintCupWrt{#5}{#3}\niceSafeMid{#4}}

\newcommand{\niceSafeCondConstraintSplitMacro}[4]{\niceSafeCondConstraintSplitMacroWrt{#1}{#2}{#3}{#4}{\spcr}}

\newcommand{\niceSafeCondConstraintSplitMacroTwoWrt}[5]{{#1}={#2}\constraintCupWrt{#5}{#3}\niceSafeMid{#4}}
\newcommand{\niceSafeCondConstraintSplitMacroTwo}[4]{\niceSafeCondConstraintSplitMacroTwoWrt{#1}{#2}{#3}{#4}{\spcr}}

\newcommand{\niceSafeCondConstraintSplitTwoMacroWrt}[5]{{#1}={#2}\constraintCupWrt{#5}{#3}\niceSafeMid{#4}}
\newcommand{\niceSafeCondConstraintSplitTwoWrt}[1]{\niceSafeCondConstraintSplitTwoMacroWrt{\Delta}{\Delta_1}{\Delta_2}{\Delta_0}{#1}}

\newcommand{\niceSafeCondConstraintSplitGen}{\niceSafeCondConstraintSplitMacroWrt{\Delta}{\Delta_1}{\Delta_m}{\Delta_0}{\spcr}}

\newcommand{\conditionalConstrainSplitGenWrt}[1]{\Delta=\Delta_1\constraintCupWrt{#1}\dots\constraintCupWrt{#1}\Delta_m\mid \Delta_0}

\newcommand{\projektion}{projection}
\newcommand{\projekting}{projecting}

\newcommand{\conditionalIndepGenNrWrt}[1]{\conditionalIndepOfInWrtGiven{\Delta_1}{\Delta_2}{\Delta}{#1}{\DeltaZero}}

\newcommand{\ie}{that is}
\newcommand{\eg}{for example}
\newcommand{\idFct}{\operatorname{id}}

%% file: textbody_stripped.tex
\section{Introduction}
An important field in the area of knowledge representation and reasoning is that of nonmonotonic reasoning. In this field, conditionals provide defeasible rules by encoding plausible, yet uncertain statements, whereby a conditional $(B|A)$ is read ``If $A$, then usually $B$''. Sets of conditionals, called \BBs, are a popular framework for inductive inference and have received much attention in the literature (\eg \cite{KrausLehmannMagidor1990short, Pearl1990systemZTARK, JH_Lehmann95, GoldszmidtPearl96, KernIsberner2004amai, KomoBeierle2022AMAI}). 
Semantics for and inference with conditionals are often notoriously difficult to compute, taking into account all possible worlds over a \BB, resulting in an exponential blowup in the signature size.
Thus, ways to localize computation of semantics and inference is of major importance in this area of research.

The syntax splitting principle \cite{KernIsbernerBeierleBrewka2020KR} provides a way to localize reasoning to only the relevant conditionals of a belief base.
The core idea of syntax splitting goes back to Parikh \cite{JH_Parikh99} 
where it is introduced in the context of belief revision, stating
that only syntactically relevant parts of a \BB\ should influence the revision result. 
More recently, syntax splitting was applied to inductive inference under the motto ``syntax splitting = relevance + independence'' \cite{KernIsbernerBeierleBrewka2020KR}.
Syntax splitting requires the 
belief base to split over disjoint subsignatures into disjoint subbases,
which is not very common in practice. Therefore, the notion of conditional syntax splitting was introduced \cite{HeyninckKernIsbernerMeyerHaldimannBeierle2023AAAI}, generalizing the notion of of syntax splitting and allowing also for an overlap of the signatures. The idea of 
focusing only on the relevant parts of a \BB
has also been investigated in other works, \eg,
\cite{Weydert98KR,PeppasWilliamsChopraFoo2015,BeierleHaldimannKernIsberner2021Boeger75Festschrift,WilhelmSezginKernIsbernerHaldimannBeierleHeyninck2023JELIA,dupin2024form,SpiegelHaldimannHeyninckKernIsbernerBeierle2025IJCAI}.

Splitting techniques provide a variety of benefits for inductive inference. They allow for localized reasoning, significantly simplifying inference tasks. Furthermore, they also uncover structural information about the \BB, making it easier to spot (in)dependencies and to understand the reasoning process.
In order to reason with \BBs, a semantics must be provided;
a popular one being ordinal conditional functions (OCFs), also called ranking functions \cite{Spohn88}.
In this article we investigate a special class of ranking functions called c-representations \cite{KernIsberner2001habil,KernIsberner2004amai}.
They can be specified via a constraint satisfaction problem (CSP) \cite{KernIsberner2001habil,BeierleEichhornKernIsbernerKutsch2018AMAI} and have been shown to comply with a variety of splitting properties \cite{KernIsbernerBeierleBrewka2020KR,BeierleHaldimannKernIsberner2021Boeger75Festschrift,WilhelmSezginKernIsbernerHaldimannBeierleHeyninck2023JELIA,BeierleSpiegelHaldimannWilhelmHeyninckKernIsberner2024KR,SpiegelHaldimannHeyninckKernIsbernerBeierle2025IJCAI}, in contrast to, \eg, reasoning according to system~P \cite{KrausLehmannMagidor1990short} or system~Z \cite{Pearl1990systemZTARK}.
In particular, splittings specifically for c-representations have been investigated in the form of constraint and solution splittings \cite{BeierleHaldimannKernIsberner2021Boeger75Festschrift,WilhelmSezginKernIsbernerHaldimannBeierleHeyninck2023JELIA}.
We extend this line of research by employing the idea underlying conditional syntax splitting and applying it to constraint and solution splittings for c-representations. We introduce the notions of conditional constraint and solution splittings for c-representations, allowing the subbases, which a \BB\ splits into, to also share conditionals, thus making localized computation of c-representations possible in many more cases.
We identify special subclasses of conditional constraint splittings, allowing us to solve the constraint system locally and we establish the relationship between the different splitting notions we introduce.
We 
present the interrelationships among our and other splitting techniques by establishing a splitting hierarchy, and
we also investigate connections of our work to safe covers \cite{WilhelmKernIsbernerAAAI24} which are motivated similarly through localized computation of c-representations.
In summary, the main contributions of this article are:
\begin{itemize}
\item We generalize constraint splittings to conditional constraint splittings, allowing subbases to depend on shared conditionals.
\item We identify the subclasses of safe and \niceSafe conditional constraint splittings, allowing us to solve the underlying constraint satisfaction problem locally on the subbases.
\item We generalize c-solution splittings to \gcondCSolSplit s, allowing subbases to also share conditionals.
\item We establish the precise relationships among safe conditional constraint splittings, \niceSafe\ conditional constraint splittings, and \gcondCSolSplit s.
\item We prove that every \niceSafe\ conditional constraint splitting yields a safe cover and vice versa.
\item We illustrate the relationship of our approaches to other splitting techniques by presenting a comprehensive hierarchy of splittings.
\end{itemize}

This article is a revised and largely extended version of our conference paper presented at FoIKS 2026 \cite{SpiegelBeierleKernIsbernerHahn2026FoIKS}. In particular, this article introduces \niceSafe\ splittings as a new subclass of conditional constraint splittings, strengthening several propositions in the process. Furthermore, this article provides an extensive analysis on the relationship between conditional constraint and conditional c-solution splittings, leading to several new results showing when conditional constraint splittings can be expected to yield conditional c-solution splittings. Moreover, this article contains all proofs not present in the conference paper and it provides additional illustrations and explanations to the introduced concepts.

The rest of this article is organized as follows. In Section 2, we recall required basics, and present conditional syntax splitting in Section 3.
In Section 4, we address c-representations and constraint splittings, and generalize the latter to conditional constraint splittings in Section 5.
In Section 6, we lift conditional splittings for c-representations to the semantical level and establish the relationship between conditional splittings on the constraint and on the solution level. In Section 7, we compare 
our approach to related work
before concluding in Section 8. 
\section{Formal Basics}
\label{sec_basics}
Let $\cL$ be a finitely generated propositional language over a signature $\signa$ with atoms $a,b,c, \ldots$ and
with formulas $A,B,C, \ldots$
We may write $AB$ instead of $A \wedge B$,
and overline formulas to indicate negation, \ie, $\notA$ means $\neg A$.
Let $\Omega$ denote the set of \emph{possible worlds} over $\cL$,
taken here simply
as the  set of all propositional interpretations over $\cL$. $\omega \models A$ means that
the propositional formula $A \in \cL$ holds in %
$\omega \in \Omega$; in this case $\omega$ is called a \emph{model} of $A$, and the set of all models of $A$ is denoted by $\Mod(A)$. For propositions $A,B \in \cL$,  $A \models B$ holds iff $\Mod(A) \subseteq \Mod(B)$, as usual. 
We will use $\omega$ both for the model and the corresponding complete conjunction of all positive or negated atoms, allowing us to use $\omega$ both as an interpretation and a proposition. %

For $\subsigna \subseteq \signa$,
let $\cL(\subsigna)$ or short \(\cL_{\subsigna}\) denote the propositional language defined by $\subsigna$, with associated set of interpretations $\Omega(\subsigna)$ or short \(\Omega_{\subsigna}\). Note that while each formula of $\cL(\subsigna)$ can also be considered as a formula of $\cL$, the interpretations $\omegasub \in \Omega(\subsigna)$ are not elements of $\Omega(\signa)$ if $\subsigna \neq \signa$. But each interpretation $\omega \in \Omega$ can be written uniquely in the form $\omega = \omegasub \omegasubbar$ with concatenated $\omegasub \in \Omega(\subsigna)$ and $\omegasubbar \in \Omega(\subsignabar)$, where $\subsignabar = \signa \backslash \subsigna$. 
The world $\omegasub$ is called the  \emph{reduct} of $\omega$ to $\subsigna$ \cite{Delgrande17}.
In the following, we will sometimes 
denote subsignatures of \(\Sigma\) by \(\Sigma_1, \Sigma_2, \dots\) and
write $\omegasubj{i}$ instead of $\omegasubj{\signa_i}$
to ease notation.
If $\Omega' \subseteq \Omega$ is a subset of models, then $\Omega'\marg_\subsigna = \{ \margpw{\omega}{\subsigna} \mid \omega \in \Omega'\} \subseteq \Omega(\subsigna)$ restricts $\Omega'$ to a subset of $\Omega(\subsigna)$.

By making use of a conditional operator $|$, we introduce the language 
\(
 (\cL | \cL) = \{(B|A) \mid A,B \in \cL \}
\)
of \emph{conditionals} over $\cL$.
Conditionals $(B|A)$  are meant to express  plausible, defeasible rules ``If $A$, then usually $B$''.
For a world \(\omega\) a conditional \(\satzCL{B}{A}\) is either \emph{verified} by \(\omega\) if \(\omega \models AB\), \emph{falsified} by \(\omega\) if \(\omega \models A\ol{B}\), or \emph{not applicable} to \(\omega\) if \(\omega \models \ol{A}\).
A conditional $(F|E)$ is called \emph{self-fulfilling}, if $E\models F$, \ie, there is no world that can falsify it.
Belief bases $\Delta=\{(B_1|A_1),\dots, (B_n|A_n)\}$ (over $\Sigma$) consist of finitely many conditionals from $\condL$.
For a conditional $\delta_i=(B_i|A_i)$ 
\begin{align*}
	\ver(\delta_i)=\{\omega\in\OmegaOf{\Sigma}\mid \omega\models A_iB_i\}\\
	\fal(\delta_i)=\{\omega\in\OmegaOf{\Sigma}\mid \omega\models A_i\ol{B_i}\}
\end{align*}
denote the sets of verifying respectively falsifying worlds.

A popular semantic framework %
for interpreting conditionals %
are
ordinal conditional functions (OCFs) $\kappa: \Omega \to \naturals \cup \{\infty\}$ with
$\kappa^{-1}(0) \neq \emptyset$. OCFs, also called \emph{ranking
    functions}, %
were introduced in a more general form, by Spohn \cite{Spohn88}. 
Intuitively, less plausible worlds are assigned higher numbers, where at least one world $\omega$ must be entirely plausible, \ie, $\kappa(\omega)=0$.
Formulas are assigned the rank of their most plausible models, \ie,
$\kappa(A) := \min\{\kappa(\omega) \mid \omega \models A\}$.
A conditional $\condAB$ is \emph{accepted}
by $\kappa$,
written as $\kappa \models \condAB$, iff
$\kappa(AB) < \kappa(\AnotB)$, \ie, iff $AB$ is strictly more plausible than $\AnotB$.
This is lifted to \BBs\ via $\kappa\models\Delta$ if $\kappa\models (B|A)$ for all $(B|A)\in\Delta$.
Consistency of such a belief base $\Delta$ can be defined in terms of OCFs \cite{Pearl1990systemZTARK}: $\Delta$ is \emph{(strongly) consistent} iff there is an OCF $\kappa$ such that $\kappa \models \Delta$ and \(\kappa(\omega) < \infty\) for all \(\omega \in \Omega\).
We focus on (strongly) consistent belief bases in the sense of \cite{Pearl1990systemZTARK,GoldszmidtPearl96} in order to elaborate our approach without having to deal with distracting technical particularities.

\section{Conditional Syntax Splitting}
Syntax splittings describe that a belief base contains completely independent information about different parts of the signature. According to \cite{KernIsbernerBeierleBrewka2020KR},
a belief base $\Delta$ \emph{splits} into subbases $\DeltaOne, \DeltaTwo$ 
if $\{\Sigma_1, \Sigma_2\}$ is a partition of $\Sigma$
such that $\Delta = \DeltaOne \cup \DeltaTwo$, $\DeltaI \subset (\cL_i | \cL_i)$, $\cL_i = \cL(\signa_i)$ for $i \in \{1, 2\}$, $\signa_1 \cap \signa_2 = \emptyset$, and $\signa_1 \cup \signa_2 = \signa$, denoted as 
\begin{equation}
	\nutzeTextstyle
	\Delta = \DeltaOne \bigcup_{\Sigma_1,\Sigma_2} \DeltaTwo.
\end{equation}
Syntax splittings are very useful for formalizing the idea that independent information about different topics should not affect each  other in reasoning \cite{KernIsbernerBeierleBrewka2020KR}.
Syntax splittings were generalized to \emph{conditional} syntax splittings \cite{HeyninckKernIsbernerMeyerHaldimannBeierle2023AAAI}, which allow subbases to share some atoms in a given subsignature $\Sigma_3$. 

\begin{definition}[conditional syntax splitting \cite{HeyninckKernIsbernerMeyerHaldimannBeierle2023AAAI}]
	A belief base $\Delta$
	\emph{splits into subbases $\DeltaOne$,$\DeltaTwo$ conditional on $\Sigma_3$}
	if there are \(\Sigma_1, \Sigma_2 \subseteq \Sigma\)
	such that $\DeltaI = \Delta \cap ({\cal L}(\Sigma_i\cup\Sigma_3)\mid {\cal L}(\Sigma_i\cup\Sigma_3))$ for $i \in \{1,2\}$, 
	and $\{\Sigma_1, \Sigma_2,\Sigma_3\}$ is a partition of $\Sigma$.
	This is denoted as
	\begin{equation}
		\label{eq:def_cond_syn_split}
		\nutzeTextstyle
		\CondSynSplitGen.
	\end{equation}
\end{definition}

Unlike syntax splitting, conditional syntax splitting does not require 
$\DeltaOne$ and $\DeltaTwo$ to be disjoint.
For the rest of this article we will use, 
for $i\in\{1,2\}$,
\begin{align*}
	\DeltaThree &= \DeltaOne\cap\DeltaTwo\\
	\DeltaIWoThree &= \DeltaI\setminus\DeltaThree
\end{align*}
for ease of reading. Given a complete conjunction over \(\Sigma_3\), which corresponds to a world in $\OmegaOf{\Sigma_3}$, conditional syntax splittings in general do not ensure complete independence of \(\DeltaOne\) and \(\DeltaTwo\); for details see \cite[Example 6]{HeyninckKernIsbernerMeyerHaldimannBeierle2023AAAI}. Therefore, \emph{safe} conditional syntax splittings were introduced.
\begin{definition}[safe conditional syntax splitting \cite{HeyninckKernIsbernerMeyerHaldimannBeierle2023AAAI}]
	\label{def:safe:splitting}
	A conditional syntax splitting $\CondSynSplitGen$ is called \emph{safe}, denoted as
	\begin{equation}
		\label{eq_safecsynsplit}
		\nutzeTextstyle
		\safeCondSynSplitGen
	\end{equation}
	if the following \emph{safety property} holds for $i,i'\in\{1,2\}$, $i\neq i'$:
	\begin{align}
		\label{eq_safety}
		\begin{split}
			\text{for every }\omegaIThree\in\OmegaOf{\Sigma_i\cup\Sigma_3},
			\text{ there is } \omegaJ\in\OmegaOf{\Sigma_{i'}}%
			\quad\text{ such that }\quad
			\omegaIThreeJ\not\models \bigvee_{(F|E)\in \DeltaJ} E\land \lnot F.
		\end{split}
	\end{align}
\end{definition}
Intuitively, the safety condition \eqref{eq_safety} guarantees proper independence with respect to falsification between the subbases $\Delta_1$ and $\Delta_2$; for a more detailed analysis we refer to \cite{HeyninckKernIsbernerMeyerHaldimannBeierle2023AAAI,SpiegelHaldimannHeyninckKernIsbernerBeierle2025IJCAI}.
In the following proposition, we show that the safety condition \eqref{eq_safety} demands, in essence, that no complete conjunction over $\Sigma_3$ may force the falsification of a conditional in $\Delta$ when considering $\Sigma$ as a whole. 
\begin{proposition}
    \label{lem_safety}
    Let $\CondSynSplitGen$. Then the safety property is equivalent to the following: 
    For every $\omega^3 \in \OmegaOf{\Sigma_3}$ there is $\omega^1\omega^2 \in \OmegaOf{\Sigma \setminus \Sigma_3}$ such that $\omega = \omega^1\omega^2\omega^3 \in \OmegaOf{\Sigma}$ does not falsify any conditional in $\Delta$.
\end{proposition}
\begin{proof}
    Direction \enquote{$\Rightarrow$}:
    Let $\omega^3 \in \OmegaOf{\Sigma_3}$. Now choose some $\widehat{\omega}^1 \in \OmegaOf{\Sigma_1}$. According to the safety property, there is $\omega_2 \in \OmegaOf{\Sigma_2}$ such that $\widehat{\omega}^1\omega^2\omega^3$ does not falsify any conditionals in $\Delta_2$. Therefore, also $\omega^2\omega^3$ does not falsify any conditional in $\Delta_2$. By applying the safety property again, we can obtain a new $\omega^1 \in \OmegaOf{\Sigma_1}$ such that $\omega^1\omega^2\omega^3$ does not falsify any conditional in $\Delta_1$. Since the conditionals in $\Delta_2$ do not contain any symbols from $\Sigma_1$, appending $\omega_1$ to $\omega^2\omega^3$ does not lead to new conditionals from $\Delta_2$ being falsified. Therefore, $\omega^1\omega^2\omega^3$ does not falsify any conditionals from $\Delta_1 \cup \Delta_2 = \Delta$.
    
    Direction ``$\Leftarrow$'': We show this by contraposition. Let $i,j\in\{1,2\},i\neq j$. Let $\omega^i\omega^3\in\Omega(\Sigma_i\cup\Sigma_3)$. Assume the safety property does not hold, \ie, (*) there is some $\omega^i\omega^3\in\Omega(\Sigma_i\cup\Sigma_3)$ such that there is no $\omega^j$ such that $\omega^i\omega^j\omega^3$ falsifies no conditional in $\Delta_j$. Because $\Delta_j\subseteq(\cL(\Sigma_j\cup\Sigma_3)|\cL(\Sigma_j\cup\Sigma_3))$ we have for any conditional $(B_l|A_l)\in\Delta_j$ that $\omega^i\omega^j\omega^3\models A_l\ol{B_l}$ iff $\omega^j\omega^3\models A_l\ol{B_l}$. Thus, (*) implies that there is $\omega^3$ such that there is no $\omega^j$ such that $\omega^3\omega^j$ falsifies no conditional in $\Delta_j$. Reversing the roles of $i$ and $j$ we obtain the corresponding result for $\omega^3\omega^i$. Thus, it follows that there cannot be a $\omega^3\omega^i\omega^j$ falsifying no conditional in $\Delta$. Then, by setting $i=1$, $j=2$, it follows that there cannot be $\omega^1\omega^2 \in \OmegaOf{\Sigma \setminus \Sigma_3}$ such that $\omega^1\omega^2\omega^3$ does not falsify any conditional in $\Delta$.
\end{proof}

Safe conditional syntax splitting provides similar benefits for inductive inference as syntax splitting \cite{HeyninckKernIsbernerMeyerHaldimannBeierle2023AAAI}. Reasoning in the language of $\Delta_1$ is independent of the conditionals in $\Delta_2$ and vice versa, given that we have full knowledge over the \emph{conditional pivot} $\Sigma_3$, \ie, we know the valuation assigned to all elements in $\Sigma_3$. The conditional pivot thereby represents the elements shared between the subbases.
The safety condition was generalized, leading to so-called generalized safe conditional syntax splittings \cite{SpiegelHaldimannHeyninckKernIsbernerBeierle2025IJCAI}, which allow for more splittings of a \BB\ and more \BBs\ with relevant splittings.

\begin{definition}[generalized safe conditional syntax splitting \cite{SpiegelHaldimannHeyninckKernIsbernerBeierle2025IJCAI}]
	\label{def_generalsafety}
	A conditional syntax splitting $\CondSynSplitGen$ is \emph{\genSafe}, denoted as
	\begin{equation}
		\label{eq_gsafecsynsplit}
		\nutzeTextstyle
		\genSafeCondSynSplitGen
	\end{equation}
	if the following \emph{\genSafety\ property} holds for $i,i'\in\{1,2\}, i\neq i'$:
	\begin{align}
		\label{eq_gsafety}
		\begin{split}
			\text{for every }\omegaIThree\in\OmegaOf{\Sigma_i\cup\Sigma_3},
			\text{ there is } \omegaJ\in\OmegaOf{\Sigma_{i'}}%
			\quad \text{ such that }\quad  \omegaIThreeJ\not\models \bigvee_{(F|E)\in \DeltaJWoThree} E\land \lnot F.
		\end{split}
	\end{align}
\end{definition}

The decisive difference in \eqref{eq_gsafety} compared to \eqref{eq_safety} is that only conditionals in $\DeltaJWoThree$ are considered for the \genSafety\ property as opposed to all conditionals in $\DeltaJ$ for the safety property.

\section{c-Representations and Constraint Splitting}
c-Representations \cite{KernIsberner2001habil,KernIsberner2004amai} are special ranking functions that are obtained by assigning integer impacts
$\kappaiminusVektor = (\eta_1,\ldots,\eta_n)$
to conditionals of a \BB\ $\Delta = \{(B_1|A_1), \ldots, (B_n|A_n)\}$ and are given by
\begin{align}\label{form:def:c-representation}
	\induzierteOCF{\kappaiminusVektor}(\omega)
	=\sum\limits_{\substack{1 \leq i \leq n\\\omega\models A_i\ol{B}_i}}\kappaiminus{i}
\end{align}
where the $\eta_i$ are chosen such that $\induzierteOCF{\kappaiminusVektor} \models \Delta$. This condition can be characterized by a constraint satisfaction problem (\CSP) \cite{KernIsberner2001habil,BeierleEichhornKernIsbernerKutsch2018AMAI}.
The constraints in this CSP can be expressed using, for each conditional $\delta_i \in \Delta$, the following \emph{constraint inducing sets} containing sets of verified and falsified conditionals \cite{BeierleKutschSauerwald2019AMAIcompilation}:  
\begin{align}
  \label{constraint-inducing_set_V}
	\VSet_{\Delta}(\delta_i)=\{\{(B_j|A_j)\in\Delta\setminus\{\delta_i\}\mid \omega\models A_j\ol{B_j}\}\mid \omega\in\ver(B_i|A_i)\}\\
  \label{constraint-inducing_set_F}
	\FSet_{\Delta}(\delta_i)=\{\{(B_j|A_j)\in\Delta\setminus\{\delta_i\}\mid \omega\models A_j\ol{B_j}\}\mid \omega\in\fal(B_i|A_i)\}
\end{align}
Employing these constraint-inducing sets \eqref{constraint-inducing_set_V}, \eqref{constraint-inducing_set_F}, the
constraint satisfaction problem $\CSP(\Delta) = \{\constraintOfWrtI{\Delta}{}{1}, \ldots, \constraintOfWrtI{\Delta}{}{n}\}$ over the constraint variables $\eta_1,\dots,\eta_n$ taking values in $\mathbb{N}$ is specified as follows for all
$\delta_i\in\Delta$:

\begin{equation}
    \constraintOfWrtI{\Delta}{}{i}: \eta_i>\min\{\sum_{\delta_j\in S} \eta_j\mid S\in \VSet_\Delta(\delta_i)\} - \min\{\sum_{\delta_j\in S} \eta_j\mid S\in \FSet_\Delta(\delta_i)\}.
\end{equation}
A solution of $\CSP(\Delta)$ is a vector $\kappaiminusVektor=(\eta_1,\dots,\eta_n)\in\mathbb{N}^n$. We denote with $\solutionsR$ the set of all solutions of $\CSP(\Delta)$.
We make use of the following notation.
\begin{definition}[$\marg$, projection to subbase]
	\label{def_crmarg}
	Let $\Delta=\{\delta_1,\dots,\delta_n\}$ be a belief base, let $\Delta'=\{\delta_{i_1},\dots,\delta_{i_m}\}\subseteq\Delta$, and let $\kappaiminusVektor=(\eta_1,\dots, \eta_n)\in\solutionsR$. Then we define the \emph{projection} of $\kappaiminusVektor$ and $\CSP(\Delta$) to $\Delta'$:
	\begin{itemize}
		\item $\kappaiminusVektor\marg_{\Delta'}=(\eta_{i_1},\dots,\eta_{i_m})$
		\item $\CSP(\Delta)\marg_{\Delta'}=\{\constraintOfWrtI{\Delta}{}{i}\mid \delta_i\in\Delta'\}$
	\end{itemize}
\end{definition}

In Definition~\ref{def_crmarg}, the operator $\marg$ is used as a projection of $\kappaiminusVektor$ and $\CSP(\Delta)$ respectively to the conditionals of $\Delta'$.
Note that $\CSP(\Delta)\marg_{\Delta'}$ might mention constraint variables not associated with a conditional in $\Delta'$ but with a conditional in $\Delta\setminus\Delta'$.
Furthermore, for two vectors $\kappaiminusVektor^1=(\eta_1^1,\dots,\eta_k^1)$ and $\kappaiminusVektor^2=(\eta_1^2,\dots,\eta_l^2)$, we denote with $(\kappaiminusVektor^1,\kappaiminusVektor^2)$ the concatenation of $\kappaiminusVektor^1$ and $\kappaiminusVektor^2$, \ie, $(\kappaiminusVektor^1,\kappaiminusVektor^2)=(\eta_1^1,\dots,\eta_k^1,\eta_1^2,\dots,\eta_l^2)$. We illustrate Definition~\ref{def_crmarg} with an example.
\begin{example}
	\label{ex_constraints}
	Let $\DeltaBird=\{\delta_1\!:\!(f|b),\ \delta_2\!:\!(\ol{f}|p),\  \delta_3:(b|p),\ \delta_4\!:\!(w|b)\}$.
	Then $\CSP(\DeltaBird)$ is given by:
	\begin{align*}
		\constraintOfWrtI{\DeltaBird}{}{1}:\eta_1 &> \min\{0,\eta_2,\eta_2+\eta_4,\eta_4\} - \min\{0,\eta_4\}\\
		\constraintOfWrtI{\DeltaBird}{}{2}:\eta_2 &> \min\{\eta_1,\eta_3,\eta_1+\eta_4\} - \min\{0,\eta_3,\eta_4\}\\
		\constraintOfWrtI{\DeltaBird}{}{3}:\eta_3 &> \min\{\eta_1,\eta_2,\eta_1+\eta_4, \eta_2+\eta_4\} - \min\{0,\eta_2\}\\
		\constraintOfWrtI{\DeltaBird}{}{4}:\eta_4 &> \min\{0,\eta_1,\eta_2\} - \min\{0,\eta_1,\eta_2\}
	\end{align*}
	\begin{table}[tb]%
		\centering%
		\begin{tabular}{ccccccccccc}
			\toprule
			$\omega$ & $(f|b)$ & $(\ol{f}|p)$ & $(b|p)$ & $(w|b)$ & \hspace{.3cm} & $\omega$ & $(f|b)$ & $(\ol{f}|p)$ & $(b|p)$ & $(w|b)$ \\
			\midrule
			$\ol{p} \, \ol{b} \, \ol{f} \, w$ &            &            &            &  && $\ol{p} \, \ol{b} \, \ol{f} \, \ol{w}$ &            &            &            &  \\
			$\ol{p} \, \ol{b} \,    {f} \, w$ &            &            &            &   && $\ol{p} \, \ol{b} \,    {f} \, \ol{w}$ &            &            &            &  \\
			$\ol{p} \,    {b} \, \ol{f} \, w$ & \textsf{f} &            &            & \textsf{v} && $\ol{p} \,    {b} \, \ol{f} \, \ol{w}$ & \textsf{f} &            &            &  \textsf{f}\\
			$\ol{p} \,    {b} \,    {f} \, w$ & \textsf{v} &            &            & \textsf{v} && $\ol{p} \,    {b} \,    {f} \, \ol{w}$ & \textsf{v} &            &            &  \textsf{f}\\
			$   {p} \, \ol{b} \, \ol{f} \, w$ &            & \textsf{v} & \textsf{f} &  && $   {p} \, \ol{b} \, \ol{f} \, \ol{w}$ &            & \textsf{v} & \textsf{f} &  \\
			$   {p} \, \ol{b} \,    {f} \, w$ &            & \textsf{f} & \textsf{f} &  && $   {p} \, \ol{b} \,    {f} \, \ol{w}$ &            & \textsf{f} & \textsf{f} &  \\
			$   {p} \,    {b} \, \ol{f} \, w$ & \textsf{f} & \textsf{v} & \textsf{v} & \textsf{v} && $   {p} \,    {b} \, \ol{f} \, \ol{w}$ & \textsf{f} & \textsf{v} & \textsf{v} &  \textsf{f}\\
			$   {p} \,    {b} \,    {f} \, w$ & \textsf{v} & \textsf{f} & \textsf{v} & \textsf{v} && $   {p} \,    {b} \,    {f} \, \ol{w}$ & \textsf{v} & \textsf{f} & \textsf{v} &  \textsf{f}\\
			\bottomrule
		\end{tabular}
		\caption{Verification and falsification behaviour of worlds \wrt\ $\DeltaBird$ in Example~\ref{ex_constraints}. The letters \textsf{v}\ and \textsf{f}\ indicate which conditionals are verified (\textsf{v}) or falsified (\textsf{f}).}
		\label{tab_ex_DeltaBird}
	\end{table}
	The verification and falsification behaviour of worlds \wrt\ $\DeltaBird$ can be seen in Table~\ref{tab_ex_DeltaBird}.
	A possible solution for $\CSP(\DeltaBird)$ is $\kappaiminusVektor\in\solutionsRof{\DeltaBird}$ with $\kappaiminusVektor=(1,2,2,1)$. The projection of $\kappaiminusVektor$ to $\DeltaBird_1=\{(\ol{f}|p),(w|b)\}$ is $\kappaiminusVektor\marg_{\DeltaBird_1}=(2,1)$. The \projektion\ of $\CSP(\DeltaBird)$ to $\DeltaBird_1$ is then given by $\CSP(\DeltaBird)\marg_{\DeltaBird_1}=\{\constraintOfWrtI{\DeltaBird}{}{2},\constraintOfWrtI{\DeltaBird}{}{4}\}$.
\end{example}
\begin{figure}[tb]
\newlength{\abstandInTabelle}
\setlength{\abstandInTabelle}{6mm}
\renewcommand{\nsrV}{\VSet_{\Delta}}
\renewcommand{\nsrF}{\FSet_{\Delta}}
\[
\begin{array}{l@{\,\,}c@{\quad}l}
\displayTransformationRule%
{\refSSVker\ \nameSSV:}%
{\displaystyle\frac{\nsrPaarKER{\nsrV \cup\{S, S'\}}{\nsrF}{i}}%
{\nsrPaarKER{\nsrV \cup\{S\}}{\nsrF}{i}}}%
{S \subsetneq S'}
\\[\abstandInTabelle]

\displayTransformationRule%
{\refSSFker\ \nameSSF:}%
{\displaystyle\frac{\nsrPaarKER{\nsrV}{\nsrF \cup\{S, S'\}}{i}}%
{\nsrPaarKER{\nsrV}{\nsrF \cup\{S\}}{i}}}%
{S \subsetneq S'}
\\[\abstandInTabelle]

\displayTransformationRule%
{\refELMker\ \nameELM:}%
{\displaystyle\frac{\nsrPaarKER{\left\{V_1 \cup \{\delta\},\ldots, V_p \cup \{\delta\}\right\}}{\left\{F_1 \cup \{\delta\},\ldots, F_q \cup \{\delta\}\right\}}{i}}%
		{\nsrPaarKER{\left\{V_1,\ldots,V_p\right\}}{\left\{F_1,\ldots,F_q\right\}}{i}}}%
{}
\\[\abstandInTabelle]

\displayTransformationRule%
{\refTRIVIAL\ \nameTRIVIAL:}%
{\displaystyle\frac{\nsrPaarKER{\nsrV}{\nsrF}{i}}%
{\nsrPaarKER{\{\leereMenge\}}{\{\leereMenge\}}{i}}}%
{\nsrV = \nsrF}
\\[\abstandInTabelle]

\displayTransformationRule%
{\refSUBSETS\ \nameSUBSETS:}%
{\displaystyle\frac{\nsrPaarKER{\left\{S_1 \dotcup T,\ldots, S_p \dotcup T\right\}}{\left\{S_1 \dotcup T',\ldots, S_p \dotcup T'\right\}}{i}}%
		{\nsrPaarKER{\left\{T\right\}}{\left\{T'\right\}}{i}}}%
{}
\\[\abstandInTabelle]

\displayTransformationRule%
{\refCIRCLE\ \nameCIRCLE:}%
{\displaystyle\frac{\nsrPaarKER{\calD \dotcup\{\{\delta_j\}\}}{\{\leereMenge\}}{i} \quad {\nsrPaarKER{\calD \dotcup\{\{\delta_i\}\}}{\{\leereMenge\}}{j}}}%
{\nsrPaarKER{\calD}{\{\leereMenge\}}{i} \quad \nsrPaarKER{\calD}{\{\leereMenge\}}{j}}}
{i \neq j}
\\[\abstandInTabelle]

\end{array}
\]
\caption{Transformation rules $\natSpcr = \{R1,\dots, R6\}$ for simplifying (the constraint-inducing sets of)
$\CSP(\Delta)$. A pair~\(\nsrPaarKER{\nsrV}{\nsrF}{i}\) represents the sets of constraint variables in the minimum expressions associated to the verification and the falsification, respectively, of the $i$-th conditional
$\delta_i \in \Delta$
in the constraint $\constraintOfWrtI{\Delta}{}{i} \in \CSP(\Delta)$ modeling the acceptance condition of $\delta_i$.} 
\label{fig:opt_transformationrules}
\end{figure}
 \begin{definition}[solution-preserving constraint reduction, spcr \cite{WilhelmSezginKernIsbernerHaldimannBeierleHeyninck2023JELIA}]
	Let $\Delta$ be a conditional belief base. Furthermore, let $\spcr : (\VSet,\FSet) \mapsto (\spcr(\VSet), \spcr(\FSet))$ be defined such that $\spcr(\VSet) \sqsubseteq \VSet$ and $\spcr(\FSet) \sqsubseteq \FSet$, with $X' \sqsubseteq X$ meaning that (a) for every $x' \in X'$, there is $x \in X$ such that $x' \subseteq x$, and (b) for every $x \in X$, there is $x' \in X'$ such that $x' \subseteq x$.
	The mapping $\CSP(\Delta) \mapsto \CSP^\spcr(\Delta)$ with $\CSP^\spcr(\Delta) = \{\constraintOfSpcrI{\Delta}{i} \mid \constraintOfWrtI{\Delta}{}{i} \in \CSP(\Delta)\}$ such that
    \begin{equation}
        \constraintOfSpcrI{\Delta}{i}: \eta_i>\min\{\sum_{\delta_j\in S} \eta_j\mid S\in \spcr(\VSet_\Delta(\delta_i))\} - \min\{\sum_{\delta_j\in S} \eta_j\mid S\in \spcr(\FSet_\Delta(\delta_i))\}.
    \end{equation}
	is a \emph{solution-preserving constraint reduction (spcr)} \wrt\ $\spcr$ if
    $\solutionsOf{\CSP^\spcr(\Delta)} = \solutionsR$ for all consistent belief bases $\Delta$.
\end{definition}
In the following, we will use 
$\spcr$ to refer both to the mapping between constraint inducing sets, as well as to the corresponding spcr \wrt\ $\spcr$, \ie, the mapping between CSPs.
For the sake of conciseness, we will sometimes write $\VSet^\spcr_\Delta(\delta_i)$ instead of $\spcr(\VSet_\Delta(\delta_i))$ and $\FSet^\spcr_\Delta(\delta_i)$ instead of $\spcr(\FSet_\Delta(\delta_i))$.
Note that choosing $\spcr$ as the identity function yields the (trivial) spcr $\idFct$, \ie, $\CSP(\Delta)=\CSP^{\idFct}(\Delta)$ for all belief bases $\Delta$.
A convenient way to implement solution-preserving constraint reductions is through the use of rewriting systems.
Figure~\ref{fig:opt_transformationrules} shows the set $\natSpcr = \{R1,\dots, R6\}$ of rewriting rules for simplifying $\CSP(\Delta)$. These rules were first given in \cite{BeierleKutschSauerwald2019AMAIcompilation} for speeding up the computation of c-representations and of c-inference \cite{BeierleEichhornKernIsbernerKutsch2018AMAI} and then extended in \cite{BeierleHaldimannKernIsberner2021Boeger75Festschrift,WilhelmGKIBeierle26KERsp}. 
The application of any sequence of rules from $\{R1,\dots,R6\}$ does not change the set of solutions $\solutionsR$, and furthermore, 
$\natSpcr$ is terminating and confluent; thus, exhaustive application of $\natSpcr$ yields a uniquely determined CSP \cite{WilhelmGKIBeierle26KERsp}.

\begin{proposition}[spcr $\natSpcr$ \cite{WilhelmGKIBeierle26KERsp}]
    \label{prop_nat-spcr}
	Let $\natSpcr = \{R1,\dots, R6\}$ be the set of transformation rules on constraint-inducing sets from Figure~\ref{fig:opt_transformationrules}, and let $\natSpcr$  also denote the mapping obtained by exhaustive application of these rules.
	Then $\natSpcr : \CSP(\Delta) \mapsto \CSP^\natSpcr(\Delta)$ is an spcr.
\end{proposition}

For the rest of this article, we will make use of the following notation for different spcrs:
\begin{itemize}
	\item $\spcr$ refers to any fixed, but arbitrary spcr.
	\item $\natSpcr$ refers to the spcr defined via $\natSpcr=\{R1,\dots,R6\}$.
	\item $\spcrSubset$ refers to the spcr defined via $\spcrSubset=\{R1,R2\}$.
\end{itemize}
It is easy to see that also $\spcrSubset$ is confluent and terminating, thus $\CSP^\spcrSubset$ is unique. We illustrate $\natSpcr$ and $\spcrSubset$ with an example.

\begin{example}
	\label{exa_natspcr}
	Applying $\spcrSubset$ to $\CSP(\DeltaBird)$ from Example~\ref{ex_constraints} yields $\CSP^\spcrSubset(\DeltaBird)$ given by 
	\begin{align*}
		\constraintOfWrtI{\DeltaBird}{\spcrSubset}{1}: \eta_1 &> \min\{0\} - \min\{0\}\\
		\constraintOfWrtI{\DeltaBird}{\spcrSubset}{2}: \eta_2 &> \min\{\eta_1,\eta_3\} - \min\{0\}\\
		\constraintOfWrtI{\DeltaBird}{\spcrSubset}{3}: \eta_3 &> \min\{\eta_1,\eta_2\} - \min\{0\}\\
		\constraintOfWrtI{\DeltaBird}{\spcrSubset}{4}: \eta_4 &> \min\{0\} - \min\{0\},
	\end{align*}
	where rules $R1$ and $R2$ eliminate the supersets in all constraints.
	Applying $\natSpcr$ to $\CSP(\DeltaBird)$ yields $\CSP^\natSpcr(\DeltaBird)$ given by 
	\begin{align*}
		\constraintOfNatspcrI{\DeltaBird}{1}: \eta_1 &> \min\{0\} - \min\{0\}\\
		\constraintOfNatspcrI{\DeltaBird}{2}: \eta_2 &> \min\{\eta_1\} - \min\{0\}\\
		\constraintOfNatspcrI{\DeltaBird}{3}: \eta_3 &> \min\{\eta_1\} - \min\{0\}\\
		\constraintOfNatspcrI{\DeltaBird}{4}: \eta_4 &> \min\{0\} - \min\{0\},
	\end{align*}
	where rule $R_6$ additionally removes the cyclic dependency between $\constraintOfWrtI{\DeltaBird}{}{2}$ and $\constraintOfWrtI{\DeltaBird}{}{3}$.
\end{example}

	\begin{figure}
	\centering
	\begin{subfigure}{.48\textwidth}
		\centering
		\begin{tikzpicture}
			\centering
			
			\usetikzlibrary{calc}
			
			\node (1) at (0,0) {$\delta_1$};
			\node (2) at (2,0) {$\delta_2$};
			\node (3) at (0,-2) {$\delta_3$};
			\node (4) at (2,-2) {$\delta_4$};
			
			\draw[->] (1) to[bend left=10] (2);
			\draw[->] (1) to[bend left=10] (4);
			
			\draw[->] (2) to[bend left=10] (1);
			\draw[->] (2) to[bend left=10] (3);
			\draw[->] (2) to[bend left=10] (4);
			
			\draw[->] (3) to (1);
			\draw[->] (3) to[bend left=10] (2);
			\draw[->] (3) to (4);
			
			\draw[->] (4) to[bend left=10] (1);
			\draw[->] (4) to[bend left=10] (2);
		\end{tikzpicture}
		\caption{
			$\CSP(\DeltaBird)$ in  
			Example~\ref{ex_constraints}}
		\label{subfig_dependencyBirdA}
	\end{subfigure}
	\begin{subfigure}{.48\textwidth}
		\centering
		\begin{tikzpicture}
			\centering
			
			\usetikzlibrary{calc}
			
			\node (1) at (0,0) {$\delta_1$};
			\node (2) at (2,0) {$\delta_2$};
			\node (3) at (0,-2) {$\delta_3$};
			\node (4) at (2,-2) {$\delta_4$};
			
			\draw[->] (2) to (1);
			\draw[->] (3) to (1);
		\end{tikzpicture}
		\caption{
			$\CSP^\natSpcr(\DeltaBird)$ in  
			Example~\ref{exa_natspcr}}
		\label{subfig_dependencyBirdB}
	\end{subfigure}
	\caption{Dependencies among the constraint variables of $\CSP(\DeltaBird)$ respectively $\CSP^\natSpcr(\DeltaBird)$. An edge $\delta_i\rightarrow\delta_j$ represents that $\delta_i$ directly depends on $\delta_j$ in $\DeltaBird$ \wrt\ $\idFct$ in (a) and \wrt\ $\natSpcr$ in (b).
	}
	\label{fig_dependencyBird}
\end{figure} 
For a conditional $\delta \in \Delta$ we denote with $\scopeOf{\Delta}{\spcr}{\delta}$ the set of conditionals whose corresponding constraint variables occur in $\constraintOfSpcrI{\Delta}{}$. More precisely, the scope of $\delta$ can be defined in terms of the constraint inducing sets $\VSet_\Delta(\delta)$ and $\FSet_\Delta(\delta)$:
\begin{equation*}
    \scopeOf{\Delta}{\spcr}{\delta} = \left(\bigcup\spcr(\VSet_\Delta(\delta))\right) \cup \left(\bigcup\spcr(\FSet_\Delta(\delta))\right) .
\end{equation*}
Note that application of an spcr may reduce the scope of constraints significantly. For instance, in Example~\ref{ex_constraints}, we have $\scopeOf{\Delta}{\id}{\delta_2} = \{\delta_1, \delta_3, \delta_4\}$, 
while, after applying $\spcrSubset$ and $\natSpcr$ in Example \ref{exa_natspcr}, we have $\scopeOf{\Delta}{\spcrSubset}{\delta_2} = \{\delta_1,\delta_3\}$ and just $\scopeOf{\Delta}{\natSpcr}{\delta_2} = \{\delta_1\}$. 
In this way, an spcr helps with reducing dependencies among the impact factors to relevant dependencies. It should be emphasized that relevance here is not based on numerical (sometimes accidental) relationships among the impact factors but on algebraic properties of arithmetics.

In \cite{BeierleHaldimannKernIsberner2021Boeger75Festschrift} notions of constraint-based dependence and independence between conditionals and sets thereof were defined.

\begin{definition}[$\directlyDependsSingle,\directlyDepends,\unabhSymbol$ \cite{BeierleHaldimannKernIsberner2021Boeger75Festschrift}]
	\label{def_constrindep}
	Let $\Delta$ be a \BB\ with $\delta_i,\delta_j\in\Delta$ and let $\spcr$ be an spcr. We say that $\delta_i$ directly depends on $\delta_j$ in $\Delta$ \wrt\ $\spcr$, written $\delta_i\directlyDependsSingleOfWrtSpcr{\Delta}\delta_j$, %
	if $\delta_j \in \scopeOf{\Delta}{\spcr}{\delta_i}$. Then $\directlyDependsOfWrtSpcr{\Delta}$ is the reflexive and transitive closure of $\directlyDependsSingleOfWrtSpcr{\Delta}$. We lift $\directlyDependsOfWrtSpcr{\Delta}$ to \BBs\ $\Delta_1,\Delta_2 \subseteq\Delta$ 
	via $\Delta_1\directlyDependsOfWrtSpcr{\Delta}\Delta_2$ if there is $\delta_1\in\Delta_1$ such that there is $\delta_2\in\Delta_2$ with $\delta_1\directlyDependsOfWrtSpcr{\Delta}\delta_2$.

	We call $\Delta_1$ and $\Delta_2$ constraint-wise independent in $\Delta$ \wrt\ $\scpr$ if
	$\Delta_1\not\directlyDependsOfWrtSpcr{\Delta}\Delta_2$ and
	$\Delta_2\not\directlyDependsOfWrtSpcr{\Delta}\Delta_1$, written $\Delta_1\unabhSymbolOfWrtSpcr{\Delta}\Delta_2$.
\end{definition}

Note that $\Delta_1\not\directlyDependsOfWrtSpcr{\Delta}\Delta_2$ means that for all $\delta_1\in\Delta_1,\delta_2\in\Delta_2$ we have \(
\delta_1\not\directlyDependsOfWrtSpcr{\Delta}\delta_2
\).
Constraint-wise (in)dependence can be visualized using constraint dependency graphs (introduced as adjacency graphs in \cite{WilhelmKernIsbernerAAAI24}) as shown in Figure~\ref{fig_dependencyBird}. For $\CSP^\spcr(\Delta)$, the constraint dependency graph is a directed graph given by $(V,E)$ with the set of nodes $V=\Delta$ and the set of edges $E=\{(\delta_i,\delta_j)\mid \delta_j\in \scopeOf{\Delta}{\spcr}{\delta_i}, \delta_i\neq\delta_j\}$. For $\delta_i,\delta_j\in\Delta$ with $\delta_i\neq\delta_j$, $\delta_i\directlyDependsSingleOfWrtSpcr{\Delta}\delta_j$ then means that there is an edge from $\delta_i$ to $\delta_j$ in the constraint dependency graph of $\CSP^\spcr(\Delta)$. Then $\delta_i\directlyDependsOfWrtSpcr{\Delta}\delta_j$ implies that there is a path from $\delta_i$ to $\delta_j$ in the constraint dependency graph of $\CSP^\spcr(\Delta)$. Lifting this to subbases $\Delta_i\subseteq\Delta$, we take the hyperedge consisting of all nodes corresponding to the conditionals in $\Delta_i$ into account instead of the singular node $\delta_i$. $\Delta_1\unabhSymbolOfWrtSpcr{\Delta}\Delta_2$ means that there is no path from the hyperedge representing $\Delta_1$ to the hyperedge representing $\Delta_2$ and vice versa in the constraint dependency graph of $\CSP^\spcr(\Delta)$ (cf. Figure~\ref{fig_constrSplits}).
Utilizing the notion of constraint (in)dependence, the notion of constraint splittings was introduced in  \cite{BeierleHaldimannKernIsberner2021Boeger75Festschrift}.

\begin{definition}[constraint splitting \cite{BeierleHaldimannKernIsberner2021Boeger75Festschrift}]	
	\label{def_constrsplit}
	Let $\Delta$ be a \BB\ and let $\spcr$ be an spcr.
	A partition $\{\Delta_1,\dots,\Delta_m\}$ of $\Delta$ is a \emph{constraint splitting of $\Delta$ \wrt\ $\spcr$}, written $\constraintSplitGen$, if, for all $i,j\in\{1,\dots,m\}$ with $i\neq j$: $\Delta_i\unabhSymbolOfWrt{\Delta}{\spcr}\Delta_j$.
\end{definition}

(Unconditional) constraint splittings in the sense of Definition~\ref{def_constrsplit} imply that the constraint systems can be computed and solved independently on the subbases after applying $\spcr$ and that each solution for $\Delta$ can be built from solutions from the subbases \cite{BeierleHaldimannKernIsberner2021Boeger75Festschrift}. We illustrate constraint splittings with an example.

\begin{example}%
\label{ex_constrsplits}
Consider again $\DeltaBird$ from Example~\ref{ex_constraints} and $\CSP(\DeltaBird)$. The constraint dependency graphs of $\DeltaBird$ \wrt\ the spcrs $\idFct$ and $\natSpcr$ can be seen in Figure~\ref{fig_dependencyBird}.
This \CSP admits no constraint splitting at all.
However, after applying 
$\natSpcr$ from Figure~\ref{fig:opt_transformationrules} we obtain $\CSP^\natSpcr(\DeltaBird)$ as seen in Example~\ref{exa_natspcr}. Thus, $\DeltaBird$ has the constraint splitting \wrt\ $\natSpcr$
\[
\{\delta_1,\delta_2,\delta_3\}\constraintCupNatSpcr\{\delta_4\}=\{(\ol{f}|p),(b|p),(f|b)\}\constraintCupNatSpcr\{(w|b)\}.
\]
This is also a safe conditional syntax splitting  with $\Sigma_1=\{f,p\},\Sigma_2=\{w\}$ and $\Sigma_3=\{b\}$.
Observe that $(\ol{f}|p)\not\directlyDependsOfWrt{\DeltaBird}{\natSpcr}(w|b)$, $(b|p)\not\directlyDependsOfWrt{\DeltaBird}{\natSpcr}(w|b)$
and
$(f|b)\not\directlyDependsOfWrt{\DeltaBird}{\natSpcr}(w|b)$. Therefore $\{(\ol{f}|p),(b|p),(f|b)\}\not\directlyDependsOfWrt{\DeltaBird}{\natSpcr}\{(w|b)\}$. Since $\{(w|b)\}\not\directlyDependsOfWrt{\DeltaBird}{\natSpcr}\{(\ol{f}|p),(b|p),(f|b)\}$ holds as well, we have $\{(\ol{f}|p),(b|p),(f|b)\}\unabhSymbolOfWrt{\DeltaBird}{\natSpcr}\{(w|b)\}$.
\end{example}

Safe conditional syntax splittings yield a constraint splitting for the case $\Delta_3=\emptyset$ \cite{WilhelmSezginKernIsbernerHaldimannBeierleHeyninck2023JELIA}. The following proposition extends this result to the case $\Delta_3 \neq \emptyset$.
\begin{proposition}
	\label{prop_safe_c}
	Let $\Delta$ be a \BB. Then
	\[
	\safeCondSynSplitGen\ \quad \text{implies} \quad \Delta=\DeltaOneWoThree\constraintCupNatSpcr\DeltaTwoWoThree\constraintCupNatSpcr\DeltaThree.
	\]
\end{proposition}
\begin{proof}
	We refer to the proof for the already established result in \cite{WilhelmSezginKernIsbernerHaldimannBeierleHeyninck2023JELIA} and extend this for the case where $\Delta_3 = \Delta_1\cap\Delta_2\neq\emptyset$.
	That means we only need to show that \safeCondSynSplitGen\ implies $\Delta_3\unabhSymbolOfWrt{\Delta}{\natSpcr}(\Delta_i\setminus\Delta_3)$ for $i\in\{1,2\}$.
	We know that $\Delta_3$ contains only 
	conditionals which cannot be falsified \cite[cf.~Lemma~19]{BeierleSpiegelHaldimannWilhelmHeyninckKernIsberner2024KR}. Therefore, these conditionals cannot appear in any constraints induced by the other conditionals. 
	Let $\delta\in\Delta_3$ be such a conditional. 
    Then clearly $\FSet_\Delta(\delta) = \emptyset$. Additionally, due to Proposition~\ref{lem_safety}, we have $\emptyset\in \VSet_\Delta(\delta)$. Then by applying R2 we obtain $\natSpcr(\VSet_\Delta(\delta))=\emptyset$ and thus, for $i\in\{1,2\}$, $\Delta_3\unabhSymbolOfWrt{\Delta}{\natSpcr}(\Delta_i\setminus\Delta_3)$. 
\end{proof}

A result analogous to Proposition~\ref{prop_safe_c} does not hold for \genSafe\ splittings.

\begin{proposition}
	\label{prop_gsnoconstraint}
	There is a belief base $\Delta$ and
	some 
	$\genSafeCondSynSplitGen$ such that $\Delta=\DeltaOneWoThree\constraintCupNatSpcr\DeltaTwoWoThree\constraintCupNatSpcr\DeltaThree$ is not a constraint splitting of $\Delta$ \wrt\ $\natSpcr$.
\end{proposition}
\begin{proof}
	Let $\DeltaC=\{(c|a),(c|b),(\ol{c}|\top)\}$ be a \BB. Then 
	\(
	\genSafeCondSynSplitMacro{\DeltaC}{\{(c|a),(\ol{c}|\top)\}}{\{(c|b),(\ol{c}|\top)\}}{\{a\}}{\{b\}}{\{c\}}
	\)
	is a generalized safe conditional syntax splitting of $\DeltaC$ but $\Delta=\{(c|a)\}\constraintCupNatSpcr\{(c|b)\}\constraintCupNatSpcr\{(\ol{c}|\top)\}$ is not a constraint splitting of $\DeltaC$ \wrt\ $\natSpcr$ because the reduced $\CSP^\natSpcr(\DeltaC)$ is given by:
	\begin{align*}
		\constraintOfNatspcrI{\DeltaC}{1}: \eta_1&> \min\{\eta_3\}-\min\{0\}\\
		\constraintOfNatspcrI{\DeltaC}{2}: \eta_2&> \min\{\eta_3\}-\min\{0\}\\
		\constraintOfNatspcrI{\DeltaC}{3}: \eta_3&> \min\{0\}-\min\{0\} %
	\end{align*}
\end{proof}

As Proposition~\ref{prop_gsnoconstraint} shows, constraint splittings are not able to capture 
\genSafe\ conditional syntax splittings.
Guided by the generalization from syntax splitting to conditional syntax splitting, we will therefore generalize constraint splittings to conditional constraint splittings in the next section.

\section{Extending Constraint Splitting}
In this section we introduce the notion of conditional constraint splitting as a generalization of constraint splittings
and, in analogy to conditional syntax splitting, we further refine this notion via safe conditional constraint splittings.

\subsection{Conditional Constraint Splitting} 
Conditional constraint splittings allow the subbases of a constraint splitting to also share conditionals. First, we
define the notion of conditional constraint (in)dependence, properly generalizing that of constraint independence given in Definition~\ref{def_constrindep}.

\begin{definition}[conditional constraint independence]
	Let $\Delta$ be a \BB\ and let $\spcr$ be an spcr. Subbases	
	$\Delta_i,\Delta_j\subseteq\Delta$ are \emph{constraint-wise independent in $\Delta$ \wrt\ $\spcr$ given $\DeltaZero\subseteq\Delta$}, 
	written
	$\conditionalIndepGen$,
	if $(\Delta_i\setminus\DeltaZero)\unabhSymbolOfWrtSpcr{\Delta\setminus\DeltaZero}(\Delta_j\setminus\DeltaZero)$.
\end{definition}

The expression $\conditionalIndepGen$ describes a conditional constraint independence of $\Delta_i$ and $\Delta_j$ given $\DeltaZero$.
That means that the conditionals of $\DeltaZero$ are ignored when checking for constraint independence between $\Delta_i$ and $\Delta_j$. 
This is meant to simulate that the \emph{conditional pivot} $\DeltaZero$ is already known or fixed, for example if the constraint variables of $\DeltaZero$ have already been assigned a fixed value.
Note that constraint independence and conditional constraint independence coincide whenever $\Delta_0=\emptyset$.
In terms of constraint dependency graphs, conditional constraint independence given $\Delta_0$  
	can intuitively be checked by ``hiding'' all nodes $\delta_0\in\Delta_0$ from the graph before checking if there is a path between the hyperedges representing $\Delta_i$ and $\Delta_j$ in the constraint dependency graph for $\CSP^\spcr(\Delta)$.
Utilizing this notion yields the following generalization of constraint splittings.

\begin{definition}[conditional constraint splitting]
	\label{def_condconstr}
	Let $\Delta$ be a \BB\ and let $\spcr$ be an spcr.
	A set $\{\Delta_1,\dots,\Delta_m\}$ is a \emph{conditional constraint splitting of $\Delta$ \wrt\ $\spcr$ given $\DeltaZero\subseteq\Delta$}, written
	\[
		\condConstraintSplitGen, 
	\]
	if 
	$\bigcup_{k=1}^{m}\Delta_k = \Delta$ and for all $i,j\in\{1,\dots,m\}$ with $i\neq j$:
	\(
		\conditionalIndepGen.
	\)
\end{definition}

Note that a conditional constraint splitting is not necessarily a partition of the \BB\ anymore but also allows the subbases to share conditionals. 
We illustrate conditional constraint splittings with an example.
\begin{example}
	\label{ex_condconstrsplits}
	Consider again $\DeltaBird=\{\delta_1\!:\!(f|b),\ \delta_2\!:\!(\ol{f}|p),\ \delta_3\!:\!(b|p),\ \delta_4:(w|b)\}$ 
	from Example~\ref{ex_constrsplits}, which
	has a conditional constraint splitting \wrt\ $\natSpcr$ given $\DeltaBird_0=\{(f|b)\}$:
	\[
	\DeltaBird=\{(\ol{f}|p),(f|b)\}\constraintCupNatSpcr\{(b|p),(f|b)\}\constraintCupNatSpcr\{(w|b)\}\mid \{(f|b)\}.
	\]	
	This can be verified by checking the constraints of the subbases $\DeltaBird_i\setminus\DeltaBird_0$ with regard to the \BB\ $\DeltaBird\setminus\DeltaBird_0$,
	resulting in a \CSP ignoring $\DeltaBird_0$.
	\begin{align*}
		\constraintOfNatspcrI{\DeltaBird\setminus\DeltaBird_0}{2}: \eta_2 &> \min\{0\} - \min\{0\}\\
		\constraintOfNatspcrI{\DeltaBird\setminus\DeltaBird_0}{3}: \eta_3 &> \min\{0\} - \min\{0\}\\
		\constraintOfNatspcrI{\DeltaBird\setminus\DeltaBird_0}{4}: \eta_4 &> \min\{0\} - \min\{0\}
	\end{align*}
	Because these constraints and thus the subbases are independent of one another, the subbases are constraint independent given $\{(f|b)\}$. For example, $\conditionalIndepOfInWrtGiven{\{(\ol{f}|p),(f|b)\}}{\{(b|p),(f|b)\}}{\DeltaBird}{\natSpcr}{\{(f|b)\}}$ because $\{(\ol{f}|p)\}\unabhSymbolOfWrt{\DeltaBird\setminus\DeltaBird_0}{\natSpcr} \{(b|p)\}$ but $\{(\ol{f}|p),(f|b)\}\notunabhSymbolOfWrt{\DeltaBird}{\natSpcr}\{(b|p),(f|b)\}$ because the transitivity of $\directlyDependsOfWrt{\DeltaBird}{\natSpcr}$ yields  $(f|b)\directlyDependsOfWrt{\DeltaBird}{\natSpcr}(f|b)$.
	
	Consider next $\DeltaC=\{(c|a),(c|b),(\ol{c}|\top)\}$ from the proof of Proposition~\ref{prop_gsnoconstraint}. Then 
	$\DeltaC=\{(c|a),(\ol{c}|\top)\}\constraintCupNatSpcr\{(c|b),(\ol{c}|\top)\}\mid \{(\ol{c}|\top)\}$.
\end{example}
As seen in Example~\ref{ex_condconstrsplits}, conditional constraint splittings allow us to find more 
splittings for a \BB\ $\Delta$ and also to find constraint splittings for \BBs\ which previously did not admit one at all. 
Next we state a basic fact regarding the conditional pivot $\DeltaZero$. 

\begin{proposition}
	\label{cor_onlyDZero}
	Let $\Delta$ be a \BB\ and let $\spcr$ be an spcr such that
	$\condConstraintSplitGen$.
	For any $\Delta_i,\Delta_j$ with $i,j\in\{1,\dots,m\}, i\neq j$ we have $\Delta_i\cap\Delta_j\subseteq\DeltaZero$
\end{proposition}
\begin{proof}
	Towards a contradiction, assume that there was some
	$\delta\in\Delta_i\cap\Delta_j$ with $\delta\notin\DeltaZero$. Note that $\delta\directlyDependsOfWrt{\Delta\setminus\DeltaZero}{\spcr}\delta$ holds because $\directlyDependsOfWrt{\Delta\setminus\DeltaZero}{\spcr}$ is reflexive by definition. Therefore, $\conditionalIndepGen$ would not hold, which means that $\condConstraintSplitGen$ would not be a conditional constraint splitting.
\end{proof}

Just like conditional syntax splittings coincide with syntax splittings when $\Sigma_3=\emptyset$ \cite{BeierleSpiegelHaldimannWilhelmHeyninckKernIsberner2024KR}, conditional constraint splittings coincide with constraint splittings whenever $\DeltaZero=\emptyset$. 

\begin{propositionrep}
	Let $\Delta$ be a \BB\ and let $\spcr$ be an spcr. Then
	\[
		\constraintSplitGen
		\quad \text{iff} \quad
		\condConstraintSplitMacro{\Delta}{\Delta_1}{\Delta_m}{\emptyset}.
	\]
\end{propositionrep}
\begin{proof}
First, observe that for $\DeltaZero=\emptyset$ the notions $\unabhSymbolOfWrt{\Delta}{\spcr}$ and $\unabhSymbolOfWrt{\Delta\setminus\DeltaZero}{\spcr}$ coincide 
	and $\Delta_i\setminus\DeltaZero=\Delta_i$.
	We still have to show that $\{\Delta_1,\dots,\Delta_m\}$ is a partition iff $\bigcup_{k=1}^{m}\Delta_k$.
	For the direction ``$\Rightarrow$'', if $\{\Delta_1,\dots,\Delta_m\}$ is a partition, then $\bigcup_{k=1}^{m}\Delta_k = \Delta$ is clear. 
	For the direction ``$\Leftarrow$'', we have $\bigcup_{k=1}^{m}\Delta_k = \Delta$.
	With Proposition~\ref{cor_onlyDZero} and the fact that $\DeltaZero=\emptyset$ it must then be the case that $\{\Delta_1,\dots,\Delta_m\}$ is a partition.
\end{proof}

In another similarity to purely syntactical conditional syntax splittings (see \cite[Example 6]{HeyninckKernIsbernerMeyerHaldimannBeierle2023AAAI}), conditional constraint splittings do not guarantee complete independence of the constraints in the subbases as the following example illustrates.

\begin{example}
	\label{exa_unsafesplit}
	Let $\DeltaO = \{\delta_1\!:\!(ab|c),\  \delta_2\!:\!(\ol{a}|\top),\  \delta_3\!:\!(\ol{b}|\top)\}$. $\CSP^\natSpcr(\DeltaO)$ is given by:
	\begin{align*}
		\constraintOfNatspcrI{\DeltaO}{1}:\eta_1 &> \min\{\eta_2+\eta_3\}-\min\{0\}\\
		\constraintOfNatspcrI{\DeltaO}{2}:\eta_2 &> \min\{0\}-\min\{0\}\\
		\constraintOfNatspcrI{\DeltaO}{3}:\eta_3 &> \min\{0\}-\min\{0\}
	\end{align*}
	Thus
	\begin{equation}
		\label{eq_unsafesplit2}
		\DeltaO=\{(ab|c),(\ol{a}|\top)\}\constraintCupNatSpcr\{(ab|c),(\ol{b}|\top)\}\mid\{(ab|c)\},
	\end{equation} 
	which is illustrated in Figure~\ref{subfig_constrSplitsO}.
	Observe that $\CSP^\natSpcr(\DeltaO)\marg_{\DeltaO_1}=\{\constraintOfNatspcrI{\DeltaO}{1},\constraintOfNatspcrI{\DeltaO}{2}\}$ mentions the constraint variable $\eta_3$ whose associated conditional $(\ol{b}|\top)$ is not part of $\DeltaO_1$. Similarly this holds for $\CSP^\natSpcr(\DeltaO)\marg_{\DeltaO_2}=\{\constraintOfNatspcrI{\DeltaO}{1},\constraintOfNatspcrI{\DeltaO}{3}\}$ and $\eta_2$ as well as for $\CSP^\natSpcr(\DeltaO)\marg_{\DeltaO_0}=\{\constraintOfNatspcrI{\DeltaO}{1}\}$ and both $\eta_2$ and $\eta_3$. Thus, none of the sets of constraints $\CSP^\natSpcr(\DeltaO)\marg_{\DeltaO_0}$, $\CSP^\natSpcr(\DeltaO)\marg_{\DeltaO_1}$ and $\CSP^\natSpcr(\DeltaO)\marg_{\DeltaO_2}$ can be solved independently from one another.
\end{example}

Example~\ref{exa_unsafesplit} shows that conditional constraint splittings alone do not guarantee that the constraints on the subbases can be solved independently.
We will therefore refine the notion of constraint splittings in the next section.

\subsection{Safe Conditional Constraint Splitting}
In analogy to (generalized) safe conditional syntax splittings, we introduce the notion of safe conditional constraint splittings.%

\begin{definition}[safe conditional constraint splitting]
	\label{def_safecondconstr}
	Let $\Delta$ be a \BB\ and let $\spcr$ be an spcr.
	A conditional constraint splitting $\condConstraintSplitGen$ of $\Delta$
	is safe, written\footnote{The notation for safe is given as superscript on the pipe symbol.} 
	\[
		\safeCondConstraintSplitGen,
	\]
	if additionally, for all $i\in\{1,\dots, m\}$:
 	$\DeltaZero\not\directlyDependsOfWrtSpcr{\Delta}(\Delta_i\setminus\DeltaZero)$.
\end{definition}

Safe conditional constraint splittings guarantee that constraints of the conditional pivot $\DeltaZero$ may only depend on conditionals in $\DeltaZero$ itself.
Thus they can guarantee independence of the constraints in the subbases given $\DeltaZero$, allowing for iteratively solving $\CSP(\Delta)$ by first computing solutions for $\CSP(\DeltaZero)$ and feeding the results into the constraints of the subbases, which can then be solved independently from one another.

In terms of constraint dependency graphs, safe conditional constraint splittings ensure that each hyperedge induced by $\Delta_i$ may be connected only to itself and/or $\Delta_0$. In particular, the hyperedge induced by the conditional pivot $\Delta_0$ 
is a self-contained ``dead end'' in the constraint dependency graph. 
An example of a safe (Fig.~\ref{subfig_constrSplitsBird}) and a not safe (Fig.~\ref{subfig_constrSplitsO}) conditional constraint splitting can be seen in Figure~\ref{fig_constrSplits}.

\begin{example}
	\label{ex_splitsolutions}
	The splitting
	\begin{equation}
		\label{eq_birdsafe}
		\DeltaBird=\{(\ol{f}|p),(f|b)\}\constraintCupNatSpcr\{(b|p),(f|b)\}\constraintCupNatSpcr\{(w|b)\}\safeMid \{(f|b)\}
	\end{equation}
	from Example~\ref{ex_condconstrsplits} is a safe conditional constraint splitting of $\DeltaBird$ (cf. Figure~\ref{subfig_constrSplitsBird}). For this splitting it is possible to solve $\constraintOfNatspcrI{\DeltaBird}{1}$ %
	first, \eg, by setting $\eta_1=1$, as this constraint is not dependent on any other constraints. Utilizing $\eta_1=1$ we can then solve $\constraintOfNatspcrI{\DeltaBird}{2}$, $\constraintOfNatspcrI{\DeltaBird}{3}$  and $\constraintOfNatspcrI{\DeltaBird}{4}$ 
	\begin{align*}
		\constraintOfNatspcrI{\DeltaBird}{2}: \eta_2 &> \min\{1\} - \min\{0\}\\
		\constraintOfNatspcrI{\DeltaBird}{3}: \eta_3 &> \min\{1\} - \min\{0\}\\
		\constraintOfNatspcrI{\DeltaBird}{4}: \eta_4 &> \min\{0\} - \min\{0\}
	\end{align*}
	independently from one another because they do not mention any of the other constraint variables.
	Thus we can obtain a solution $\kappaiminusVektor{}\in\solutionsR$ with, \eg, $\kappaiminusVektor{}=(1,2,2,1)$ by solving $\constraintOfNatspcrI{\DeltaBird}{2}$, $\constraintOfNatspcrI{\DeltaBird}{3}$  and $\constraintOfNatspcrI{\DeltaBird}{4}$ independently. 
	Note that it would also be possible to choose $\DeltaBird_0=\{(f|b),(w|b)\}$ because $\constraintOfNatspcrI{\DeltaBird}{4}$ also does not mention any other constraint variable.
	Utilizing Figure~\ref{subfig_constrSplitsO}, we can also see that the conditional constraint splitting of $\DeltaO$ (cf. Example~\ref{exa_unsafesplit}) given by
	\begin{equation*}
		\DeltaO=\{(ab|c)\}\constraintCupNatSpcr\{(\ol{a}|\top),(\ol{b}|\top)\}\safeMid\{(\ol{a}|\top),(\ol{b}|\top)\}
	\end{equation*}
	is safe.

	\begin{figure}
	\begin{subfigure}{.48\textwidth}
		\centering
		\begin{tikzpicture}
			\centering
			
			\node (1) at (0,0) {$\delta_1$};
			\node (2) at (2,0) {$\delta_2$};
			\node (3) at (0,-2) {$\delta_3$};
			\node (4) at (2,-2) {$\delta_4$};
			
			\draw[rounded corners=5pt] (1.north west) -- (2.north east) -- (2.south east) -- (1.south west) -- cycle;
			\draw[rounded corners=5pt] (1.north west) -- (1.north east) -- (3.south east) -- (3.south west) -- cycle;
			
			\draw (4) circle [radius=0.3];
			
			\draw[->] (2) to (1);
			\draw[->] (3) to (1);
		\end{tikzpicture}
		\caption{Splitting \eqref{eq_birdsafe} of $\DeltaBird$ from Example \ref{ex_splitsolutions}.}
		\label{subfig_constrSplitsBird}
	\end{subfigure}
	\begin{subfigure}{.48\textwidth}
		\centering
		\begin{tikzpicture}
			\centering
			
			\node (1) at (0,0) {$\delta_1$};
			\node (2) at (2,0) {$\delta_2$};
			\node (3) at (0,-2) {$\delta_3$};
			
			\draw[rounded corners=5pt] (1.north west) -- (2.north east) -- (2.south east) -- (1.south west) -- cycle;
			\draw[rounded corners=5pt] (1.north west) -- (1.north east) -- (3.south east) -- (3.south west) -- cycle;

			\draw[->] (1) to (2);
			\draw[->] (1) to (3);
		\end{tikzpicture}
		\caption{Splitting \eqref{eq_unsafesplit2} of $\DeltaO$ from Example~\ref{exa_unsafesplit}.}
		\label{subfig_constrSplitsO}
	\end{subfigure}
	\caption{Visualization of conditional constraint splittings in Examples \ref{exa_unsafesplit} and \ref{ex_splitsolutions}. An edge $\delta_i\rightarrow\delta_j$ represents that $\delta_i$ directly depends on $\delta_j$ in $\DeltaBird$ (subfig. (a)) and in $\DeltaO$ (subfig. (b)) respectively \wrt\ $\natSpcr$.
	A hyperedge $\{\delta_i,\dots,\delta_n\}$  
	represents a subbase in a conditional constraint splitting, where subfigure (a) shows the splitting \eqref{eq_birdsafe} and subfigure (b) shows the splitting \eqref{eq_unsafesplit2}.}
	\label{fig_constrSplits}
\end{figure} \end{example}

The following proposition presents an important characterization of safe conditional constraint splittings in terms of independence of subbases excluding $\DeltaZero$.

\begin{propositionrep}
	\label{prop_safeAlt}
	Let $\Delta$ be a \BB and let $\spcr$ be an spcr.
	\[
	\safeCondConstraintSplitGen
	\quad \text{iff} \quad
	\text{for all } i,j\in\{0,\dots,m\}, i\neq j: \Delta_i\not\directlyDependsOfWrtSpcr{\Delta}(\Delta_j\setminus\DeltaZero).
	\]
\end{propositionrep}
\begin{proof}
	$\Rightarrow:$ $\safeCondConstraintSplitGen$ implies
	$(\Delta_i\setminus\DeltaZero)\not\directlyDependsOfWrtSpcr{\Delta\setminus\DeltaZero}(\Delta_j\setminus\DeltaZero)$.
	Then either 
	$(\Delta_i\setminus\DeltaZero)\not\directlyDependsOfWrtSpcr{\Delta}(\Delta_j\setminus\DeltaZero)$
	or there is some $\DeltaZero'\subseteq\DeltaZero$ such that $\Delta_i\directlyDependsOfWrtSpcr{\Delta}\DeltaZero'$ and $\DeltaZero'\directlyDependsOfWrtSpcr{\Delta}\Delta_j$.
	However, because the splitting is safe $\DeltaZero\not\directlyDependsOfWrtSpcr{\Delta}(\Delta_j\setminus\DeltaZero)$
	and thus
	$\DeltaZero'\not\directlyDependsOfWrtSpcr{\Delta}(\Delta_j\setminus\DeltaZero)$.
	In both cases
	$\Delta_i\not\directlyDependsOfWrtSpcr{\Delta}(\Delta_j\setminus\DeltaZero)$.
	
	$\Leftarrow:$ Assume $\Delta_i\not\directlyDependsOfWrtSpcr{\Delta}(\Delta_j\setminus\DeltaZero)$. Then clearly $(\Delta_i\setminus\DeltaZero)\not\directlyDependsOfWrtSpcr{\Delta\setminus\DeltaZero}(\Delta_j\setminus\DeltaZero)$,
	and in particular also $\Delta_0\not\directlyDependsOfWrtSpcr{\Delta}(\Delta_j\setminus\DeltaZero)$.
\end{proof}

Now we
show that every \genSafe\ conditional syntax splitting yields a safe conditional constraint splitting \wrt\ the spcr given by $\spcrSubset=\{R1,R2\}$, 
closing the gap put forth by Proposition~\ref{prop_gsnoconstraint}.

\begin{proposition}
	\label{prop_gensafe_safecondconstr}
	Let $\Delta$ be a \BB. Then
	\[
		\genSafeCondSynSplitGen\quad\text{implies}\quad \Delta=\DeltaOne\constraintCupWrt{\spcrSubset}\DeltaTwo\safeMid(\Delta_1\cap\Delta_2). 
	\]
\end{proposition}
\begin{proof}
 	Let \genSafeCondSynSplitGen\ and let $\DeltaZero=\Delta_1\cap\Delta_2$. First we show that $\Delta=\DeltaOne\constraintCupWrt{\spcrSubset}\DeltaTwo\safeMid\DeltaZero$ is a conditional constraint splitting \wrt\ $\spcrSubset$, \ie,
 	$\conditionalIndepGenNrWrt{\spcrSubset}$. 
 	Towards a contradiction, assume
 	$\conditionalIndepGenNrWrt{\spcrSubset}$ 
 	were not the case. Then either 
 	$(\Delta_1\setminus\DeltaZero)\directlyDependsOfWrt{\Delta\setminus\DeltaZero}{\spcrSubset}(\Delta_2\setminus\DeltaZero)$ or $(\Delta_2\setminus\DeltaZero)\directlyDependsOfWrt{\Delta\setminus\DeltaZero}{\spcrSubset}(\Delta_1\setminus\DeltaZero)$. 
 	We consider the first case here, the second case is symmetrical. This means there must be some $\delta_1\in\Delta_1\setminus\DeltaZero$ such that there is a $\delta_2\in\Delta_2\setminus\DeltaZero$ with $\delta_1\directlyDependsOfWrt{\Delta\setminus\DeltaZero}{\spcrSubset}\delta_2$.
 	Note that $\delta_2\notin\DeltaZero$ implies with Corollary~\ref{cor_onlyDZero} that $\delta_2\notin\Delta_1$.
 	This means there is some set $S$ with $\delta_2\in S$ and either $S\in\VSet_{\Delta}(\delta_1)$ or $S\in\FSet_{\Delta}(\delta_1)$. Again we assume the first case as the second is analogous. 
 	Because $\delta_2\in\Delta_2\setminus\DeltaZero$, the generalized safety property implies that there must be a set $S'$ with $S'\subsetneq S$ and $\delta_2\notin S'$.
 	This can be seen by the following: Let $\omega=\omega^1\omega^2\omega^3$ be such that $\omega$ verifies $\delta_1$ and $\{(B|A)\in\Delta\mid \omega\models A\ol{B}\}=S$, that is, $\omega$ is ``responsible for'' $S$. Then $\omega^1\omega^3$ sufficiently describes the falsification behaviour \wrt\ $\Delta_1$, \ie, $\{(B|A)\in\Delta_1\mid \omega\models A\ol{B}\}=\{(B|A)\in\Delta_1\mid \omega^1\omega^3\models A\ol{B}\}$. With safety there is then ${\omega^2}'$ such that $\omega^1{\omega^2}'\omega^3$ does not falsify $\delta_2$ (and in fact no other conditional from $\Delta_2\setminus\Delta_0$). 
 	Thus $R1$ (or $R2$ in the second case) from $\spcrSubset$ would remove this set $S$ from $\VSet_{\Delta}(\delta_1)$. In this way $\delta_2$ is eliminated from $\VSet_{\Delta}(\delta_1)$ and therefore $\delta_1\not\directlyDependsOfWrt{\Delta\setminus\DeltaZero}{\spcrSubset}\delta_2$, contradiction.
 	To show that $\DeltaOne\constraintCupWrt{\spcrSubset}\DeltaTwo\safeMid\DeltaZero$ is safe we can apply the same arguments as above, as the generalized safety condition demands that the minimal constraint inducing sets for conditionals in $\DeltaZero$ can only include conditionals in $\DeltaZero$ itself. 
\end{proof}
Note that in the case where $\Delta_1\cap\Delta_2=\emptyset$, Proposition~\ref{prop_gensafe_safecondconstr} and Proposition~\ref{prop_safe_c} coincide. 

	Proposition~\ref{prop_gensafe_safecondconstr} extends also to the spcr $\natSpcr$, as every conditional constraint splitting of a \BB\ $\Delta$ \wrt\ $\spcrSubset$ is also a conditional constraint splitting of $\Delta$ \wrt\ $\natSpcr$.
	\begin{proposition}
		Let $\Delta$ be a belief base, let $\spcrSubset=\{R1,R2\}$ and let $\natSpcr=\{R1,\dots,R6\}$. Then
		\[
			\conditionalConstrainSplitGenWrt{\spcrSubset}\quad\text{implies}\quad\conditionalConstrainSplitGenWrt{\natSpcr}.
		\]
	\end{proposition}
	\begin{proof}
		Since every rule $R1,\dots,R6$ only ever removes constraint variables, but never adds them, the additional application of rules cannot undo already existing independencies.
	\end{proof}

Definition~\ref{def_safecondconstr} allows that, for a given splitting $\safeCondConstraintSplitGen$, 
the conditional pivot $\Delta_0$ may not occur in every subbase that depends on it, as the following Example shows.
\begin{example}
	\label{exa_nicesafe1}
	Let $\DeltaBird$ be as from Example~\ref{ex_constraints}. Then the following safe conditional constraint splittings of $\DeltaBird$ exist:	
	\begin{align}
		\DeltaBird&=\{(\ol{f}|p),(f|b)\}\constraintCupNatSpcr\{(b|p),(f|b)\}\constraintCupNatSpcr\{(w|b)\}\safeMid \{(f|b)\}\label{eq_nicesafe1}\\
		\DeltaBird&=\{(\ol{f}|p),(f|b)\}\constraintCupNatSpcr\{(b|p)\}\constraintCupNatSpcr\{(w|b)\}\safeMid \{(f|b)\}\label{eq_nicesafe2}\\
		\DeltaBird&=\{(\ol{f}|p)\}\constraintCupNatSpcr\{(b|p),(f|b)\}\constraintCupNatSpcr\{(w|b)\}\safeMid \{(f|b)\}\label{eq_nicesafe3}\\
		\DeltaBird&=\{(\ol{f}|p),(f|b)\}\constraintCupNatSpcr\{(b|p),(f|b)\}\constraintCupNatSpcr\{(w|b),(f|b)\}\safeMid \{(f|b)\}\label{eq_nicesafe4}
	\end{align}
	Note that these four splittings describe the same independencies and thus imply each other. However, taking a closer look at $\CSP^\natSpcr(\DeltaBird)$ we see that the subbase $\{(w|b)\}$ is actually not dependent on $\DeltaBird_0 = \{(f|b)\}$. 	
	It is not derivable from any of the splittings \eqref{eq_nicesafe1} to \eqref{eq_nicesafe4} that this is the case. 
	Furthermore, the subbase $\{(b|p)\}$ in splitting \eqref{eq_nicesafe2} is dependent on $\DeltaBird_0$, yet $\DeltaBird_0$ is not included in it. 
\end{example}

Ideally, every subbase $\Delta_i$ should already contain all conditionals relevant to $\CSP(\Delta_i)$, without having to take $\Delta_0$ into account implicitly. We therefore introduce the following notion of \niceSafe\ splittings.

	\begin{definition}[$\niceSafeMid$, \niceSafe\ conditional constraint splitting]
		\label{def_niceSafe}
		Let $\Delta$ be a \BB\ and let $\spcr$ be an spcr.
		A conditional constraint splitting $\condConstraintSplitGen$
		is called \emph{\niceSafe}, written
		\begin{equation}
			\niceSafeCondConstraintSplitGen,
		\end{equation}
		if, for all $i\in\{0,\dots, m\}$:
		$\Delta_i\not\directlyDependsOfWrtSpcr{\Delta}(\Delta\setminus\Delta_i)$.
	\end{definition}

Every \niceSafe\ splitting is also safe which can easily be seen via Proposition~\ref{prop_safeAlt}. 
	
\begin{corollary}
	$\niceSafeCondConstraintSplitGen$ implies $\safeCondConstraintSplitGen$
\end{corollary}
\begin{proof}
	Follows directly from Proposition~\ref{prop_safeAlt}.
\end{proof}

We give an example of \niceSafe\ splittings. 
\begin{example}
	Among the splittings \eqref{eq_nicesafe1} to \eqref{eq_nicesafe4} from Example~\ref{exa_nicesafe1}, only \eqref{eq_nicesafe1} and \eqref{eq_nicesafe4} are \niceSafe\ splittings. Note that  $\DeltaBird=\{(\ol{f}|p),(f|b)\}\constraintCupNatSpcr\{(b|p),(f|b)\}\constraintCupNatSpcr\{(w|b)\}\niceSafeMid \{(f|b)\}$ implies $\{(w|b)\}\not\directlyDependsOfWrtSpcr{\DeltaBird}\DeltaBird_0$ due to $\DeltaBird_0\not\subseteq \{(w|b)\}$ which is not derivable from \eqref{eq_nicesafe1}, \eqref{eq_nicesafe2}, or \eqref{eq_nicesafe3}. %
\end{example}

Every safe conditional constraint splitting yields a \niceSafe\ conditional constraint splitting by taking the conditionals of the conditional pivot $\Delta_0$ into account in all subbases.
	\begin{lemma}
		\label{lemm_includeDZero}
		Let $\Delta$ be a \BB\ and let $\spcr$ be an spcr.
		Then
		\[
			\safeCondConstraintSplitGen\quad\text{implies}\quad \niceSafeCondConstraintSplitMacro{\Delta}{(\Delta_1\cup\Delta_0)}{(\Delta_m\cup\Delta_0)}{\Delta_0}.
		\]
	\end{lemma}
	\begin{proof}
		Let $i,j\in\{1,\dots,m\}, i\neq j$. 
		$\Delta_0\not\directlyDependsOfWrtSpcr{\Delta}(\Delta\setminus\DeltaZero)$ holds by the fact that $\Delta_0$ is unchanged.
		Since $\safeCondConstraintSplitGen$, \(\Delta_i\not\directlyDependsOfWrtSpcr{\Delta}(\Delta_j\setminus\DeltaZero)\) follows from Proposition~\ref{prop_safeAlt}. By  Definition~\ref{def_safecondconstr} $\DeltaZero\not\directlyDependsOfWrtSpcr{\Delta}(\Delta_j\setminus\DeltaZero)$. Thus \(\Delta_i\cup\Delta_0\not\directlyDependsOfWrtSpcr{\Delta}(\Delta_j\setminus\DeltaZero)\), in particular \(\Delta_i\cup\Delta_0\not\directlyDependsOfWrtSpcr{\Delta}(\Delta_j\setminus(\Delta_i\cup\Delta_0))\) and thus  \(\Delta_i\cup\Delta_0\not\directlyDependsOfWrtSpcr{\Delta}(\Delta\setminus(\Delta_i\cup\Delta_0))\), \ie, $\safeCondConstraintSplitMacro{\Delta}{(\Delta_1\cup\Delta_0)}{(\Delta_m\cup\Delta_0)}{\Delta_0}$.
	\end{proof}

Lemma~\ref{lemm_includeDZero} shows that \niceSafe\ splittings mostly act as an additional constraint ensuring that each subbase is self-contained without having to take $\Delta_0$ into account separately, simplifying notation and many proofs in this article. Nevertheless, \niceSafe\ splittings can make explicit that a subbase is not dependent on the conditional pivot $\Delta_0$.

	For $\niceSafeCondConstraintSplitGen$, Definition~\ref{def_niceSafe} ensures that each subbase $\Delta_i$ is self-contained, \ie, all conditionals that $\Delta_i$ depends on are included in $\Delta_i$ itself.
	\begin{proposition}
		\label{prop_closedunderdepends}
		Let $\Delta$ be a \BB\ and let $\spcr$ be an spcr such that $\niceSafeCondConstraintSplitGen$. Then, for $i\in\{0,\dots,n\}$, each $\Delta_i$ is closed under $\directlyDependsOfWrtSpcr{\Delta}$. 
	\end{proposition}
	\begin{proof}
		Follows directly from Definition~\ref{def_niceSafe}.
	\end{proof}

We can now strengthen Proposition~\ref{prop_gensafe_safecondconstr} by showing that the conditional constraint splitting obtained from a generalized safe conditional syntax splitting is not only safe, but \niceSafe.
\begin{proposition}
	\label{prop_gensafe_nicesafecondconstr}
	Let $\Delta$ be a \BB. Then
	\[
		\genSafeCondSynSplitGen\quad\text{implies}\quad \Delta=\DeltaOne\constraintCupWrt{\spcrSubset}\DeltaTwo\niceSafeMid(\Delta_1\cap\Delta_2).
	\]
\end{proposition}
\begin{proof}
	Observe that $\Delta_0=(\Delta_1\cap\Delta_2)\subseteq\Delta_1 and \Delta_0=(\Delta_1\cap\Delta_2)\subseteq\Delta_2$, thus $\Delta_1=\Delta_1\cup\Delta_0$ and $\Delta_2=\Delta_2\cup\Delta_0$ and therefore Proposition~\ref{prop_gensafe_safecondconstr} and Lemma~\ref{lemm_includeDZero} imply $\Delta=\DeltaOne\constraintCupWrt{\spcrSubset}\DeltaTwo\niceSafeMid(\Delta_1\cap\Delta_2)$.
\end{proof}

The key takeaway in Proposition~\ref{prop_gensafe_nicesafecondconstr} is that a belief base that generalized safely splits into $\Delta_1$ and $\Delta_2$ syntactically, also can be split \niceSafe\ into $\Delta_1$ and $\Delta_2$ along its constraints.

\section{Safe Conditional c-Solution Splitting}
In this section, we lift the notion of constraint splittings to the semantical level by introducing the notion of \gcondCSolSplit.
To keep notation concise and to avoid being distracted by technical particularities we investigate only binary splittings in this section.

In previous work, non-conditional semantic splittings for c-representations were characterized by the notion of c-solution splittings  \cite{BeierleHaldimannKernIsberner2021Boeger75Festschrift}.
\begin{definition}[c-solution splitting \cite{BeierleHaldimannKernIsberner2021Boeger75Festschrift}]
	\label{def_solsplit}
	Let $\Delta$ be a \BB. A partition $\{\Delta_1,\Delta_2\}$ of $\Delta$ is a c-solution splitting for $\Delta$ if the following equation holds:
	\begin{equation}
		\label{eq_solsplit}
		\solutionsR=\{(\kappaiminusVektor^1,\kappaiminusVektor^2)\mid \kappaiminusVektor^1\in\solutionsRof{\Delta_1}, \kappaiminusVektor^2\in\solutionsRof{\Delta_2}\}
	\end{equation}
\end{definition}
Definition~\ref{def_solsplit} means that
the solutions of $\CSP(\Delta)$ are exactly the concatenations of the solutions of $\CSP(\Delta_1)$ and $\CSP(\Delta_2)$. A straightforward adaptation of c-solution splitting to the conditional case fails however, because the solutions of the subbases might not match on their intersection as shown in the following example.

\begin{example}
	\label{ex_nocondsem}
	Recall $\DeltaBird=\{\delta_1\!:\!(f|b),\ \delta_2\!:\!(\ol{f}|p),\ \delta_3\!:\!(b|p),\ \delta_4:(w|b)\}$ from Example~\ref{exa_natspcr} with its associated simplified constraints. Observe that
	$\DeltaBird=\{(f|b), (\ol{f}|p), (b|p)\}\constraintCupNatSpcr\{(f|b), (w|b)\}\safeMid\{(f|b)\}$.
	Then we have
	$(1,2,2)\in\solutionsRofNatSpcr{\DeltaBird_1}$, $(3,1)\in\solutionsRofNatSpcr{\DeltaBird_2}$ and $(1)\in\solutionsRofNatSpcr{\DeltaBird_1\cap\DeltaBird_2}$. 
	But these cannot be combined 
	into a solution for $\CSP^\natSpcr(\DeltaBird)$;
	for example $(3,2,2,1)\notin\solutionsRofNatSpcr{\DeltaBird}$ as both $\constraintOfNatspcrI{\DeltaBird}{2}$ and $\constraintOfNatspcrI{\DeltaBird}{3}$ are violated by this solution.
\end{example}

Note that Example~\ref{ex_nocondsem} exploits the freedom of choosing arbitrarily high impacts for conditionals. Indeed, there is no real reason to select $(3,1)\in\solutionsRof{\DeltaTwo}$ as opposed to the smaller vector $(1,1)\in\solutionsRof{\DeltaTwo}$ which also matches the other impacts. 
The crucial point here is that the constraints induced by the intersection of the subbases may be solved differently in the respective \CSPs.
It is therefore reasonable to require that this intersection should be ``treated in the same way'' in all subbases. 
This requirement then implies that the intersection $\Delta_1\cap\Delta_2$ plays the role of the conditional pivot for the solution splitting.
This is in analogy to inference with conditional syntax splitting where it is presupposed that full information on the conditional pivot $\Sigma_3$ is available (cf. \cite{HeyninckKernIsbernerMeyerHaldimannBeierle2023AAAI,SpiegelHaldimannHeyninckKernIsbernerBeierle2025IJCAI}); 
this is lifted to the semantic level here by demanding that 
the solutions of $\CSP(\Delta_1\cap\Delta_2)$ 
are available and 
can be matched with.
By setting $\Delta_0=\Delta_1\cap\Delta_2$ this is also in line with the
other notions of conditional splittings,
leading to the notion of 
safe conditional c-solution splittings.

\begin{definition}[\gcondCSolSplit]
	\label{def_condCSolSplit}
	Let $\Delta$ be a belief base and let $\Delta_1,\Delta_2\subseteq\Delta$.
	Then $\{\Delta_1,\Delta_2\}$ is a \emph{\gcondCSolSplit} of $\Delta$ %
	if $\Delta=\Delta_1\cup\Delta_2$ and
	\begin{equation}
		\label{eq_csolcover}
		\solutionsR =
		\{\kappaiminusVektor\mid \kappaiminusVektor\marg_{\Delta_1}
		\in\solutionsRof{\Delta_1},%
		\kappaiminusVektor\marg_{\Delta_2}%
		\in\solutionsRof{\Delta_2}\}
	\end{equation}
\end{definition}

Safe conditional c-solution splittings generalize solutions splittings by no longer requiring a partition of $\Delta$, but only subbases of $\Delta$ that yield $\Delta$ when combined;
while $\kappaiminusVektor^1$ and $\kappaiminusVektor^2$ in \eqref{eq_solsplit} are defined over disjoint subbases, an implication of \eqref{eq_csolcover} is that $\kappaiminusVektor^1$ and $\kappaiminusVektor^2$ must coincide on the intersection of $\Delta_1$ and $\Delta_2$, \ie,
$\kappaiminusVektor^1\marg_{\Delta_1\cap\Delta_2}=\kappaiminusVektor^2\marg_{\Delta_1\cap\Delta_2}$.
Thus, \gcondCSolSplit s still demand that any solution for $\CSP(\Delta)$ can be derived by a combination of solutions of $\DeltaOne$ and $\DeltaTwo$, but with the additional stipulation that the solutions match with respect to the intersection $\CSP(\Delta_1\cap\Delta_2)$. Conversely, if there are solutions for $\CSP(\DeltaOne)$ and $\CSP(\DeltaTwo)$ that match on the intersection, then the combination of these vectors must yield a solution for $\CSP(\Delta)$.
Note that \gcondCSolSplit s are not dependent on an spcr. They are defined purely on the semantic level, taking into account only the solution space but not the structure of the underlying constraint system.

\begin{example}
	For $\DeltaBird$ from Example~\ref{ex_nocondsem}, 
	$\{\DeltaBird_1,\DeltaBird_2\}=\{\{\delta_1,\delta_2,\delta_3\},\{\delta_1,\delta_4\}\}=
	\{\{(f|b), (\ol{f}|p), (b|p)\},\{(f|b), (w|b)\}\}$ is a safe conditional solution splitting of $\DeltaBird$. For example, $(1,2,2,1)\in\solutionsOf{\DeltaBird}$ implies $(1,2,2)\in\solutionsRof{\DeltaBird_1}$ and $(1,1)\in\solutionsRof{\DeltaBird_2}$. Conversely, $(2,3,3)\in\solutionsRof{\DeltaBird_1}$ and $(2,1)\in\solutionsRof{\DeltaBird_2}$ implies $(2,3,3,1)\in\solutionsRof{\DeltaBird}$ because the two vectors match on $\DeltaBird_1\cap\DeltaBird_2=\{(f|b)\}$.
	But $(1,2,2)\in\solutionsRof{\DeltaBird_1}$ and $(3,1)\in\solutionsRof{\DeltaBird_2}$ does not imply $(3,2,2,1)\in\solutionsRof{\DeltaBird}$ because the solutions don't match on $\DeltaBird_1\cap\DeltaBird_2$.
\end{example}

We now compare \gcondCSolSplit s to other splitting notions. First we state the following straightforward result.

\begin{proposition}
	Every \cSolSplit\ is a \gcondCSolSplit.
\end{proposition}
\begin{proof}
	Let $\{\Delta_1,\Delta_2\}$ be a \cSolSplit\ of a \BB\ $\Delta$. Then
	\eqref{eq_solsplit} holds.
	Since $\{\Delta_1,\Delta_2\}$ is a partition, \eqref{eq_solsplit} implies \eqref{eq_csolcover} and thus $\{\Delta_1,\Delta_2\}$ is also a \gcondCSolSplit.
\end{proof}

Somewhat surprisingly, safe conditional constraint splittings in general are not safe conditional solution splittings.

\begin{propositionrep}
	\label{prop_constrnosol}
	There is a \BB\ $\Delta$ and an spcr $\spcr$ such that there is $\niceSafeCondConstraintSplitMacroTwo{\Delta}{\Delta_1}{\Delta_2}{\DeltaZero}$ such that $\{\Delta_1,\Delta_2\}$ is not a \gcondCSolSplit\ of $\Delta$.
\end{propositionrep}
\begin{proof}
	$\DeltaBird$ from Example~\ref{ex_constraints} 
	has the safe conditional constraint splitting 
	\begin{equation}
		\label{eq_constrnosol}
		\DeltaBird=\{(\ol{f}|p),(f|b)\}\constraintCupNatSpcr\{(b|p),(f|b),(w|b)\}\niceSafeMid \{(f|b)\}.
	\end{equation}
	Then $\{\DeltaBird_1=\{(f|b),(\ol{f}|p)\},\DeltaBird_2=\{(f|b),(b|p),(w|b)\}\}$ is no safe conditional solution splitting. 
	For this, recall $\CSP^\natSpcr(\DeltaBird)$ from Example~\ref{exa_natspcr} given by
	\begin{align*}
		\constraintOfNatspcrI{\DeltaBird}{1}: \eta_1 &> \min\{0\} - \min\{0\}\\
		\constraintOfNatspcrI{\DeltaBird}{2}: \eta_2 &> \min\{\eta_1\} - \min\{0\}\\
		\constraintOfNatspcrI{\DeltaBird}{3}: \eta_3 &> \min\{\eta_1\} - \min\{0\}\\
		\constraintOfNatspcrI{\DeltaBird}{4}: \eta_4 &> \min\{0\} - \min\{0\}.
	\end{align*}
	$\CSP^\natSpcr(\DeltaBird_1)$ and $\CSP^\natSpcr(\DeltaBird_2)$ are given by
		\begin{align*}
		\constraintOfNatspcrDI{\DeltaBird_1}{\delta_1}: \eta_1^1 &> \min\{0\} - \min\{0\}\\
		\constraintOfNatspcrDI{\DeltaBird_1}{\delta_2}: \eta_2^1 &> \min\{0\} - \min\{0\}\\
		\constraintOfNatspcrDI{\DeltaBird_2}{\delta_1}: \eta_1^2 &> \min\{0\} - \min\{0\}\\
		\constraintOfNatspcrDI{\DeltaBird_2}{\delta_3}: \eta_3^2 &> \min\{0\} - \min\{0\}\\
		\constraintOfNatspcrDI{\DeltaBird_2}{\delta_4}: \eta_4^2 &> \min\{0\} - \min\{0\}.
	\end{align*}
	We observe that $(1,1)\in\solutionsRof{\DeltaBird_1}$ and $(1,1,1)\in\solutionsRof{\DeltaBird_2}$ match on $\DeltaBird_1\cap\DeltaBird_2$ but $(1,1,1,1)\notin\solutionsRof{\DeltaBird}$.
\end{proof}

\begin{proposition}
	\label{prop_subconstraintssplitimplication}
	Let $\Delta$ be a belief base and let $\spcr$ be an spcr such that
	$\niceSafeCondConstraintSplitTwoWrt{\spcr}$. 
	If, for $i\in\{1,2\}$, $\CSP^{\spcr}(\Delta)|_{\Delta_i}=\CSP^{\spcr}(\Delta_i)$ then $\{\Delta_1,\Delta_2\}$ is a safe conditional solution splitting of $\Delta$.
\end{proposition}
\begin{proof}
	We show both set inclusions, \ie, we show (1.) $\kappaiminusVektor\in\solutionsOf{\CSP^{\spcr}(\Delta)}$ implies $\kappaiminusVektor\marg_{\Delta_i}\in\solutionsOf{\CSP^{\spcr}(\Delta_i)}$ for $i\in\{1,2\}$ and (2.) If $\kappaiminusVektor\marg_{\Delta_i}\in\solutionsOf{\CSP^{\spcr}(\Delta_i)}$ for all $i\in\{1,2\}$ then $\kappaiminusVektor\in\solutionsOf{\CSP^{\spcr}(\Delta)}$.
	
	Let $i\in\{1,2\}$. For (1.) observe that $\kappaiminusVektor\in\solutionsOf{\CSP^{\spcr}(\Delta)}$ implies $\kappaiminusVektor\marg_{\Delta_i}\in\solutionsOf{\CSP^{\spcr}(\Delta)\marg_{\Delta_i}}$ and thus per assumption $\kappaiminusVektor\marg_{\Delta_i}\in\solutionsOf{\CSP^{\spcr}(\Delta_i)}$.

	For (2.), assume $\kappaiminusVektor\marg_{\Delta_i}\in\solutionsOf{\CSP^{\spcr}(\Delta_i)}$. Towards a contradiction, assume $\kappaiminusVektor\notin\solutionsOf{\CSP^{\spcr}(\Delta)}$. Then there is $\constraintOfSpcrI{\Delta}{k}\in \CSP^{\spcr}(\Delta)$ that is violated by $\kappaiminusVektor$. Let $\delta_k\in\Delta_i$. Then  $\constraintOfSpcrI{\Delta}{k}\in \CSP^{\spcr}(\Delta)\marg_{\Delta_i}$. Thus, per assumption $\constraintOfSpcrI{\Delta}{k}\in \CSP^{\spcr}(\Delta_i)$, \ie, $\constraintOfSpcrI{\Delta}{k}=\constraintOfSpcrI{\Delta_i}{k}$. 
	Thus, $\kappaiminusVektor$, and also $\kappaiminusVektor\marg_{\Delta_i}$, must violate $\constraintOfSpcrI{\Delta_i}{k}$, which is in contradiction to $\kappaiminusVektor\marg_{\Delta_i}\in\solutionsOf{\CSP^{\spcr}(\Delta_i)}$.
\end{proof}

However, 
there is an spcr
such that the following implication holds.

\begin{propositionrep}	
	\label{prop_idsol}
	Let $\Delta$ be a \BB\ and let $\spcr$ be given by the identity function.
	If $\niceSafeCondConstraintSplitMacroTwoWrt{\Delta}{\Delta_1}{\Delta_2}{\DeltaZero}{\idFct}$,
	then $\{\Delta_1,\Delta_2\}$ is a \gcondCSolSplit\ of $\Delta$.
\end{propositionrep}
\begin{proof}
	Let $\niceSafeCondConstraintSplitMacroTwoWrt{\Delta}{\Delta_1}{\Delta_2}{\DeltaZero}{\idFct}$.
	Per definition $\Delta_i\not\directlyDependsOfWrt{\Delta}{\idFct}(\Delta\setminus\Delta_i)$. 
	This means that 
	$\CSP(\Delta_i)$ does
	not mention any constraint variable for conditionals in $\Delta\setminus\Delta_i$.
	Since no spcr is applied to $\CSP(\Delta)$ nor $\CSP(\Delta_i)$, 
	the constraints in $\Delta$ and $\Delta_i$ regarding the conditionals of $\Delta_i$ then are exactly the same, \ie, 
	$\CSP(\Delta)\marg_{\Delta_i}=\CSP(\Delta_i)$. Thus, with Proposition~\ref{prop_subconstraintssplitimplication}, $\{\Delta_1,\Delta_2\}$ is a \gcondCSolSplit\ of $\Delta$.
\end{proof}

Proposition~\ref{prop_constrnosol} reveals an important distinction to be made between solution and constraint splittings. Safe conditional constraint splittings regard the constraint system built upon the entire \BB\ and delegate only the solving process to the subbases. Safe conditional c-solution splittings demand that the constraint systems themselves can be built and solved directly on the subbases, without taking the other subbases into account at all.
Thus, the deciding difference between the two splittings is the level at which the constraint system can be computed.
From an application perspective,
this means that any spcr that satisfies the implication in Proposition~\ref{prop_idsol} allows not only for localized solving but also for localized computation of the constraint systems.
Thus, while there are safe conditional constraint splittings that are not safe conditional c-solution splittings, the latter are computationally more advantageous.
The difference in the computation of constraint and c-solutions splittings is illustrated in Figure~\ref{fig_constraint_solsplits}.

\begin{figure}[tb]
\begin{subfigure}{.48\textwidth}
	\begin{tikzpicture}
		\node (D) at (0,0) {$\Delta$};
		\node (CD) at (0,-1) {$\CSP(\Delta)$};
		
		\node (sCD) at (0,-2) {$\CSP^\spcr(\Delta)$};
		
		\node (CD1) at (-2,-3) {$\CSP^\spcr(\Delta)|_{\Delta_1}$};
		\node (CDD) at (0,-3) {$\cdots$};
		\node (CDN) at (2,-3) {$\CSP^\spcr(\Delta)|_{\Delta_n}$};

		\draw[->,  thick] (D) -- (CD);
		\draw[->,  thick] (CD) -- (sCD) node [midway, left] {$\spcr$};
		\draw[->,  thick] (sCD) -- (CD1);
		\draw[->,  thick] (sCD) -- (CDD);
		\draw[->,  thick] (sCD) -- (CDN);

	\end{tikzpicture}
	\centering
	\caption{safe conditional constraint splitting}
\end{subfigure}
\begin{subfigure}{.48\textwidth}
	\begin{tikzpicture}
		\node (D) at (0,0) {$\Delta$};
		
		\node (D1) at (-2,-1) {$\Delta_1$};
		\node (DD) at (0,-1) {$\cdots$};
		\node (DN) at (2,-1) {$\Delta_n$};
		
		\node (CD1) at (-2,-2) {$\CSP(\Delta_1)$};
		\node (CDD) at (0,-2) {$\cdots$};
		\node (CDN) at (2,-2) {$\CSP(\Delta_n)$};
		
		\node (sCD1) at (-2,-3) {$\solutionsOf{\CSP(\Delta_1)}$};
		\node (sCDD) at (0,-3) {$\cdots$};
		\node (sCDN) at (2,-3) {$\solutionsOf{\CSP(\Delta_n)}$};
		
		\node (ssCD1) at (-2,-3.8) {$\solutionsOf{\CSP^\spcr(\Delta_1)}$};
		\node (ssCDD) at (0,-3.8) {$\cdots$};
		\node (ssCDN) at (2,-3.8) {$\solutionsOf{\CSP^\spcr(\Delta_n)}$};
		
		\node (ssCD1) at (-2,-3.4) {$=$};
		\node (ssCDD) at (0,-3.4) {$=$};
		\node (ssCDN) at (2,-3.4) {$=$};
		
		\draw[->,  thick] (D) -- (D1);
		\draw[->,  thick] (D) -- (DD);
		\draw[->,  thick] (D) -- (DN);
		
		\draw[->,  thick] (D1) -- (CD1);
		\draw[->,  thick] (DD) -- (CDD);
		\draw[->,  thick] (DN) -- (CDN);
		
		\draw[->,  thick] (CD1) -- (sCD1); %
		\draw[->,  thick] (CDD) -- (sCDD); %
		\draw[->,  thick] (CDN) -- (sCDN); %
	\end{tikzpicture}
	\centering
	\caption{safe c-solution splitting}
\end{subfigure}
\caption{Illustration on how the computation of a c-representation of $\Delta$ can be localized, given that $\Delta$ splits into subbases $\Delta_1,\dots,\Delta_n$. Figure (a) shows the localization if $\safeCondConstraintSplitGen$ while Figure (b) illustrates the localization if $\{\Delta_1,\dots,\Delta_n\}$ is a safe c-solution splitting. Note that, in general, $\CSP^\spcr(\Delta)|_{\Delta_i}\neq\CSP^\spcr(\Delta_i)$.}
\label{fig_constraint_solsplits}
\end{figure} 

Note that the ``culprit'' in the proof of Proposition~\ref{prop_constrnosol} is rule R6 as can be seen in the following Example.
	\begin{example}
		\label{exa_r6bad}
		Let $\DeltaBird$ from Example~\ref{ex_constraints} with $\CSP(\DeltaBird)$ given by
		\begin{align*}
			\constraintOfWrtI{\DeltaBird}{}{1}:\eta_1 &> \min\{0,\eta_2,\eta_2+\eta_4,\eta_4\} - \min\{0,\eta_4\}\\
			\constraintOfWrtI{\DeltaBird}{}{2}:\eta_2 &> \min\{\eta_1,\eta_3,\eta_1+\eta_4\} - \min\{0,\eta_3,\eta_4\}\\
			\constraintOfWrtI{\DeltaBird}{}{3}:\eta_3 &> \min\{\eta_1,\eta_2,\eta_1+\eta_4, \eta_2+\eta_4\} - \min\{0,\eta_2\}\\
			\constraintOfWrtI{\DeltaBird}{}{4}:\eta_4 &> \min\{0,\eta_1,\eta_2\} - \min\{0,\eta_1,\eta_2\}.
		\end{align*}
		For $\spcrSubset=\{R1,R2\}$ we obtain $\CSP^\spcrSubset(\DeltaBird)$ given by
		\begin{align*}
			\constraintOfWrtI{\DeltaBird}{\spcrSubset}{1}:\eta_1 &> \min\{0\} - \min\{0\}\\
			\constraintOfWrtI{\DeltaBird}{\spcrSubset}{2}:\eta_2 &> \min\{\eta_1,\eta_3\} - \min\{0\}\\
			\constraintOfWrtI{\DeltaBird}{\spcrSubset}{3}:\eta_3 &> \min\{\eta_1,\eta_2\} - \min\{0\}\\
			\constraintOfWrtI{\DeltaBird}{\spcrSubset}{4}:\eta_4 &> \min\{0\} - \min\{0\}.
		\end{align*}
		Observe that for $\spcrSubset$ every safe conditional constraint splitting $\safeCondConstraintSplitTwoMacroWrt{\DeltaBird}{\DeltaBird_1}{\DeltaBird_2}{\DeltaBird_0}{\spcrSubset}$ yields a conditional c-solution splitting $\{\DeltaBird_1,\DeltaBird_2\}$ of $\DeltaBird$.
		However, applying $\natSpcr=\{R1,\dots,R6\}$ to $\CSP(\DeltaBird)$ leads to $\CSP^\natSpcr(\DeltaBird)$ as shown in the proof of Proposition~\ref{prop_constrnosol} and we obtain splitting \eqref{eq_constrnosol}, which is no conditional c-solution splitting. 
		Note that while $\spcrSubset$ and $\natSpcr$ both apply $R1$ and $R2$, only $\natSpcr$ applies $R6$.
	\end{example}
 
$R6$ is the only rule of $\natSpcr$ that is applied across two conditionals instead of only taking one conditional into account at a time. 
Indeed, upon closer inspection, the rules from $R1$-$R6$ all exploit varying degrees of connections between the constraint variables. Rules $R1$ and $R2$ only make use of basic algebraic properties of the minimum. Rules $R3,R4,$ and $R5$ additionally compare the sets of verifying and falsifying conditionals, exploiting also properties of the subtraction operator. $R6$ applies reductions across constraints, exploiting properties of the constraint system itself. It is therefore not surprising that $R1$-$R6$ behave differently with respect to the structural properties of the constraint system. In the following, we will show that an spcr applying only rules $R1$ and $R2$ satisfies the implication in Proposition~\ref{prop_idsol} as well. We focus on $R1$ and $R2$ for two reasons. First, they are the most ``benign'' rules with respect to their structural impact, only making use of arithmetic properties of the minimum, which is an intentional deliberation also done, \eg, in \cite{KernIsbernerEtalKR2026}. Second, it has been shown experimentally that the rules $R1$ and $R2$ yield the most benefit for speeding up implementations of inductive inference with c-representations \cite{BeierlevonBergSanin2022FLAIRSarticle, vonBergSaninBeierle2024IJAR} and are therefore the most interesting ones to investigate from a practical point of view.

In the rest of this section we will take a closer look at conditions under which conditional constraint splittings yield conditional c-solution splittings. In particular, we show that they hold for $\spcrSubset=\{R1,R2\}$. 
To facilitate this analysis, we introduce the following notion of constraint restriction. For a \BB\ $\Delta$ with subbase $\Delta_1\subseteq\Delta$, the constraint restriction of $\CSP(\Delta)$ to $\Delta_1$ is based on $\CSP(\Delta)$ (unlike $\CSP(\Delta_1)$) but takes only the conditionals of $\Delta_1$ into account. Note that this is different from \projekting\ $\CSP(\Delta)$ to $\Delta_1$, as $\CSP(\Delta)\marg_{\Delta_1}$ may also mention constraint variables associated with conditionals outside $\Delta_1$.

\begin{definition}[constraint restriction]
	\label{def_crFunc}
	Let $\Delta$ be a \BB\ and let $\Delta_1\subseteq\Delta$ with $\delta_i\in\Delta_1$. We define the \emph{constraint restriction} operator $\crFuncGen$ %
	for sets of sets of conditionals $\mathcal{S}\subseteq 2^\Delta$ by
		\[
		\crFuncGenOf{\mathcal{S}}=\{S\cap\Delta_1\mid S\in \mathcal{S}\},
		\]
	for constraints $\constraintOfWrtI{\Delta}{\spcr}{i}$ by
		\[
		 \crFuncGenOf{\constraintOfWrtI{\Delta}{\spcr}{i}}= \eta_i>\min\{\sum_{\delta_j\in S} \eta_j\mid S\in \crFuncGenOf{\VSet^\spcr_\Delta(\delta_i)}\} - \min\{\sum_{\delta_j\in S} \eta_j\mid S\in \crFuncGenOf{\FSet^\spcr_\Delta(\delta_i)}\},
		\]
	and for CSPs by
		\[
		 \crFuncGenOf{\CSP^\spcr(\Delta)}=\{\crFuncGenOf{\constraintOfWrtI{\Delta}{\spcr}{j}}\mid \delta_j\in\Delta_1\}.
		\]

\end{definition}
Note that $\crFunc$ is a constraint reduction, but in general not solution preserving and therefore no spcr. 
In general  $\crFuncGenOf{\CSP^{\spcr}(\Delta)}\neq\CSP^{\spcr}(\Delta)\marg_{\Delta_1}$ and $\crFuncGenOf{\CSP^{\spcr}(\Delta)}\neq\CSP^{\spcr}(\Delta_1)$.
We illustrate Definition~\ref{def_crFunc} with an example.

\begin{example}
	\label{exa_crFunc}
	Recall $\DeltaBird$ from Example~\ref{ex_constraints} with $\CSP(\DeltaBird)$ given by:
	\begin{align*}
		\constraintOfWrtI{\DeltaBird}{}{1}:\eta_1 &> \min\{0,\eta_2,\eta_2+\eta_4,\eta_4\} - \min\{0,\eta_4\}\\
		\constraintOfWrtI{\DeltaBird}{}{2}:\eta_2 &> \min\{\eta_1,\eta_3,\eta_1+\eta_4\} - \min\{0,\eta_3,\eta_4\}\\
		\constraintOfWrtI{\DeltaBird}{}{3}:\eta_3 &> \min\{\eta_1,\eta_2,\eta_1+\eta_4, \eta_2+\eta_4\} - \min\{0,\eta_2\}\\
		\constraintOfWrtI{\DeltaBird}{}{4}:\eta_4 &> \min\{0,\eta_1,\eta_2\} - \min\{0,\eta_1,\eta_2\}.
	\end{align*}
	We first show that constraint restriction and projection do not necessarily match.
	Let $\DeltaBird_1=\{\delta_1,\delta_2\}$. 
	Then $\crFuncOfWrt{\CSP(\DeltaBird)}{\DeltaBird_1}$ is given by
	\begin{align*}
		\crFuncOfWrt{\constraintOfWrtI{\DeltaBird}{}{1}}{\DeltaBird_1}:\eta_1 &> \min\{0,\eta_2\} - \min\{0\}\\
		\crFuncOfWrt{\constraintOfWrtI{\DeltaBird}{}{2}}{\DeltaBird_1}:\eta_2 &> \min\{\eta_1,0\} - \min\{0\}.
	\end{align*}
	$\CSP(\DeltaBird)\marg_{\DeltaBird_1}$ is given by	 
	\begin{align*}
		\constraintOfWrtI{\DeltaBird}{}{1}:\eta_1 &> \min\{0,\eta_2,\eta_2+\eta_4,\eta_4\} - \min\{0,\eta_4\}\\
		\constraintOfWrtI{\DeltaBird}{}{2}:\eta_2 &> \min\{\eta_1,\eta_3,\eta_1+\eta_4\} - \min\{0,\eta_3,\eta_4\}.
	\end{align*}
	Thus, $\crFuncOfWrt{\CSP(\DeltaBird)}{\DeltaBird_1} = \CSP(\DeltaBird_1)\neq \CSP(\DeltaBird)\marg_{\DeltaBird_1}$.
	
	Next, we will see that constraint restriction also does not necessarily match with constraints derived on the subbase when spcrs are involved.
	Recall $\CSP^\natSpcr(\DeltaBird)$ from Example~\ref{exa_natspcr} given by
	\begin{align*}
		\constraintOfNatspcrI{\DeltaBird}{1}: \eta_1 &> \min\{0\} - \min\{0\}\\
		\constraintOfNatspcrI{\DeltaBird}{2}: \eta_2 &> \min\{\eta_1\} - \min\{0\}\\
		\constraintOfNatspcrI{\DeltaBird}{3}: \eta_3 &> \min\{\eta_1\} - \min\{0\}\\
		\constraintOfNatspcrI{\DeltaBird}{4}: \eta_4 &> \min\{0\} - \min\{0\}.
	\end{align*} 
	Then  $\crFuncOfWrt{\CSP^\natSpcr(\DeltaBird)}{\DeltaBird_1}$ is given by 
	\begin{align*}
		\crFuncOfWrt{\constraintOfWrtI{\DeltaBird}{\natSpcr}{1}}{\DeltaBird_1}:\eta_1 &> \min\{0\} - \min\{0\}\\
		\crFuncOfWrt{\constraintOfWrtI{\DeltaBird}{\natSpcr}{2}}{\DeltaBird_1}:\eta_2 &> \min\{\eta_1\} - \min\{0\}.
	\end{align*}
	But $\CSP^\natSpcr(\DeltaBird_1)$ is given by 
	\begin{align*}
		\constraintOfWrtI{\DeltaBird_1}{\natSpcr}{1}:\eta_1 &> \min\{0\} - \min\{0\}\\
		\constraintOfWrtI{\DeltaBird_1}{\natSpcr}{2}:\eta_2 &> \min\{0\} - \min\{0\}.
	\end{align*}
	Thus, $\crFuncOfWrt{\CSP^\natSpcr(\DeltaBird)}{\DeltaBird_1} = \CSP^\natSpcr(\DeltaBird)\marg_{\DeltaBird_1}\neq \CSP^\natSpcr(\DeltaBird_1)$.
	
	Finally, observe that the order of application of the spcr and $\crFunc$ matters, as can be seen by $\natSpcr(\crFuncOfWrt{\CSP(\DeltaBird)}{\DeltaBird_1})$:
	\begin{align*}
		\natSpcr(\crFuncOfWrt{\constraintOfWrtI{\DeltaBird}{}{1}}{\DeltaBird_1}):\eta_1 &> \min\{0\} - \min\{0\}\\
		\natSpcr(\crFuncOfWrt{\constraintOfWrtI{\DeltaBird}{}{2}}{\DeltaBird_1}):\eta_2 &> \min\{0\} - \min\{0\}.
	\end{align*}
	Thus, $\crFuncOfWrt{\CSP^\natSpcr(\DeltaBird)}{\DeltaBird_1}\neq \natSpcr(\crFuncOfWrt{\CSP(\DeltaBird)}{\DeltaBird_1})=\CSP^\natSpcr(\DeltaBird_1)$.
\end{example}

As illustrated in Example~\ref{exa_crFunc}, the order of application of $\crFunc$ and the spcr is significant. Unless specified otherwise, we will assume that the spcr is always applied first.

Now we will examine conditions under which $\crFuncGenOf{\CSP^{\spcr}(\Delta)}=\CSP^{\spcr}(\Delta)\marg_{\Delta_1}$ and $\crFuncGenOf{\CSP^{\spcr}(\Delta)}=\CSP^{\spcr}(\Delta_1)$ hold. The following lemma states that in the special case where the spcr is $\idFct$ (\ie, no spcr is applied to $\CSP(\Delta)$), the equality $\crFuncGenOf{\CSP^{\idFct}(\Delta)}=\CSP^{\idFct}(\Delta_1)$ always holds.

\begin{lemma}
	\label{lemm_crFunc_id}
	Let $\Delta$ be a \BB\ and $\Delta_1\subseteq\Delta$. Then $\crFuncGenOf{\CSP(\Delta)}=\CSP(\Delta_1)$.
\end{lemma}
\begin{proof}
	Let $\Delta=\{(B_1|A_1),\dots,(B_n|A_n)\}$. Let $\Delta'\subseteq\Delta$ and let $(B_i|A_i)\in\Delta_1$.
	In order to show $\crFuncGenOf{\CSP(\Delta)}=\CSP(\Delta_1)$ we need to show $\crFuncGenOf{\constraintOfWrtI{\Delta}{}{i}}=\constraintOfWrtI{\Delta_1}{}{i}$ for all $\delta_i\in\Delta$. Then it suffices to show  $\crFuncGenOf{\VSet_\Delta(\delta_i)}= \VSet_{\Delta_1}(\delta_i)$ and $\crFuncGenOf{\FSet_\Delta(\delta_i)}= \FSet_{\Delta_1}(\delta_i)$.
	We show $\crFuncGenOf{\VSet_\Delta((B_i|A_i))}= \VSet_{\Delta_1}((B_i|A_i))$, the case $\crFuncGenOf{\FSet_\Delta((B_i|A_i))}= \FSet_{\Delta_1}((B_i|A_i))$ is analogous.
	We show both set inclusions.
	
	"$\subseteq$": Let $V\in \crFuncGenOf{\VSet_\Delta((B_i|A_i))}$. Then there is $V'\in \VSet_\Delta((B_i|A_i))$ such that $V'\cap\Delta_1=V$. Then there is $\omega_{V'}$ such that $\omega_{V'}\models A_iB_i$ and $\omega_{V'}\models A_j\ol{B_j}$ for all $(B_j|A_j)\in V'$. Then clearly $\omega_{V'}\models A'_j\ol{B'_j}$ for all $(B'_j|A'_j)\in V=V'\cap\Delta_1$. Because $\omega_{V'}\models B_i|A_i)$, we also have $\omega_{V'}\in \arg\VSet_{\Delta_1}((B_i|A_i))$. Thus	
	$V\in \VSet_{\Delta_1}((B_i|A_i))$.
	
	"$\supseteq$": Let $V_1\in \VSet_{\Delta_1}((B_i|A_i))$. Then there is $\omega_{V_1}$ such that $\omega_{V_1}\models A_iB_i$ and $\omega_{V_1}\models A_j\ol{B_j}$ for all $(B_j|A_j)\in V_1$. Because $\omega_{V_1}\models A_iB_i$, we have $\omega_{V_1}\in \arg\VSet_{\Delta}((B_i|A_i))$. Then there must be some $V$ with $V_1\subset V$ such that $V\in \VSet_{\Delta}((B_i|A_i))$ and $\omega_{V_1}\models A'_j\ol{B'_j}$ for all $(B'_j|A'_j)\in V$. Then $V_1=V\cap\Delta_1$ and thus $V_1\in\crFuncGenOf{\VSet_\Delta((B_i|A_i))}$.
\end{proof}

Next, the following proposition shows that equivalence under \projektion\ implies equivalence under constraint restriction.

\begin{proposition}
	\label{prop_margimpliesfunc}
	Let $\Delta$ be a \BB\ and let $\spcr$ be an spcr.
	$\CSP^{\spcr}(\Delta)|_{\Delta_1}=\CSP^{\spcr}(\Delta_1)$ implies $\crFuncGenOf{\CSP^\spcr(\Delta)}=\CSP^\spcr(\Delta_1)$ but not vice versa.
\end{proposition}
\begin{proof}
	Observe that $\crFuncGenOf{\constraintOfWrtI{\Delta}{\spcr}{i}}\in\crFuncGenOf{\CSP^\spcr(\Delta)}$ iff $\constraintOfWrtI{\Delta}{\spcr}{i}\in\CSP^{\spcr}(\Delta)\marg_{\Delta_1}$.
	Now, assume $\CSP^{\spcr}(\Delta)|_{\Delta_1}=\CSP^{\spcr}(\Delta_1)$.
	Then $\constraintOfWrtI{\Delta}{\spcr}{i}=\constraintOfWrtI{\Delta_1}{\spcr}{i}$ for all $\delta_i\in\Delta_1$. Thus, also $\crFuncGenOf{\constraintOfWrtI{\Delta}{\spcr}{i}}=\constraintOfWrtI{\Delta_1}{\spcr}{i}$ and therefore $\crFuncGenOf{\CSP^\spcr(\Delta)}=\CSP^\spcr(\Delta_1)$.
	
	Example~\ref{exa_crFunc} provides a counterexample for the other direction.
\end{proof}

The next proposition considers the inverse implication of Proposition~\ref{prop_margimpliesfunc} for the special case that $\CSP^{\spcr}(\Delta)\marg_{\Delta_1}$ mentions no constraint variables outside $\Delta_1$.

\begin{proposition}
	\label{prop_funcimpliesmarg}
	Let $\Delta$ be a \BB\ and let $\spcr$ be an spcr.
	If $\CSP^{\spcr}(\Delta)\marg_{\Delta_1}$ mentions no constraint variables outside $\Delta_1$, then 
	$\CSP^{\spcr}(\Delta)|_{\Delta_1}=\CSP^{\spcr}(\Delta_1)$ iff $\crFuncGenOf{\CSP^\spcr(\Delta)}=\CSP^\spcr(\Delta_1)$
\end{proposition}
\begin{proof}
	$\Rightarrow$ Follows from Proposition~\ref{prop_margimpliesfunc}.
	
	$\Leftarrow$: 
	First, observe that $\crFuncGenOf{\constraintOfWrtI{\Delta}{\spcr}{i}}\in\crFuncGenOf{\CSP^\spcr(\Delta)}$ iff $\constraintOfWrtI{\Delta}{\spcr}{i}\in\CSP^{\spcr}(\Delta)\marg_{\Delta_1}$.
	Assume that $\CSP^{\spcr}(\Delta)\marg_{\Delta_1}$ mentions no constraint variables outside $\Delta_1$. Then $\crFuncGenOf{\VSet^\spcr_\Delta(\delta_i)}=\VSet^\spcr_\Delta(\delta_i)$ and $\crFuncGenOf{\FSet^\spcr_\Delta(\delta_i)}=\FSet^\spcr_\Delta(\delta_i)$ per definition.
	Then also $\crFuncGenOf{\constraintOfWrtI{\Delta}{\spcr}{i}}=\constraintOfWrtI{\Delta}{\spcr}{i}$ and thus
	$\crFuncGenOf{\CSP^\spcr(\Delta)}=\CSP^\spcr(\Delta)\marg_{\Delta_1}$. Due to  $\crFuncGenOf{\CSP^\spcr(\Delta)}=\CSP^\spcr(\Delta_1)$ we then obtain $\CSP^\spcr(\Delta)\marg_{\Delta_1}= \CSP^\spcr(\Delta_1)$.
\end{proof}

Finally, we use and apply Lemma~\ref{lemm_crFunc_id} and Propositions~\ref{prop_margimpliesfunc} and \ref{prop_funcimpliesmarg} to the case where the spcr is given by $\spcrSubset$.
\begin{proposition}
	\label{prop_subsetcsplitimpliessolsplit}
	Let $\Delta$ be a belief base, let $\spcrSubset=\{R1,R2\}$ and let $\niceSafeCondConstraintSplitTwoWrt{\spcrSubset}$. Then 
	\[
		\CSP^{\spcrSubset}(\Delta)|_{\Delta_1}=\CSP^{\spcrSubset}(\Delta_1)\quad\text{and}\quad\CSP^{\spcrSubset}(\Delta)|_{\Delta_2}=\CSP^{\spcrSubset}(\Delta_2).
	\]
\end{proposition}
\begin{proof}
	In the following we will consider non-exhaustive applications of the rules $\spcrSubset$ by looking at a specific application 
	$r$
	of $R1$ or $R2$.
	Note that $r$ is defined on tuples $\langle\mathcal{S},\mathcal{S'}\rangle$ with $\mathcal{S},\mathcal{S'}\in 2^\Delta$, but $R1$ and $R2$ only ever apply to set $\mathcal{S}$ or set $\mathcal{S'}$ respectively, leaving the other set unchanged. To ease notation, we will therefore say that $r$ is applied to a single set $\mathcal{S}\subseteq 2^\Delta$ when $r$ is actually applied to a tuple $\langle\mathcal{S},\mathcal{S'}\rangle$ (or $\langle\mathcal{S}',\mathcal{S}\rangle$), where the other set $\mathcal{S}'$ is unchanged. This omits $\mathcal{S}'$ but since $\mathcal{S}$ and $r$ are sufficient to describe the effect of applying $r$ to $\langle\mathcal{S},\mathcal{S'}\rangle$ (or $\langle\mathcal{S}',\mathcal{S}\rangle$), no information will be lost.

	Then, for a given set of sets of conditionals $\mathcal{S}\subseteq 2^\Delta$, $r$ denotes a single, specific (ground) application of rule $R1$ or $R2$ that is applicable to $\mathcal{S}$. The result of applying $r$ to $\mathcal{S}$ is denoted $\mathcal{S}^r$. %
	Thus, there are sets $S,S'\in \mathcal{S}$ with $S\subsetneq S'$ such that $\mathcal{S}^r=\mathcal{S}\setminus \{S'\}$; $r$ removes $S'$ from $\mathcal{S}$.
	Note that $\mathcal{S}^r$ is uniquely defined.
	Similarly, $r'=(r_1,\dots,r_n)$
	denotes a sequence
	of specific applications of rules $R1$ or $R2$, applying exactly the reductions $(r_1,\dots,r_n)$ one at a time. Such a sequence $r'=(r_1,\dots,r_n)$ is applicable to a set $\mathcal{S}\subseteq 2^\Delta$, if $r_1$ is applicable to $\mathcal{S}$, and, for $i\in\{2,\dots, n\}$, $r_{i}$ is applicable to $\mathcal{S}^{r_{i-1}}$. Then $\mathcal{S}^{r'}$ denotes the application of $r'$ to $\mathcal{S}$.
	
	Assume $\niceSafeCondConstraintSplitTwoWrt{\spcrSubset}$. We will show that either $\CSP^{\spcrSubset}(\Delta)|_{\Delta_1}=\CSP^{\spcrSubset}(\Delta_1)$ and $\CSP^{\spcrSubset}(\Delta)|_{\Delta_2}=\CSP^{\spcrSubset}(\Delta_2)$ or we will derive a contradiction. We focus only on $\Delta_1$ here because the case for $\Delta_2$ is analogous.
	
	First, we make the following observations:
	\begin{enumerate}
		\item  For $\constraintOfWrtI{\Delta}{}{i}$ and $\constraintOfWrtI{\Delta_1}{}{i}$, $\crFuncGen$ is defined in such a way that $\crFuncGenOf{\constraintOfWrtI{\Delta}{}{i}}=\constraintOfWrtI{\Delta_1}{}{i}$  iff $\crFuncGenOf{\VSet_{\Delta}(\delta_i)}=\VSet_{\Delta_1}(\delta_i)$ and $\crFuncGenOf{\FSet_{\Delta}(\delta_i)}=\FSet_{\Delta_1}(\delta_i)$. 
		\item $\CSP^\spcr(\Delta)|_{\Delta_1}=\CSP^\spcr(\Delta_1)$ iff $\constraintOfWrtI{\Delta}{\spcr}{i}=\constraintOfWrtI{\Delta_1}{\spcr}{i}$ for all $\delta_i\in\Delta_1$.
		\item Because $\spcrSubset$ takes only a single constraint into account, it is sufficient to show $\constraintOfWrtI{\Delta}{\spcr}{i}=\constraintOfWrtI{\Delta_1}{\spcr}{i}$ for some arbitrary $\delta_i\in\Delta_1$.
	\end{enumerate}

	Let $\constraintOfWrtI{\Delta}{}{i}\in \CSP(\Delta)$ with $\delta_i\in\Delta_1$. We show:
	\begin{enumerate}
		\item[(i)] $\crFuncGenOf{\constraintOfWrtI{\Delta}{\spcrSubset}{i}}=\constraintOfWrtI{\Delta_1}{\spcrSubset}{i}$.
		\item[(ii)] $\constraintOfWrtI{\Delta}{\spcrSubset}{i}$ does not mention any constraint variables outside $\Delta_1$.
	\end{enumerate}
	
	(i) and (ii) together with Proposition~\ref{prop_funcimpliesmarg} then imply $\CSP(\Delta)|_{\Delta_1}=\CSP(\Delta_1)$.
	
	We show (ii) first. Towards a contradiction, assume $\constraintOfWrtI{\Delta}{\spcrSubset}{i}$ mentions a constraint variable $\eta_j$ such that $\delta_j\notin\Delta_1$. Then $\delta_j\in\Delta_2$, because $\Delta_1\cup\Delta_2=\Delta$. But then $\delta_i\directlyDependsOfWrt{\Delta}{\spcrSubset}\delta_j$, and thus $\Delta_1\directlyDependsOfWrt{\Delta}{\spcrSubset}\Delta_2$ which is in contradiction to the assumption $\niceSafeCondConstraintSplitTwoWrt{\spcrSubset}$.
	
	We show (i) next. In order to show $\crFuncGenOf{\constraintOfWrtI{\Delta}{\spcrSubset}{i}}=\constraintOfWrtI{\Delta_1}{\spcrSubset}{i}$ we need to show $\crFuncGenOf{\VSet^{\spcrSubset}_{\Delta}(\delta_i)}=\VSet^{\spcrSubset}_{\Delta_1}(\delta_i)$ and $\crFuncGenOf{\FSet^{\spcrSubset}_{\Delta}(\delta_i)}=\FSet^{\spcrSubset}_{\Delta_1}(\delta_i)$. 
	In order to show this, we show the following: 

	($\dagger$) Let $\mathcal{S}\subseteq 2^\Delta$ and let $\mathcal{S}_1\subseteq 2^{\Delta_1}$. If $\crFuncGenOf{\mathcal{S}}=\mathcal{S}_1$ and $\mathcal{S}^\spcrSubset\subseteq2^{\Delta_1}$, then every application $r$ of $\spcrSubset$ to $\mathcal{S}_1$ can be mirrored by applying some sequence $r'$ of $\spcrSubset$ to $\mathcal{S}$ such that $\crFuncGenOf{\mathcal{S}^{r'}}=\mathcal{S}_1^r$ and vice versa.

	Consider a reduction $r$ of $\mathcal{S}_1$. Then there is some $S_1\in \mathcal{S}_1$, but $S_1\notin \mathcal{S}_1^r$. In order for $\crFuncGenOf{\mathcal{S}^{r'}}=\mathcal{S}_1^r$ to hold, there must be $r'$ such that $r'$ removes all sets $\{S\in \mathcal{S} \mid \crFuncGenOf{S}=S_1\}$ from $\mathcal{S}$. Towards a contradiction, assume such an $r'$ does not exist. Then there is some $S\in \mathcal{S}$ such that $\crFuncGenOf{S}=S_1$ but $S$ cannot be removed from $\mathcal{S}$ by application of $\spcrSubset$. We make a case distinction: Either $S\subseteq\Delta_1$ or $S\not\subseteq\Delta_1$. In the second case $S\cap\Delta_2\neq\emptyset$. Because $S$ cannot be removed by $\spcrSubset$, we have $S\in \mathcal{S}^{\spcrSubset}$ which is in contradiction to $\mathcal{S}^\spcrSubset\subseteq2^{\Delta_1}$.	
	Therefore assume $S\subseteq\Delta_1$. Because $S_1$ was removed, there must be $S_1'\in \mathcal{S}_1\cap \mathcal{S}_1^{r}$ such that $S_1'\subsetneq S_1$. Because $S$ could not be removed by $\spcrSubset$, for all  $\{S'\in S \mid \crFuncGenOf{S'}=S'_1\}$ it must hold that $S'\not\subseteq S$. Again we make a case distinction: Either $(S'\setminus S)\subseteq\Delta_1$ or $(S'\setminus S)\not\subseteq\Delta_1$. In the second case, again $S\cap\Delta_2\neq\emptyset$, thus there must either be a sequence of applications of $\spcrSubset$ removing the conditionals of $\Delta_2$, or we contradict $S^\spcrSubset\subseteq2^{\Delta_1}$. 
	If there is such a sequence, then, after applying this sequence, $S$ can also be removed by applying $\spcrSubset$ one more time utilizing $S'$, contradiction. Therefore assume, $(S'\setminus S)\subseteq\Delta_1$. But then $S=S_1$ and $S'=S_1'$ due to $\crFuncGenOf{S}=S_1$ and $\crFuncGenOf{S'}=S'_1$ which is in contradiction to $S_1'\subsetneq S_1$.
	Vice versa, consider some application $p$ of $\spcrSubset$ to $\mathcal{S}$. If $\crFuncGenOf{\mathcal{S}^p}=\mathcal{S}_1$ still holds we do not need to apply any reduction to $\mathcal{S}_1$ and are done. Therefore assume $\crFuncGenOf{\mathcal{S}}\neq \mathcal{S}_1$.
	
	Then there is $T_1\in \mathcal{S}_1$ such that $\{T\in \mathcal{S}^p\mid\crFuncGenOf{T}=T_1\}=\emptyset$.
	Then there is $T\in \mathcal{S}$ such that $T\notin \mathcal{S}^p$ and $T_1\in \crFuncGenOf{T}$. Thus there must be $T'\in \mathcal{S}\cap \mathcal{S}^p$ with $T'\subsetneq T$. By definition then $\crFuncGenOf{T'}\subseteq \crFuncGenOf{T}$. 
	Then there is $T_1'\in\crFuncGenOf{T'}$ such that $T_1'\subseteq T_1$. The case $T_1'= T_1$ is in contradiction to $\{T\in \mathcal{S}^p\mid\crFuncGenOf{T}=T_1\}=\emptyset$, thus $T_1'\subsetneq T_1$. Then there is an application $p'$ of $\spcrSubset$ that removes $T_1$ from $\mathcal{S}_1$. Thus $\crFuncGenOf{\mathcal{S}^p}=\mathcal{S}_1^{p'}$. 

	Observe that the prerequisites of $(\dagger)$ are met in (i), \ie, $\VSet_{\Delta}(\delta_i)^\spcrSubset\subseteq 2^{\Delta_1}$ and $\FSet_{\Delta}(\delta_i)^\spcrSubset\subseteq 2^{\Delta_1}$ holds due to (ii) and 
	$\crFuncGenOf{\VSet_{\Delta}(\delta_i)}=\VSet_{\Delta_1}(\delta_i)$ and $\crFuncGenOf{\FSet_{\Delta}(\delta_i)}=\FSet_{\Delta_1}(\delta_i)$ holds by Lemma~\ref{lemm_crFunc_id}.
	Thus, $(\dagger)$ can be applied for every step of the reduction $\spcrSubset$ until $\spcrSubset$ is exhaustively applied and (i) follows. 
	The usage of $\crFunc$ in the proof of (i) is illustrated in Figure~\ref{fig_crFunc}.

\begin{figure}
	\centering
	\begin{tikzpicture}
		\centering
		\node (A1) at (-1,0) {$\CSP(\Delta)$};
		\node (A2) at (-1,-2) {$\CSP(\Delta_1)$};
		
		\node (B1) at (2,0) {$\constraintOfWrtI{\Delta}{}{i}$};
		\node (B2) at (2,-2) {$\constraintOfWrtI{\Delta_1}{}{i}$};
		
		\node (C1) at (4,0) {$\VSet_{\Delta}(\delta_i)$};
		\node (C2) at (4,-2) {$\VSet_{\Delta_1}(\delta_i)$};

		\node (D1) at (6,0) {$\VSet_{\Delta}^{r'}(\delta_i)$};
		\node (D2) at (6,-2) {$\VSet_{\Delta_1}^{r}(\delta_i)$};

		\node (E1) at (8,0) {$\VSet_{\Delta}^{\spcrSubset}(\delta_i)$};
		\node (E2) at (8,-2) {$\VSet_{\Delta_1}^{\spcrSubset}(\delta_i)$};
		
		\node (R1) at (7,0) {$\dots$};
		\node (R2) at (7,-2) {$\dots$};
		
		\node (F1) at (10,0) {$\constraintOfWrtI{\Delta}{\spcrSubset}{i}$};
		\node (F2) at (10,-2) {$\constraintOfWrtI{\Delta_1}{\spcrSubset}{i}$};
		
		\node (G1) at (13,0) {$\CSP^\spcrSubset(\Delta)$};
		\node (G2) at (13,-2) {$\CSP^\spcrSubset(\Delta_1)$};

		\draw[->] (A1) -- node[left] {$\crFuncGen$} (A2);
		\draw[->] (B1) -- node[left] {$\crFuncGen$} (B2);
		\draw[->] (C1) -- node[left] {$\crFuncGen$} (C2);
		\draw[->] (D1) -- node[left] {$\crFuncGen$} (D2);
		\draw[->] (E1) -- node[left] {$\crFuncGen$} (E2);
		
		\draw[->] (F1) -- node[left] {$\crFuncGen$} (F2);
		\draw[->] (G1) -- node[left] {$\crFuncGen$} (G2);

		\draw[->] (C1) -- node[above] {$r'$} (D1);
		\draw[->] (C2) -- node[above] {$r$} (D2);
		
		\draw[->] (C1) to[bend left] node[above] {$\spcrSubset$} (E1);
		\draw[->] (C2) to[bend right] node[above] {$\spcrSubset$} (E2);
		
		\draw[->] (A1) -- node[above] {$\forall \delta_i\in\Delta_1$} (B1);
		\draw[->] (B1) -- (C1);
		
		\draw[->] (A2) -- node[above] {$\forall \delta_i\in\Delta_1$} (B2);
		\draw[->] (B2) -- (C2);
		
		\draw[->] (D1) -- (R1);
		\draw[->] (D2) -- (R2);
		
		\draw[->] (R1) -- (E1);
		\draw[->] (R2) -- (E2);
		
		\draw[->] (E1) -- (F1);
		\draw[->] (E2) -- (F2);
		
		\draw[->] (F1) -- node[above] {$\forall \delta_i\in\Delta_1$} (G1);
		\draw[->] (F2) -- node[above] {$\forall \delta_i\in\Delta_1$} (G2);
	\end{tikzpicture}
	\caption{Illustration of the usage of the constraint restriction $\crFunc$ in the presence of $\niceSafeCondConstraintSplitTwoWrt{\spcrSubset}$ with $\spcrSubset=\{R1,R2\}$ in the proof for Proposition~\ref{prop_subsetcsplitimpliessolsplit}. The paths for $\FSet_{\Delta}(\delta_i)$ and $\FSet_{\Delta_1}(\delta_i)$ are omitted as they are analogous to $\VSet_{\Delta}(\delta_i)$ and $\VSet_{\Delta_1}(\delta_i)$.}
	\label{fig_crFunc}
\end{figure} \end{proof}

The next proposition now shows that each conditional constraint splittings \wrt\ $\spcrSubset$ yields a conditional c-solutions splitting.

\begin{proposition}
	\label{prop_subsetspcrcsolsplit}
	Let $\Delta$ be a belief base, and let $\spcrSubset=\{R1,R2\}$. Then $\niceSafeCondConstraintSplitTwoWrt{\spcrSubset}$ implies that $\{\Delta_1,\Delta_2\}$ is a safe conditional c-solution splitting of $\Delta$
\end{proposition}
\begin{proof}
	Follows directly from Propositions~\ref{prop_subconstraintssplitimplication} and \ref{prop_subsetcsplitimpliessolsplit}.
\end{proof}

\begin{figure}[tb]
	\centering
	\begin{tikzpicture}
		\centering
		\node (CSI) at (-3.5,-1) {$\niceSafeCondConstraintSplitTwoWrt{\sf id}$};
		\node (CSR) at (0,-1) {$\niceSafeCondConstraintSplitTwoWrt{\spcrSubset}$};
		\node (CSN) at (3.5,-1) {$\niceSafeCondConstraintSplitTwoWrt{\natSpcr}$};
		
		\node (SYN) at (8,-1) {$\genSafeCondSynSplitGen$};
		
		\node (MARG) at (0,-2) {$\CSP(\Delta)\marg_{\Delta_i}=\CSP(\Delta_i)$};
		
		\node (SOL) at (0,-3) {safe conditional c-solution splitting};

		\draw[->] (CSN.south west) to node {\rotatebox{45}{$\times$}} (MARG.north east);
		\draw[->] (CSR.south) to (MARG.north);
		\draw[->] (CSI.south east) to (MARG.north west);
		\draw[->] (MARG.south) to node[right] {$\ast$} (SOL.north);
		
		\draw[->, dashed] (SYN.south) to (8,-2) to (MARG.east);
		
		\draw[->, dashed] (8,-2) to (8,-3) to (SOL.east);

		\draw[->] (SYN.north) to (8,0) to (0,0) to (CSR.north);
		
		\draw[->] (CSR.east)  to (CSN.west);
		
	\end{tikzpicture}
	\caption{
		A closer analysis of the relationship among conditional constraint and conditional c-solution splittings as well as conditional syntax splittings. The investigated spcrs are id, $\spcrSubset=\{R1,R2\}$ and $\natSpcr=\{R1,\dots,R6\}$. Pictured is only the binary case as c-solution splittings are only defined for the binary case. An arrow ($S_1\rightarrow S_2$) from splitting $S_1$ to splitting $S_2$ means that each splitting of type $S_1$ yields a splitting of type $S_2$. A $\times$-symbol at an arrow indicates that the implication does not hold.
		Solid arrows ($\rightarrow$) indicate results from this article while dashed arrows ($\dashrightarrow$) indicate results from \cite{SpiegelHaldimannHeyninckKernIsbernerBeierle2025IJCAI, KernIsbernerEtalKR2026}. 
		The implication marked with $\ast$ has been utilized in a different context in \cite{SpiegelHaldimannHeyninckKernIsbernerBeierle2025IJCAI, KernIsbernerEtalKR2026}, but we have made it explicit for conditional c-solution splittings here.} 
	\label{fig_spcrsplits}
\end{figure}
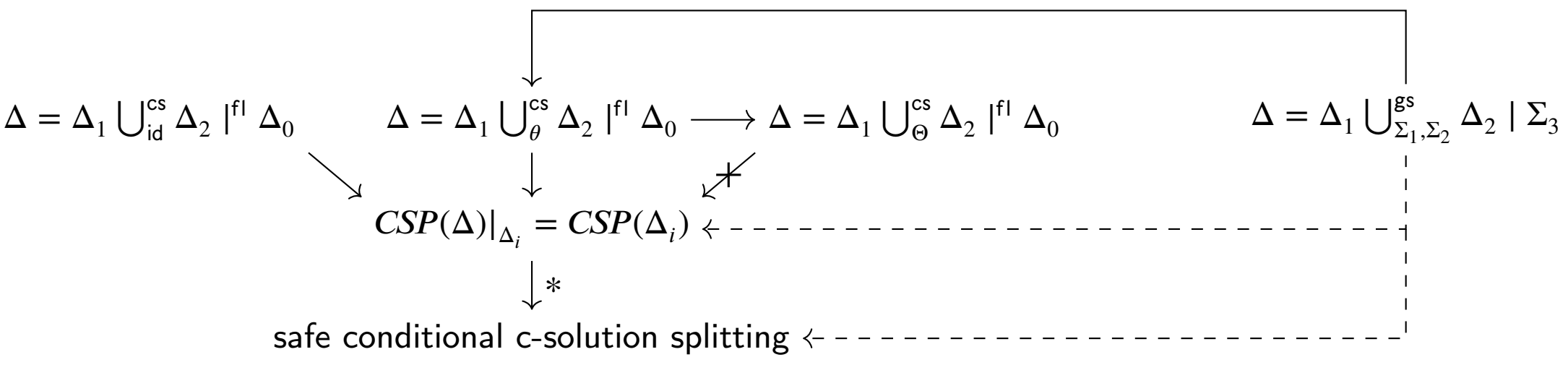

In this section we have shown that the connection between conditional constraint splittings and conditional solution splittings is crucially dependent on the spcr underlying the constraint splitting. For spcrs $\idFct$ and $\spcrSubset$, each \niceSafe\ conditional constraint splitting \wrt\ them yields a safe conditional c-solution splitting, while not every \niceSafe\ conditional constraint splitting \wrt\ $\natSpcr$ does so.
Figure~\ref{fig_spcrsplits} provides an overview of the relationships among conditional syntax, conditional c-solution and conditional constraint splittings \wrt\ to different spcrs $\idFct$, $\spcrSubset$ and $\natSpcr$. The figure shows that conditional constraint splittings can be placed as an intermediary between splittings on the syntax and splittings on the solutions, providing a more in-depth analysis on the dependencies among conditionals.	

\section{Related Work}
There are several works on splitting  techniques in the literature, \eg,
\cite{JH_Parikh99,PeppasWilliamsChopraFoo2015,KernIsbernerBrewka2017,KernIsbernerBeierleBrewka2020KR,HeyninckKernIsbernerMeyerHaldimannBeierle2023AAAI,WilhelmSezginKernIsbernerHaldimannBeierleHeyninck2023JELIA}. Some papers \cite{KernIsbernerBeierleBrewka2020KR,BeierleHaldimannKernIsberner2021Boeger75Festschrift,BeierleSpiegelHaldimannWilhelmHeyninckKernIsberner2024KR,SpiegelHaldimannHeyninckKernIsbernerBeierle2025IJCAI} investigate reasoning with c-representations in the context of syntax splitting postulates for inductive inference. 
Next to the works on %
constraint splitting \cite{BeierleHaldimannKernIsberner2021Boeger75Festschrift,WilhelmSezginKernIsbernerHaldimannBeierleHeyninck2023JELIA}, which provide the starting point of this article, there is only one other paper that concerns itself chiefly with splittings for c-representations based on their constraints %
\cite{WilhelmKernIsbernerAAAI24}.
Following a similar line of thought to ours, the notion of \emph{safe covers} \cite{WilhelmKernIsbernerAAAI24} aims at breaking down \BBs\ into smaller subbases to compute c-representations locally.

\begin{definition}[safe cover \cite{WilhelmKernIsbernerAAAI24}]
	Let $\Delta$ be a \BB\ and $\spcr$ be an spcr. A set $\{\Delta_1,\dots,\Delta_m\}$ is a \emph{safe cover of $\Delta$ \wrt\ $\spcr$} if 
	$\bigcup_{k=1}^{m}\Delta_k = \Delta$ and $\Delta_i$ is closed under $\directlyDependsOfWrtSpcr{\Delta}$ for all $i\in\{1,\dots,m\}$.
\end{definition}

Constraint splittings are also safe covers \cite{WilhelmKernIsbernerAAAI24}. 
The following example shows that conditional constraint splittings
are more general than safe covers.

\begin{example}
	Consider $\DeltaC=\{\delta_1\!:\!(c|a\lor b),\ \delta_2\!:\!(c|a),\ \delta_3\!:\!(c|b)\}$ 
	with corresponding constraints
	\begin{align}
		\constraintOfNatspcrI{\DeltaO}{1}:\eta_1 &> \min\{0\} - \min\{\eta_2,\eta_3\}\\
		\constraintOfNatspcrI{\DeltaO}{2}:\eta_2 &> \min\{0\} - \min\{\eta_1\}\\
		\constraintOfNatspcrI{\DeltaO}{3}:\eta_3 &> \min\{0\} - \min\{\eta_1\}.
	\end{align}
	Then
	$\condConstraintSplitMacroTwoNatSpcr{\DeltaC}{\{(c|a\lor b),(c|a)\}}{\{(c|a\lor b),(c|b)\}}{\{(c|a\lor b)\}}$, cf. Example~\ref{exa_unsafesplit}.
	However, the only safe cover of $\DeltaC$ is the trivial cover $\{\DeltaC\}$.
\end{example}

On the other hand, safe conditional constraint splittings are as expressive as safe covers as is shown in the following two propositions.
In order to construct a safe cover from a safe conditional constraint splitting, the conditionals in $\DeltaZero$ must be taken into account in all relevant subbases which is the case for \niceSafe\ splittings.
\begin{propositionrep}
	\label{prop_splitcover}
	Let $\Delta$ be a \BB\ and let $\spcr$ be an spcr.
	If
	$\niceSafeCondConstraintSplitGen$
	then
	$\{\Delta_0,\Delta_1,\dots,\Delta_m\}$
	is a safe cover of $\Delta$ \wrt\ $\spcr$.
\end{propositionrep}
\begin{proof}
$\bigcup_{k=0}^{m}\Delta_k = \Delta$ holds by definition, and for $i\in\{0,\dots, m\}$, the fact that $\Delta_i$ is closed under $\directlyDependsOfWrt{\Delta}{\spcr}$ follows from Proposition~\ref{prop_closedunderdepends}.
\end{proof}

To construct a safe conditional constraint splitting from a safe cover, the conditional pivot $\DeltaZero$ needs to be constructed by taking into account the overlap between any two subbases of the safe cover.
\begin{propositionrep}
	\label{prop_coversplit}
	If $\{\Delta_1,\dots,\Delta_m\}$ is a safe cover of $\Delta$ \wrt\ $\spcr$, then
	$\niceSafeCondConstraintSplitGen$ with 
	\[
	\DeltaZero= \bigcup_{\substack{i,j\in\{1,\dots,n\}\\ i\neq j}}(\Delta_i\cap\Delta_j).
	\]
\end{propositionrep}
\begin{proof}
$\bigcup_{k=0}^{m}(\Delta_k\cup\DeltaZero) = \Delta$ holds by definition.
	First we show 
	$\condConstraintSplitGen$.
	We need to show that $\conditionalIndepGen$ holds for $i,j\in\{1,\dots,m\}, i\neq j$. 
	Because $\{\Delta_1,\dots,\Delta_m\}$ is a safe cover, each $\Delta_i$ is closed under $\directlyDependsOfWrt{\Delta}{\spcr}$.
	Towards a contradiction, assume there are $i,j\in\{1,\dots,m\},i\neq j$ with
	$(\Delta_i\setminus\DeltaZero)\directlyDependsOfWrt{\Delta\setminus\DeltaZero}{\spcr}(\Delta_j\setminus\DeltaZero)$. 
	Then also 
	$(\Delta_i\setminus\DeltaZero)\directlyDependsOfWrt{\Delta}{\spcr}(\Delta_j\setminus\DeltaZero)$.
	Then there are conditionals $\delta_i,\delta_j$ with $\delta_i\directlyDependsOfWrt{\Delta}{\spcr}\delta_j$ and
	$\delta_i\in\Delta_i\setminus\DeltaZero,\delta_j\in\Delta_j\setminus\DeltaZero$.
	Because $\Delta_i$ %
	is closed under $\directlyDependsOfWrt{\Delta}{\scpr}$ it must be the case that $\delta_i\in\Delta_j$.
	But per construction, $\delta_i\in\Delta_i$ and $\delta_i\in\Delta_j$ would require $\delta_i\in\DeltaZero$ which is in contradiction to $\delta_i\in\Delta_i\setminus\DeltaZero$. 
	With the same arguments we obtain for $i=0$ that $\delta_j\in\DeltaZero$, meaning for all $j\in\{1,\dots,m\}$ we have $\DeltaZero\not\directlyDependsOfWrt{\Delta}{\spcr}\Delta_j$ and thus $\niceSafeCondConstraintSplitGen$. %
\end{proof}

Propositions~\ref{prop_splitcover} and \ref{prop_coversplit} confirm that our approach to conditional constraint splittings can be utilized to compute c-representations locally.
On the other hand, our comparison underlines the concept of safe covers by showing that they can be understood as a conditional splitting of constraints derived in analogy to notions of conditional syntax splitting.

Safety of decompositions of conditional belief bases is also a major topic in \cite{KernIsbernerEtalKR2026}. However, the focus of  that paper is different from (though complementary to) this article in that it mainly deals with (conditional) splittings of the subsignatures underlying the subbases. In that paper, subbases are usually disjoint but their respective subsignatures overlap. In particular, no pivot subbase $\Delta_0$ that contains all intersections of subbases (see Proposition \ref{cor_onlyDZero}) is relevant there because these intersections may usually be empty. However in \cite{KernIsbernerEtalKR2026}, the intersections of the subsignatures are usually not empty and play an important role to transfer information (for the constraint systems) from one subbase to others.
Nevertheless, this article and the paper \cite{KernIsbernerEtalKR2026} both use constraint splittings  \cite{BeierleHaldimannKernIsberner2021Boeger75Festschrift} as a starting point and are related to safe covers \cite{WilhelmKernIsbernerAAAI24}. Proposition 15 of \cite{KernIsbernerEtalKR2026} and its proof show that safe decompositions of conditional belief bases as defined in \cite{KernIsbernerEtalKR2026} yield constraint splittings and safe covers.

Splittings at the semantical level beyond c-representations have been investigated in the literature utilizing the notion of model combinations \cite{BeierleHaldimannKernIsberner2021Boeger75Festschrift} as a means to combine ranking functions via the operators $\oplus$ and $\ominus$.
\begin{align*}
	M_1 \oplus M_2 &= \{\kappa \mid \kappa(\omega) = \kappa_1(\omega) + \kappa_2(\omega),
	\kappa_1 \in M_1, \kappa_2 \in M_2 \}
	\\
	M_1 \ominus M_2 &= \{\kappa \mid \kappa(\omega) = \kappa_1(\omega) - \kappa_2(\omega),
	\kappa_1 \in M_1, \kappa_2 \in M_2 \}
\end{align*}
These combinations were employed to define the notion of semantic splitting \cite{BeierleHaldimannKernIsberner2021Boeger75Festschrift}. 
If a \BB\ $\Delta$ semantically splits into a partition $\{\Delta_1,\Delta_2\}$ \wrt\ a semantics $Sem$, then the models of $\Delta$
can be obtained by combining the models of the subbases $\Delta_1,\Delta_2$ as stated in the following equation.
\begin{equation}
	Mod_\Sigma^{\textit{Sem}}(\R) = \Mod_\Sigma^{\textit{Sem}}(\R_1) \oplus \Mod_\Sigma^{\textit{Sem}}(\R_2)
\end{equation}
This notion was extended to conditional semantic splitting \cite{BeierleSpiegelHaldimannWilhelmHeyninckKernIsberner2024KR,BeierleSpiegelHaldimannWilhelmHeyninckKernIsberner2026EAI} where the subbases $\Delta_1$ and $\Delta_2$ may overlap. This overlap means that the intersection of the subbases is ``counted'' twice and must be subtracted again to obtain models of $\Delta$ as stated in the next equation.
\begin{equation}
	\label{eq_csemsplit}
	\Mod_\Sigma^{\textit{Sem}}(\R) = \Mod_\Sigma^{\textit{Sem}}(\R_1) \oplus \Mod_\Sigma^{\textit{Sem}}(\R_2) \ominus  \Mod_\Sigma^{\textit{Sem}}(\R_1\cap \R_2)
\end{equation}

As Example~\ref{ex_nocondsem} shows, a straightforward adaptation of \eqref{eq_csemsplit} to the conditional case for c-representations fails, as we need to additionally demand that the c-representations for the subbases match on their intersection. Thus, conditional semantic splittings are not immediately applicable for our approach.

\section{Conclusion and Future Work}
\begin{figure}[tb!]
	\centering
	\begin{tikzpicture}[scale=0.8]
		\node (TS) at (0,0) {trivial \Aufteilung*};
		\node (SS) at (-6,-1.5) {syntax \Aufteilung*};
		\node (CS) at (6,-1.5) {case splitting*};
		\node (SCSS) at (-6,-3) {safe cond. syntax \Aufteilung*};
		\node (sCSP) at (0,-3) {strong constraint \Aufteilung}; 
		\node (gsCSS) at (-6,-4.5) {gen. safe cond. syntax \Aufteilung*}; 
		\node (CSS) at (-6,-6) {cond. syntax \Aufteilung*};
		\node (CSP) at (0,-4.5) {constraint \Aufteilung};
		\node (cSS) at (6,-6) {\cSolSplitAbr};
		\node (scCSP) at (0,-6) {\textbf{\niceSafeSplittAbr}};
		\node (scCSP2) at (0,-7.5) {\textbf{safe cond. constraint \Aufteilung}};
		\node (cCSP) at (0,-9) {\textbf{cond. constraint \Aufteilung}};
		\node (SC) at (-6,-7.5) {safe cover};
		\node (sccSS) at (6,-7.5) {\textbf{\safecondCSolSplitAbr}};
		
		\usetikzlibrary{arrows.meta}
		\draw[->] (TS.south west) -- (SS);
		\draw[->] (SS) -- (SCSS);
		\draw[->] (SCSS) -- (gsCSS);
		\draw[->] (gsCSS) -- (CSS);
		\draw[->] (SCSS.south east) -- (CSP);
		\draw[->] (sCSP) -- (CSP);
		\draw[->] (CSP.south east) -- (cSS);
		\draw[->] (TS.south east) -- (CS);
		\draw[->] (CS.south west) -- (sCSP);
		\draw[line width=.25mm, -Latex] (CSP) -- (scCSP);
		\draw[line width=.25mm, -Latex] (gsCSS.south east) -- (scCSP);
		\draw[line width=.25mm, Latex-Latex] (scCSP) -- (scCSP2);
		\draw[line width=.25mm, -Latex] (scCSP2) -- (cCSP);
		\draw[line width=.25mm, Latex-Latex] (scCSP) -- (SC);
		\draw[line width=.25mm, -Latex] (cSS) -- (sccSS);
		\draw[line width=.25mm, dotted, -Latex] (scCSP) -- (sccSS);
	\end{tikzpicture}
	\caption{
		Hierarchy of splittings of \BBs.
		Splitting techniques marked with $\ast$ are general splitting techniques independent of the semantics
		of c-representations.
		A solid arrow ($S_1 \rightarrow S_2$) indicates that each splitting of type $S_1$ yields also a splitting of type $S_2$ while dotted arrows (\protect\tikz[baseline=-0.5ex]{\protect\draw[line width=.25mm, dotted, -Latex] (0,0) -- (0.5,0);}) indicate that this holds in special cases (cf. Propositions~\ref{prop_idsol} and \ref{prop_subsetspcrcsolsplit}).
		The splitting notions in bold and the bold arrows (\protect\tikz[baseline=-0.5ex]{\protect\draw[line width=.25mm, -Latex] (0,0) -- (0.5,0);}, \protect\tikz[baseline=-0.5ex]{\protect\draw[line width=.25mm, dotted, -Latex] (0,0) -- (0.5,0);})
		are introduced and shown in this article; for the other splitting notions see \cite{JH_Parikh99,KernIsbernerBeierleBrewka2020KR,BeierleHaldimannKernIsberner2021Boeger75Festschrift,HeyninckKernIsbernerMeyerHaldimannBeierle2023AAAI,WilhelmSezginKernIsbernerHaldimannBeierleHeyninck2023JELIA,WilhelmKernIsbernerAAAI24,SpiegelHaldimannHeyninckKernIsbernerBeierle2025IJCAI}.}
		\label{fig_splithierarchy}
\end{figure}
 In this article we applied
generalizations from syntax splitting to the notion of constraint splitting for c-representations by
introducing conditional constraint splittings, allowing the subbases to also share conditionals. 
We identified the subclasses of safe and \niceSafe\ conditional constraint splittings, allowing the constraint system to be solved locally on the subbases, provided the constraint system of the conditional pivot is solved first.
Furthermore, we applied these techniques to c-solution splittings, leading to the definition of safe conditional c-solution splittings and generalizing c-solutions splittings to the case where subbases may share conditionals. 
We thoroughly investigated the connection between conditional syntax splittings, conditional constraint splittings, and conditional c-solution splittings. Our investigations showed that each generalized safe conditional syntax splitting yields a \niceSafe\ conditional constraint splitting, and that \niceSafe\ conditional constraint splittings \wrt\ the solution preserving constraint reductions given by the identity function and $\spcrSubset=\{R1,R2\}$ yield conditional c-solution splittings.
Additionally, we compared our approaches to
safe covers.
Our results allowed us to embed our approaches in the hierarchy shown in Figure~\ref{fig_splithierarchy}.

For future work we plan to exploit these new concepts for the computation and implementation of inference with c-representations in the online reasoning platform InfOCF-Web\footnote{https://www.fernuni-hagen.de/wbs/research/infocf-web/index.html} and its underlying library \cite{BeierleHaldimannSaninSchwarzerSpangSpiegelvonBerg2024SUM,BeierleHaldimannSaninSpangSpiegelvonBerg2025JELIA}.
In particular, we want to investigate the difference in computational gain between safe conditional constraint and conditional solution splittings.
We will also extend the investigations presented here by considering not only strongly consistent \BBs\ but also \BBs\ that are only weakly consistent \cite{HaldimannBeierleKernIsberner2024FoIKS,HaldimannBeierleKernIsberner2025KER}. 